\documentclass{article}

\PassOptionsToPackage{numbers, compress}{natbib}

\usepackage[preprint]{neurips_2023}

\usepackage[utf8]{inputenc} 
\usepackage[T1]{fontenc}    
\usepackage{hyperref}       
\usepackage{url}            
\usepackage{booktabs}       
\usepackage{amsfonts}       
\usepackage{nicefrac}       
\usepackage{microtype}      
\usepackage{xcolor}         

\usepackage{amsthm, amsmath, amssymb}
\usepackage{bm}
\usepackage{epsfig}
\usepackage{graphicx}
\usepackage{multicol, multirow, makecell}
\usepackage{subcaption}

\title{KernelWarehouse: Towards Parameter-Efficient Dynamic Convolution}

\author{%
  Chao Li\thanks{This work was done by Chao Li when he was an intern at Intel Labs China. Anbang Yao led the project and the paper writing. $^\dagger$ Corresponding author.}\\
  Intel Labs China\\
  \texttt{chao.li3@intel.com}\\
  \And
  Anbang Yao$^\dagger$\\
  Intel Labs China\\
  \texttt{anbang.yao@intel.com} \\
}

\begin{document}

\maketitle

\begin{abstract}
  Dynamic convolution learns a linear mixture of $n$ static kernels weighted with their sample-dependent attentions, demonstrating superior performance compared to normal convolution. However, existing designs are parameter-inefficient: they increase the number of convolutional parameters by $n$ times. This and the optimization difficulty lead to no research progress in dynamic convolution that can allow us to use a significant large value of $n$ (e.g., $n>100$ instead of typical setting $n<10$) to push forward the performance boundary. In this paper, we propose~\textbf{KernelWarehouse}, a more general form of dynamic convolution, which can strike a favorable trade-off between parameter efficiency and representation power. Its key idea is to redefine the basic concepts of "\textbf{kernels}" and "\textbf{assembling kernels}" in dynamic convolution from the perspective of reducing kernel dimension and increasing kernel number significantly. In principle, KernelWarehouse enhances convolutional parameter dependencies within the same layer and across successive layers via tactful kernel partition and warehouse sharing. Specifically, KernelWarehouse sequentially divides a static kernel at any convolutional layer of a ConvNet into $m$ disjoint kernel cells having the same dimensions first, and then computes each kernel cell as a linear mixture based on a predefined "\textbf{warehouse}" consisting of $n$ kernel cells (e.g., $n=108$) which is also shared to multiple neighboring convolutional layers, and finally replaces the static kernel by assembling its corresponding $m$ mixtures in order, yielding a high degree of freedom to fit a desired parameter budget. To facilitate the learning of the attentions for summing up kernel cells, we also present a new attention function. We validate our method on ImageNet and MS-COCO datasets with different ConvNet architectures, and show that it attains state-of-the-art results. For instance, the ResNet18|ResNet50|MobileNetV2|ConvNeXt-Tiny model trained with KernelWarehouse on ImageNet reaches 76.05\%|81.05\%|75.92\%|82.51\% top-1 accuracy. Thanks to its flexible design, KernelWarehouse can even reduce the model size of a ConvNet while improving the accuracy, e.g., our ResNet18 model with 36.45\%|65.10\% parameter reduction to the baseline shows 2.89\%|2.29\% absolute improvement to top-1 accuracy. Code and pre-trained models are available at \url{https://github.com/OSVAI/KernelWarehouse}.

  \end{abstract}

\section{Introduction}

Convolution is a fundamental operation in 
modern convolutional neural networks (ConvNets)~\citep{CNN_AlexNet,CNN_GoogLeNet,CNN_ResNet,CNN_MobileNets,CNN_MobileNetV2,CNN_ConvNeXt}. In a convolutional layer, normal convolution $\mathbf{y} = \mathbf{W}*\mathbf{x}$ computes the output $\mathbf{y}$ by applying the same convolutional kernel $\mathbf{W}$ defined as a set of convolutional filters to every input sample $\mathbf{x}$. \textit{For brevity, we refer to "convolutional kernel" as "kernel" and omit the bias term throughout this paper}. Although the efficacy of normal convolution is extensively validated with various types of ConvNets in numerous 
computer vision applications, recent progress in the efficient ConvNet architecture design shows that dynamic convolution, also known as CondConv in~\citep{DynamicConv_CondConv} and DY-Conv in~\citep{DynamicConv_DyConv}, achieves large performance gains. The basic idea of dynamic convolution is to replace the static kernel in normal convolution by a linear mixture of $n$ same dimensioned kernels $\mathbf{W}=\alpha_{1}\mathbf{W}_1+...+\alpha_{n}\mathbf{W}_n$, where $\alpha_{1},...,\alpha_{n}$ are scalar attentions generated by an input-dependent attention module. Benefiting from the additive property of $n$ kernels $\mathbf{W}_1,...,\mathbf{W}_n$ and compact attention module designs, dynamic convolution improves the feature learning ability with little extra multiply-add cost. However, it increases the number of convolutional parameters by $n$ times, which leads to a huge rise in model size because the convolutional layers of a modern ConvNet occupy the vast majority of parameters. To alleviate this problem, DCD
~\citep{DynamicConv_ResDyConv} directly learns a base kernel and a sparse residual to approximate dynamic convolution via matrix decomposition. This coarse approximation abandons the basic mixture learning paradigm
, and cannot well retain the representation power of dynamic convolution when $n$ becomes large. 
Recently, ODConv~\citep{DynamicConv_ODConv} presents a more powerful attention module to dynamically weight static kernels along different dimensions instead of one single dimension, which can get competitive performance with a reduced number of kernels. Under the same setting of $n$, ODConv has more parameters than vanilla dynamic convolution. In brief, existing dynamic convolution methods based on the mixture learning paradigm 
are not parameter-efficient, and they typically set $n=8$~\citep{DynamicConv_CondConv} or $n=4$~\citep{DynamicConv_DyConv,DynamicConv_ODConv}. More importantly, a plain fact is that the improved capacity of a ConvNet constructed with dynamic convolution comes from increasing the number of kernels per convolutional layer facilitated by the attention mechanism.~\textit{This causes an intrinsic conflict between the desired model size and capacity, which prevents researchers to explore the performance boundary of dynamic convolution with a significantly large kernel number (e.g., $n>100$). In this paper, we attempt to break down this barrier, and shed new insights on research in dynamic convolution}.

To this end, we present \textit{KernelWarehouse} (Figure~\ref{fig:architecture}), a more general form of dynamic convolution, which can strike a favorable trade-off between parameter efficiency and representation power. 
The formulation of KernelWarehouse is inspired by two observations. On the one hand, we note that existing dynamic convolution methods treat all parameters in a regular convolutional layer as a kernel, and increase the kernel number from 1 to $n$, and use their attention modules to assemble $n$ kernels into a linearly mixed kernel. Though straightforward and effective, they pay no attention to parameter dependencies within the static kernel at a convolutional layer. On the other hand, we notice that existing dynamic convolution methods allocate a different set of $n$ kernels for every individual convolutional layer of a ConvNet, 
ignoring parameter dependencies across successive layers. We hypothesize that the barrier discussed above can be removed by way of flexibly leveraging these two types of convolutional parameter dependencies to reformulate dynamic convolution. Driven by this analysis, we introduce two simple strategies when formulating KernelWarehouse, namely kernel partition and 
warehouse sharing. By them, 
we redefine the basic concepts of "\textit{kernels}" and "\textit{assembling kernels}" in dynamic convolution from the perspective of reducing kernel dimension and increasing kernel number significantly, taking advantage of the aforementioned parameter dependencies as easy as possible. With kernel partition, the static kernel for a regular convolutional layer is sequentially divided into $m$ disjoint kernel cells of the same dimensions, and then the linear mixture can be defined in terms of much smaller kernel cells instead of holistic kernels. Specifically, each of $m$ kernel cells is represented as a linear mixture based on a predefined "\textit{warehouse}" consisting of $n$ kernel cells (e.g., $n=108$), and the static kernel will be replaced by assembling its corresponding $m$ mixtures in order. With 
warehouse sharing, multiple neighboring convolutional layers of a ConvNet can share the same warehouse as long as the same kernel cell size is used in the kernel partition process, further enhancing its parameter efficiency and representation power
. Nevertheless, with a significantly large value of $n$, the optimization of KernelWarehouse becomes more challenging than existing methods. We find that popular attention functions for dynamic convolution
~\citep{DynamicConv_CondConv,DynamicConv_DyConv,DynamicConv_ODConv} do not work well under this circumstance. We solve this problem by designing a new and simple attention function. These simple concept shifts provide a high degree of flexibility to balance parameter efficiency and representation power of KernelWarehouse under different parameter budgets.

As a drop-in replacement of normal convolution, KernelWarehouse can be easily used to different ConvNet architectures. Extensive experiments on ImageNet and MS-COCO show that our method achieves better results than its dynamic convolution counterparts.

\section{Related Work}

\textbf{ConvNet Architectures.} In the past decade, a lot of pioneering ConvNet architectures such as AlexNet~\citep{CNN_AlexNet}, VGGNet~\citep{CNN_VGGNet}, GoogLeNet~\citep{CNN_GoogLeNet}, ResNet~\citep{CNN_ResNet}, DenseNet~\citep{CNN_DenseNet}, ResNeXt~\citep{CNN_ResNeXt} and RegNet~\citep{CNN_RegNet} have been presented, which significantly advance the performance on the ImageNet benchmark. Around the same time, some lightweight architectures like MobileNet~\citep{CNN_MobileNets,CNN_MobileNetV2,CNN_MobileNetV3}, ShuffleNet~\citep{CNN_ShuffleNet} and EfficientNet~\citep{CNN_EfficientNet} have been designed for resource-constrained applications. Recently, Liu et al. presented ConvNeXt~\citep{CNN_ConvNeXt} whose performance can match newly emerging vision transformers~\citep{SelfAttention_ViT,SelfAttention_Swin}. As a plug-and-play design, our method could be used to improve their performance. 

\textbf{Feature Recalibration.} An effective way to enhance the capacity of a ConvNet is feature recalibration. It relies on attention mechanisms to adaptively refine the feature maps learnt by a convolutional block. Popular feature recalibration modules such as RAN~\citep{Attention_ResidualAttention}, SE~\citep{Attention_SE}, BAM~\citep{Attention_BAM}, CBAM~\citep{Attention_CBAM}, GE~\citep{Attention_GE}, SRM~\citep{Attention_SRM} and ECA~\citep{Attention_ECA} focus on different design aspects: using channel attention, or spatial attention, or hybrid attention to emphasize important features and suppress necessary ones. Differing from feature recalibration in multiple aspects, dynamic convolution replaces the static kernel of a convolutional layer by a linear mixture of $n$ kernels weighted with the attention mechanism.

\textbf{Dynamic Weight Networks.} In the field of deep learning, many research efforts have been made on developing effective methods to generate the weights for a neural network. Jaderberg et al.~\citep{WeightGeneration_STNetworks} propose a Spatial Transformer module which uses a localisation network that predicts the feature transformation parameters conditioned on the learnt feature itself. Dynamic Filter Network~\citep{WeightGeneration_DynamicFilterNetworks} and Kernel Prediction Networks~\citep{WeightGeneration_KPN, WeightGeneration_KPN_BurstDenoising} introduce two filter generation frameworks which share the same idea: using a deep neural network to generate sample-adaptive filters conditioned on the input. Based on this idea, DynamoNet~\citep{WeightGeneration_DynamoNet}  uses dynamically generated motion filters to boost human action recognition in videos. CARAFE~\citep{WeightGeneration_CARAFE} and Involution~\citep{WeightGeneration_Involution} further extend this idea by designing efficient generation modules to predict the weights for extracting informative features. By connecting this idea with SE, WeightNet~\citep{WeightGeneration_WeightNet}, CGC~\citep{DynamicConv_CGC} and WE~\citep{DynamicConv_WeightExcitation} design different attention modules to adjust the weights in convolutional layers of a ConvNet. Unlike these methods, hypernetwork~\citep{WeightGeneration_Hypernetworks} uses a small network to generate the weights for a larger recurrent network, and MetaNet~\citep{WeightGeneration_MetaNetworks} introduces a meta learning model consisting of a base learner and a meta learner, allowing the learnt network for rapid generalization across different tasks. This paper focuses on advancing dynamic convolution research, which differs from these works in formulation.

\section{Method}

\begin{figure*}[t]
\vskip -0.12 in
	\begin{center}
\includegraphics[width=0.93\linewidth]{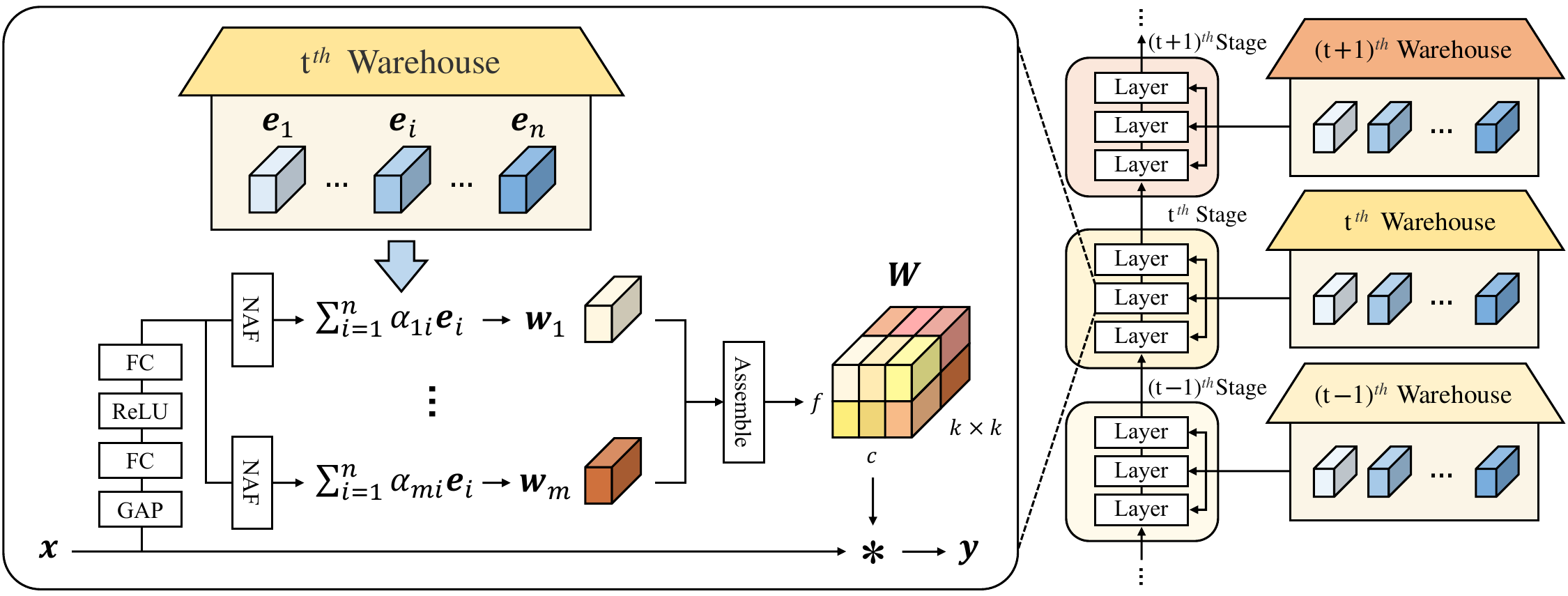}
	\end{center}
\vskip -0.11in
	\caption{Schematic illustration of KernelWarehouse. Briefly speaking, KernelWarehouse sequentially divides the static kernel $\mathbf{W}$ at any regular convolutional layer of a ConvNet into $m$ disjoint kernel cells $\mathbf{w}_1$,...,$\mathbf{w}_m$ having the same dimensions first, and then computes each kernel cell $\mathbf{w}_i$ as a linear mixture $\mathbf{w}_i=\alpha_{i1} \mathbf{e}_1+...+\alpha_{in}\mathbf{e}_n$ based on a predefined "\textit{warehouse}" (consisting of $n$ same dimensioned kernel cells $\mathbf{e}_1,...,\mathbf{e}_n$, e.g., $n=108$) which is shared to all same-stage convolutional layers, and finally replaces the static kernel $\mathbf{W}$ by assembling its corresponding $m$ mixtures in order, yielding a high degree of freedom to fit a desired convolutional parameter budget. The input-dependent scalar attentions $\alpha_{i1},...,\alpha_{in}$ are computed with a new attention function (NAF).}   
	\label{fig:architecture}
\vskip -0.21in
\end{figure*}

Figure~\ref{fig:architecture} shows a schematic illustration of our method called \textbf{KernelWarehouse}, which has three key components, namely kernel partition, warehouse sharing and a new attention function. Next, we describe the formulation of KernelWarehouse and clarify these components.


\subsection{Formulation of KernelWarehouse}


For a convolutional layer, let $\mathbf{x} \in \mathbb{R}^{h \times w \times c}$ and $\mathbf{y} \in \mathbb{R}^{h \times w \times f}$ be the input having $c$ feature channels with the resolution $h \times w$ and the output having $f$ feature channels with the same resolution to the input, respectively. Normal convolution $\mathbf{y} = \mathbf{W}*\mathbf{x}$ uses a single static kernel $\mathbf{W} \in \mathbb{R}^{k \times k \times c \times f}$ consisting of $f$ convolutional filters with the spatial size $k\times k$, while dynamic convolution replaces this static kernel by $\mathbf{W}=\alpha_{1} \mathbf{W}_1+...+\alpha_{n} \mathbf{W}_n$, a linear mixture of $n$ same dimensioned static kernels weighted with their input-dependent scalar attentions $\alpha_{1},...,\alpha_{n}$. Like existing dynamic convolution methods~\citep{DynamicConv_CondConv,DynamicConv_DyConv,DynamicConv_ODConv}, KernelWarehouse also adopts the attentive mixture learning paradigm. 
Unlike them, KernelWarehouse applies the attentive mixture learning paradigm to a static kernel in dense local scales instead of a single holistic scale via kernel partition and warehouse sharing. 

\textbf{Kernel Partition.} The basic idea of kernel partition is to reduce kernel dimension and increase kernel number via explicitly enhancing parameter dependencies within the same convolutional layer. First, we sequentially divide the static kernel $\mathbf{W}$ at a regular convolutional layer into $m$ disjoint parts $\mathbf{w}_1$,...,$\mathbf{w}_m$ called "\textit{kernel cells}" that have the same dimensions.~\textit{For brevity, here we omit to define kernel cell dimensions}. Kernel partition can be defined as
\begin{equation}
\label{eq:01}
 \mathbf{W} = \mathbf{w}_1\cup...\cup \mathbf{w}_m, \ \mathrm{and}\ \forall\ i,j\in\{1,...,m\}, i \ne j, \ \mathbf{w}_i \cap \mathbf{w}_j = \mathbf{\emptyset}.
\end{equation}
After kernel partition, we treat kernel cells $\mathbf{w}_1$,...,$\mathbf{w}_m$ as "\textit{local kernels}", and define a "\textit{warehouse}" containing $n$ kernel cells $\mathbf{E}=\{\mathbf{e}_1,...,\mathbf{e}_n\}$, where $\mathbf{e}_1,...,\mathbf{e}_n$ have the same dimensions as $\mathbf{w}_1$,...,$\mathbf{w}_m$. Then, by sharing the warehouse $\mathbf{E}=\{\mathbf{e}_1,...,\mathbf{e}_n\}$ in the same convolutional layer, we represent each of $m$ kernel cells $\mathbf{w}_1$,...,$\mathbf{w}_m$ as
\begin{equation}
\label{eq:02}
\mathbf{w}_i =\alpha_{i1} \mathbf{e}_1+...+\alpha_{in} \mathbf{e}_n, \ \mathrm{and}\ i\in\{1,...,m\},
\end{equation}
where $\alpha_{i1},...,\alpha_{in}$ are the scalar attentions generated by an attention module $\phi(x)$ conditioned on the input $\mathbf{x}$. Finally, the static kernel $\mathbf{W}$ in a regular convolutional layer is replaced by assembling its corresponding $m$ linear mixtures in order. Note that the dimensions of a kernel cell can be much smaller than the dimensions of the static kernel $\mathbf{W}$ (e.g., when $m=16$, the number of parameters in a kernel cell is 1/16 to that of the static kernel $\mathbf{W}$). Under the same parameter budget, this allows a warehouse can have a large setting of $n$ (e.g., $n>100$), in sharp contrast to existing dynamic convolution methods which define the linear mixture in terms of $n$ "\textit{holistic kernels}" and typically uses a much smaller setting of $n$ (e.g., $n<10$) due to its parameter-inefficient shortcoming.

\textbf{Warehouse Sharing.} Following the simple design philosophy of kernel partition, the main goal of warehouse sharing is to further improve parameter efficiency and representation power of KernelWarehouse through explicitly enhancing parameter dependencies across successive convolutional layers. Specifically, we share a warehouse $\mathbf{E}=\{\mathbf{e}_1,...,\mathbf{e}_n\}$ to represent every kernel cell at $l$ neighboring convolutional layers in the same-stage building blocks of a ConvNet, allowing it can use a much larger setting of $n$ 
against kernel partition. This is mostly easy, as modern ConvNets such as ResNet~\citep{CNN_ResNet}, MobileNet~\citep{CNN_MobileNets,CNN_MobileNetV2,CNN_MobileNetV3} and ConvNeXt~\citep{CNN_ConvNeXt} typically adopt a modular design scheme in which static kernels in the same-stage convolutional layers usually have the same or similar dimensions.~\textit{In implementation, given a desired convolutional parameter budget, we simply use common dimension divisors for $l$ static kernels as the uniform kernel cell dimensions for kernel partition, which naturally determine the values of $m$ and $n$}. On the other hand, parameter dependencies across neighboring convolutional layers are also critical in strengthening the capacity of a ConvNet (see Table~\ref{table:ablation_sharingscale}).

\textbf{Parameter Efficiency and Representation Power.} Let $n$ be the number of kernel cells in a warehouse shared to $l$ convolutional layers of a ConvNet, and let $m_{t}$ be the total number of kernel cells in these $l$ convolutional layers ($m_{t} = m$, when $l = 1$). Then,~\textit{$b=n/m_{t}$ can be server as a scaling factor to indicate the convolutional parameter budget of KernelWarehouse relative to normal convolution}. Here, we do not consider the number of parameters in the attention module $\phi(x)$ which generates $nm_{t}$ scalar attentions, as it is much smaller than the total number of parameters for normal convolution at $l$ convolutional layers.~\textit{In implementation, we use the same value of $b$ to all convolutional layers of every ConvNet}. Under such a condition, we can see that KernelWarehouse can easily scale up or scale down the model size of a ConvNet by changing $b$. Compared to normal convolution: (1) when $b<1$, KernelWarehouse tends to get the reduced model size; (2) when $b=1$, KernelWarehouse tends to get the similar model size; (3) when $b>1$, KernelWarehouse tends to get the increased model size. Intriguingly, given a desired value of $b$, a proper and large value of $n$ can be obtained by simply changing $m$, providing a representation power guarantee for KernelWarehouse. Therefore, KernelWarehouse can strike a favorable trade-off between parameter efficiency and representation power, under different convolutional parameter budgets in terms of $b$.

\textbf{KernelWarehouse vs. Dynamic Convolution.} According to its above formulation, KernelWarehouse will degenerate into vanilla dynamic convolution~\citep{DynamicConv_CondConv,DynamicConv_DyConv} when uniformly setting $m=1$ in kernel partition (i.e., all kernel cells in each warehouse have the same dimensions as the static kernel $\mathbf{W}$ in normal convolution) and $l=1$ in warehouse sharing (i.e., each warehouse is limited to its specific convolutional layer). Therefore, KernelWarehouse is a more general form of dynamic convolution.

\subsection{Attention Module of KernelWarehouse}

Following existing dynamic convolution methods~\citep{DynamicConv_CondConv,DynamicConv_DyConv,DynamicConv_ODConv}, KernelWarehouse also adopts a compact SE-typed structure as the attention module $\phi(x)$ (illustrated in Figure~\ref{fig:architecture}) to generate attentions for weighting kernel cells in a warehouse. For any convolutional layer with a static kernel $\mathbf{W}$, it starts with a channel-wise global average pooling (GAP) operation that maps the input $\mathbf{x}$ into a feature vector, followed by a fully connected (FC) layer, a rectified linear unit (ReLU), another FC layer, and a new attention function. The first FC layer reduces the length of the feature vector by 16, and the second FC layer generates $m$ sets of $n$ feature logits in parallel which are finally normalized by the attention function set by set.

\textbf{Attention Function.} Comparatively, 
the optimization of KernelWarehouse is more challenging, mainly due to its new formulation. For the linear mixture in KernelWarehouse, (1) the number of kernel cells in a warehouse is much larger (e.g., $n=108$ vs. $n<10$); (2) a warehouse is not only shared to represent each of $m$ kernel cells for a specific convolutional layer of a ConvNet, but also is shared to represent every kernel cell for the other $l-1$ same-stage convolutional layers. Under this circumstance, a proper choice of the attention function is critical. An ideal attention function should have the property of easily allocating diverse attentions for all linear mixtures simultaneously, making the mixed kernel cells at $l$ convolutional layers can learn informative features hieratically. However, we experimentally find that popular attention functions such as Sigmoid and Softmax do not work well for KernelWarehouse. Even with the temperature annealing~\citep{DynamicConv_DyConv}, 
they get worse results than vanilla dynamic convolution (see Table~\ref{table:classification_cnn_300epoch} and Table~\ref{table:ablation_function}). To address this joint optimization problem, we present a new attention function. For $i^{th}$ kernel cell in the static kernel $\mathbf{W}$, let $z_{i1},...,z_{in}$ be the feature logits generated by the second FC layer, then our attention function is defined as
\begin{equation}
\label{eq:03}
\alpha_{ij} = (1-\tau) \frac{z_{ij}}{\sum^{n}_{p=1}{|z_{ip}|}} + \tau\beta_{ij}, \ \mathrm{and}\ j\in\{1,...,n\},
\end{equation}
where $\tau$ is a temperature which linearly reduces from $1$ to $0$ in the early training stage; $\frac{z_{ij}}{\sum^{n}_{p=1}{|z_{ip}|}}$ is a linear normalization function which can have negative attention outputs that are essential to encourage the network to learn adversarial attention relationships among all linear mixtures sharing the same warehouse; $\beta_{ij}$ is a binary value (0 or 1) which is used for initializing the attentions. Note that the settings of $\beta_{ij}$ at $l$ convolutional layers are very important, which assure the shared warehouse can assign (1) at least one specified kernel cell ($\beta_{ij}=1$) to every linear mixture, given a desired convolutional parameter budget $b\geq1$, and (2) at most one specific kernel cell ($\beta_{ij}=1$) to every linear mixture, given $b<1$.~\textit{
In implementation, we adopt a simple strategy to assign one of the total $n$ kernel cells in a shared warehouse to each of $m_{t}$ linear mixtures at $l$ convolutional layers without repetition.
When $n$ is less than $m_{t}$,
we let the remaining linear mixtures always have $\beta_{ij}=0$ once $n$ kernel cells are used up.} In the Appendix, we provide visualization examples to illustrate this strategy, and a set of ablative experiments to compare it with other alternatives. Extensive experiments validate the effectiveness of our attention function (see Figure~\ref{figure:attention_resnet18_heatmap} and Table~\ref{table:ablation_function}).


\section{Experiments}
In this section, we conduct comprehensive experiments on image classification, object detection and instance segmentation to evaluate the effectiveness of our KernelWarehouse ("\textbf{KW}" for short), compare it with other attention based methods, and provide lots of ablations to study its design.
\subsection{Image Classification on ImageNet}
Our main experiments are conducted on ImageNet dataset~\citep{Dataset_ImageNet}, which consists of over 1.2 million training images and 50,000 validation images with 1,000 object categories.

\textbf{ConvNet Backbones.} We select backbones from MobileNetV2~\citep{CNN_MobileNetV2}, ResNet~\citep{CNN_ConvNeXt} and ConvNeXt~\citep{CNN_MobileNetV2} families for experiments, including both lightweight networks and larger ones. Specifically, we use MobileNetV2 ($1.0\times$), MobileNetV2 ($0.5\times$), ResNet18, ResNet50 and ConvNeXt-Tiny.

\textbf{Experimental Setup.}
In the experiments, we make comparisons of KernelWarehouse with related methods to demonstrate its effectiveness.
On the ResNet18 backbone, we compare our design with various state-of-the-art attention based methods, including: (1) SE~\citep{Attention_SE}, CBAM~\citep{Attention_CBAM} and ECA~\citep{Attention_ECA}, which focus on recalibration of feature maps; (2) CGC~\citep{DynamicConv_CGC} and WeightNet~\citep{WeightGeneration_WeightNet}, which focus on adjusting convolutional weights; (3) CondConv~\citep{DynamicConv_CondConv}, DY-Conv~\citep{DynamicConv_DyConv}, DCD~\citep{DynamicConv_ResDyConv} and ODConv~\citep{DynamicConv_ODConv}, which focus on dynamic convolution.
%
As DY-Conv and ODConv are state-of-the-art dynamic convolution methods which achieve top performance in our comparative experiments on the ResNet18 and are closely related to KernelWarehouse, we select them as our key reference methods. We compare our method with them on all the other ConvNet backbones except ConvNeXt-Tiny (since there is no publicly available implementation of them on ConvNeXt).
To make fair comparisons, all the methods are implemented using the public codes with the same settings for training and testing. 
\textit{In the experiments, we use $b\times$ to denote the convolutional parameter budget of each dynamic convolution method relative to normal convolution, the values of $n$ and $m$ in KernelWarehouse, and the experimental details for each network are provided in the Appendix.}

\begin{figure}[t]
\vskip -0.10 in
\centering
\begin{minipage}[t!]{0.5\linewidth}
\centering
\vskip -0.6 in
\captionof{table}{Results comparison on ImageNet with the ResNet18 backbone using the traditional training strategy. Best results are bolded.}
\vskip -0.10 in
\label{table:classification_resnet_100epoch}
\begin{center}
\resizebox{0.99\linewidth}{!}{
\begin{tabular}{l|c|c|c}
\hline
Models & Params & Top-1 Acc (\%) & Top-5 Acc (\%)\\
\hline
ResNet18 & 11.69M & 70.25 & 89.38 \\
\hline
+ SE~\citep{Attention_SE} & 11.78M & 70.98 ($\uparrow$0.73) & 90.03 ($\uparrow$0.65) \\
+ CBAM~\citep{Attention_CBAM} & 11.78M & 71.01 ($\uparrow$0.76) & 89.85 ($\uparrow$0.47) \\
+ ECA~\citep{Attention_ECA} & 11.69M & 70.60 ($\uparrow$0.35) & 89.68 ($\uparrow$0.30) \\
+ CGC~\citep{DynamicConv_CGC} & 11.69M & 71.60 ($\uparrow$1.35) & 90.35 ($\uparrow$0.97) \\
+ WeightNet~\citep{DynamicConv_CGC} & 11.93M & 71.56 ($\uparrow$1.31) & 90.38 ($\uparrow$1.00) \\
+ DCD~\citep{DynamicConv_ResDyConv} & 14.70M & 72.33 ($\uparrow$2.08) & 90.65 ($\uparrow$1.27) \\
+ CondConv ($8\times$)~\citep{DynamicConv_CondConv} & 81.35M & 71.99 ($\uparrow$1.74) & 90.27 ($\uparrow$0.89) \\
+ DY-Conv ($4\times$)~\citep{DynamicConv_DyConv} & 45.47M & 72.76 ($\uparrow$2.51) & 90.79 ($\uparrow$1.41) \\
+ ODConv ($4\times$)~\citep{DynamicConv_ODConv} & 44.90M & 73.97 ($\uparrow$3.72) & 91.35 ($\uparrow$1.97) \\
\hline
+ KW ($1/2\times$) & 7.43M & 72.81 ($\uparrow$2.56) & 90.66 ($\uparrow$1.28) \\
+ KW ($1\times$) & 11.93M & 73.67 ($\uparrow$3.42) & 91.17 ($\uparrow$1.79) \\
+ KW ($2\times$) & 23.24M & \textbf{74.03} ($\uparrow$\textbf{3.78}) & \textbf{91.37} ($\uparrow$\textbf{1.99}) \\
+ KW ($4\times$) & 45.86M & 73.54 ($\uparrow$3.29) & 90.94 ($\uparrow$1.56) \\
\hline
\end{tabular}
}
\end{center}
\vskip -0.0 in
\end{minipage}
\hfill
\begin{minipage}[t!]{0.48\linewidth}
\begin{center}
\vskip 0.1 in
\includegraphics[width=0.99\linewidth, trim=0 8 0 0, clip]{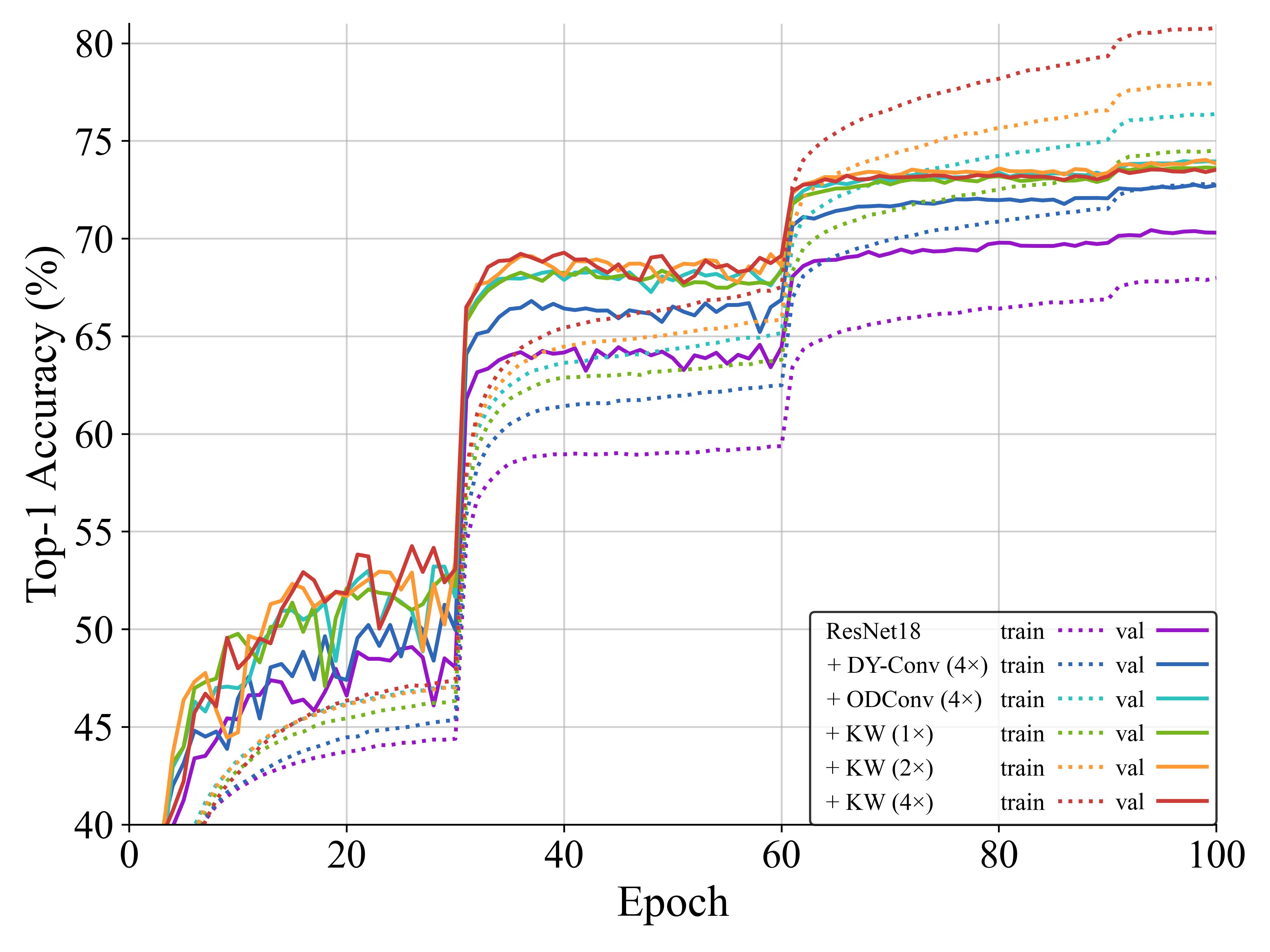}
\vskip 0.1 in
\caption{Curves of top-1 training accuracy and validation accuracy of ResNet18 models trained for 100 epochs on ImageNet with DY-Conv ($4\times$), ODConv ($4\times$) and KW ($1\times, 2\times, 4\times$).
}
\label{fig:training_curve}
\end{center}
\vskip -0.07 in
\end{minipage}

\begin{minipage}[t]{0.5\linewidth}
\vskip - 0.55 in
\captionof{table}{Results comparison on ImageNet with the ResNet18, ResNet50 and ConvNeXt-Tiny backbones trained using the advanced training strategy proposed in ConvNeXt. Best results are bolded.}
\vskip -0.105 in
\label{table:classification_cnn_300epoch}
\begin{center}
\resizebox{0.99\linewidth}{!}{
\begin{tabular}{l|c|c|c}
\hline
Models & Params & Top-1 Acc (\%) & Top-5 Acc (\%)\\
\hline
ResNet18 & 11.69M & 70.44 & 89.72 \\
+ DY-Conv ($4\times$) & 45.47M & 73.82 ($\uparrow$3.38) & 91.48 ($\uparrow$1.76) \\
+ ODConv ($4\times$) & 44.90M & 74.45 ($\uparrow$4.01) & 91.67 ($\uparrow$1.95)\\
+ KW ($1/4\times$) & 4.08M & 72.73 ($\uparrow$2.29) & 90.83 ($\uparrow$1.11) \\
+ KW ($1/2\times$) & 7.43M & 73.33 ($\uparrow$2.89) & 91.42 ($\uparrow$1.70) \\
+ KW ($1\times$) & 11.93M & 74.77 ($\uparrow$4.33) & 92.13 ($\uparrow$2.41) \\
+ KW ($2\times$) & 23.24M & 75.19 ($\uparrow$4.75) & 92.18 ($\uparrow$2.46) \\
+ KW ($4\times$) & 45.86M & \textbf{76.05} ($\uparrow$\textbf{5.61}) & \textbf{92.68} ($\uparrow$\textbf{2.96}) \\
\hline
ResNet50 & 25.56M & 78.44 & 94.24 \\
+ DY-Conv ($4\times$) & 100.88M & 79.00 ($\uparrow$0.56) & 94.27 ($\uparrow$0.03) \\
+ ODConv ($4\times$) & 90.67M & 80.62 ($\uparrow$2.18) & 95.16 ($\uparrow$0.92)\\
+ KW ($1/2\times$) & 17.64M & 79.30 ($\uparrow$0.86) & 94.71 ($\uparrow$0.47) \\
+ KW ($1\times$) & 28.05M & 80.38 ($\uparrow$1.94) & 95.19 ($\uparrow$0.95) \\
+ KW ($4\times$) & 102.02M & \textbf{81.05} ($\uparrow$\textbf{2.61}) & \textbf{95.21} ($\uparrow$\textbf{0.97}) \\
\hline
ConvNeXt-Tiny & 28.59M & 82.07 & 95.86 \\
+ KW ($1\times$) & 39.37M & \textbf{82.51} ($\uparrow$\textbf{0.44}) & \textbf{96.07} ($\uparrow$\textbf{0.21}) \\
\hline
\end{tabular}
}
\end{center}
\end{minipage}
\hfill
\begin{minipage}[t]{0.48\linewidth}
\vskip 0.13 in
\captionof{table}{Results comparison on ImageNet with the MobileNetV2 ($1.0\times$, $0.5\times$) backbones trained for 150 epochs. Best results are bolded.}
\vskip -0.1 in
\label{table:classification_mobilenetv2}
\begin{center}
\resizebox{1.0\linewidth}{!}{
\begin{tabular}[b]{l|c|c|c}
\hline
Models & Params & Top-1 Acc (\%) & Top-5 Acc (\%) \\
\hline
MobileNetV2 ($1.0\times$) & 3.50M & 72.02 & 90.43  \\
+ DY-Conv ($4\times$) & 12.40M & 74.94 ($\uparrow$2.92) & 91.83 ($\uparrow$1.40)  \\
+ ODConv ($4\times$) & 11.52M & 75.42 ($\uparrow$3.40) & 92.18 ($\uparrow$1.75)   \\
+ KW ($1/2\times$) & 2.65M & 72.59 ($\uparrow$0.57) & 90.71 ($\uparrow$0.28)  \\
+ KW ($1\times$) & 5.17M & 74.68 ($\uparrow$2.66) & 91.90 ($\uparrow$1.47)  \\
+ KW ($4\times$) & 11.38M & \textbf{75.92} ($\uparrow$\textbf{3.90}) & \textbf{92.22} ($\uparrow$\textbf{1.79})  \\
\hline
MobileNetV2 ($0.5\times$) & 1.97M & 64.30 & 85.21 \\
+ DY-Conv ($4\times$) & 4.57M & 69.05 ($\uparrow$4.75) & 88.37 ($\uparrow$3.16) \\
+ ODConv ($4\times$) & 4.44M & 70.01 ($\uparrow$5.71) & 89.01 ($\uparrow$3.80)  \\
+ KW ($1/2\times$) & 1.47M & 65.19 ($\uparrow$0.89) & 85.98 ($\uparrow$0.77) \\
+ KW ($1\times$) & 2.85M & 68.29 ($\uparrow$3.99) & 87.93 ($\uparrow$2.72)  \\
+ KW ($4\times$) & 4.65M & \textbf{70.26} ($\uparrow$\textbf{5.96}) & \textbf{89.19} ($\uparrow$\textbf{3.98}) \\
\hline
\end{tabular}
}
\end{center}
\end{minipage}
\vskip -0.15 in
\end{figure}

\textbf{Results Comparison on ResNets18 with the Traditional Training Strategy.} 
We first use the traditional training strategy adopted by lots of previous studies such as DY-Conv~\citep{DynamicConv_DyConv}, ODConv~\citep{DynamicConv_ODConv}, where models are trained for 100 epochs. The results are shown in Table~\ref{table:classification_resnet_100epoch}.
%
It can be observed that the dynamic convolution methods (CondConv, DY-Conv and ODConv), which introduce obviously more extra parameters, tend to have better performance compared with other methods (SE, CBAM, ECA, CGC, WeightNet and DCD). \textit{Note that our KW ($1/2\times$), which has 36.45\% parameters less than the baseline, can even outperform all the above attention based methods except ODConv ($4\times$), including CondConv ($8\times$) and DY-Conv ($4\times$) which increase the model size to about 6.96|3.89 times. Our KW ($2\times$) achieves the best results, which further surpasses ODConv ($4\times$) by 0.06\% top-1 gain with roughly only half of its parameters.
}
However, when we increase KW from $2\times$ to $4\times$, it shows a decline in top-1 gain from 3.78\% to 3.29\%.
Figure~\ref{fig:training_curve} illustrates the training and validation accuracy curves of ResNet18 backbone with ODConv ($4\times$) and our KW ($1\times, 2\times, 4\times$).
We can see that KW ($2\times$) already gets higher training accuracy than ODConv ($4\times$).
While compared to KW ($2\times$), KW ($4\times$) further brings 2.79\% improvement on training set but 0.49\% drop on validation set.
We consider the reason of validation accuracy decline is that KW ($4\times$) largely enhances the capacity of ResNet18 backbone, but also suffers from potential overfitting, when using the traditional training strategy.

\textbf{Results Comparison on ResNets and ConvNeXts with the Advanced Training Strategy.}
With the traditional training strategy, our KW ($1/2\times,1\times,2\times$) have shown promising results on the ResNet18 backbone.
In order to further explore the potential of KernelWarehouse on large ConvNets with a larger value of $b$, we next adopt the training strategy proposed in ConvNeXt~\citep{CNN_ConvNeXt}, with a longer training schedule (300 epochs) and more aggressive augmentations for comparisons on the ResNet18, ResNet50 and ConvNeXt-Tiny.
From the results shown in Table~\ref{table:classification_cnn_300epoch}, we can observe:
(1) with the advanced training strategy, DY-Conv ($4\times$), ODConv ($4\times$) and our KW ($1/2\times, 1\times, 4\times$) show better results and larger top-1 gains to baseline on the ResNet18 backbone, compared to the traditional training strategy. It can be seen that the advanced training strategy successfully alleviates the overfitting of KW ($4\times$), which gets the best results, bringing an absolute top-1 gain of 5.61\%. \textit{Even with 36.45\%|65.10\% parameter reduction, KW ($1/2\times$)|KW($1/4\times$) brings 2.89\%|2.29\% top-1 accuracy gain to the baseline;}
(2) on the larger ResNet50 backbone, while the vanilla dynamic convolution method (DY-Conv) shows much lower performance gain, KW ($1/2\times, 1\times, 4\times$) still bring great performance improvement to the baseline model. With 30.99\% parameter reduction, KW ($1/2\times$) improves ResNet50 backbone by a top-1 gain of 0.86\%. Our KW ($4\times$) consistently outperforms both DY-Conv ($4\times$) and ODConv ($4\times$) by 2.05\%|0.43\% top-1 gain.
Beside ResNets, we also apply KernelWarehouse to the ConvNeXt-Tiny backbone to investigate its performance on the state-of-the-art ConvNet architecture. Our method can generalize well on ConvNeXt-Tiny, brings 0.44\% top-1 gain to the baseline model with KW ($1\times$).

\textbf{Results Comparison on MobileNets.}
We further apply KernelWarehouse to MobileNetV2 ($1.0\times$, $0.5\times$) to validate its effectiveness on lightweight ConvNet architectures.
Since the lightweight MobileNetV2 models have lower capacity compared to ResNet and ConvNeXt models, we don't use aggressive augmentations for MobileNetV2.
The results are shown in Table~\ref{table:classification_mobilenetv2}.
%
Our KernelWarehouse can strike a favorable trade-off between parameter efficiency and representation power for lightweight ConvNets as well as larger ones.
\textit{Even on the lightweight MobileNetV2 ($1.0\times, 0.5\times$) with 3.50M|1.97M parameters, our KW ($1/2\times$) can reduce the model size by 24.29\%|25.38\% while bringing top-1 gain of 0.55\%|0.89\%.} Again, our KW ($4\times$) obtains the best results on both MobileNetV2 ($1.0\times$) and MobileNet ($0.5\times$).


\begin{table}[t]
\vskip -0.15 in
\caption{Results comparison on the MS-COCO 2017 validation set. Best results are bolded.}
\vskip -0.0 in
\label{table:detection}
\begin{center}
\resizebox{.95\linewidth}{!}
{
\begin{tabular}{c|l|c|c|c|c|c|c|c|c|c|c|c|c}
\hline
\multirow{2}*{\makecell[c]{Detectors}} & \multirow{2}*{\makecell[c]{Backbone Models}} & \multicolumn{6}{c|}{Object Detection} &  \multicolumn{6}{c}{Instance Segmentation} \\
\cline{3-14}
& & $AP $ & $AP_{50} $ & $AP_{75} $ & $AP_{S} $ & $AP_{M} $ & $AP_{L} $ & $AP $ & $AP_{50} $ & $AP_{75} $ & $AP_{S} $ & $AP_{M} $ & $AP_{L} $ \\
\hline
\multirow{12}*{\makecell[c]{Mask R-CNN}} & ResNet50 & 39.6 & 61.6 & 43.3 & 24.4 & 43.7 & 50.0 & 36.4 & 58.7 & 38.6 & 20.4 & 40.4 & 48.4 \\
& + DY-Conv ($4\times$) & 39.6 & 62.1 & 43.1 & 24.7 & 43.3 & 50.5 & 36.6 & 59.1 & 38.6 & 20.9 & 40.2 & 49.1 \\
& + ODConv ($4\times$) & 42.1 & 65.1 & 46.1 & \textbf{27.2} & 46.1 & 53.9 & 38.6 & 61.6 & 41.4 & \textbf{23.1} & 42.3 & 52.0 \\
& + KW ($1\times$) & 41.8 & 64.5 & 45.9 & 26.6 & 45.5 & 53.0 & 38.4 & 61.4 & 41.2 & 22.2 & 42.0 & 51.6 \\
& + KW ($4\times$) & \textbf{42.4} & \textbf{65.4} & \textbf{46.3} & \textbf{27.2} & \textbf{46.2} & \textbf{54.6} & \textbf{38.9} & \textbf{62.0} & \textbf{41.5} & 22.7 & \textbf{42.6} & \textbf{53.1} \\
\cline{2-14}
& MobileNetV2 ($1.0\times$) & 33.8 & 55.2 & 35.8 & 19.7 & 36.5 & 44.4 & 31.7 & 52.4 & 33.3 & 16.4 & 34.4 & 43.7 \\
& + DY-Conv ($4\times$) & 37.0 & 58.6 & 40.3 & 21.9 & 40.1 & 47.9 & 34.1 & 55.7 & 36.1 & 18.6 & 37.1 & 46.3 \\
& + ODConv ($4\times$) & 37.2 & 59.4 & 39.9 & 22.6 & 40.0 & 48.0 & 34.5 & 56.4 & 36.3 & 19.3 & 37.3 & 46.8 \\
& + KW ($1\times$) & 36.4 & 58.3 & 39.2 & 22.0 & 39.6 & 47.0 & 33.7 & 55.1 & 35.7 & 18.9 & 36.7 & 45.6 \\
& + KW ($4\times$) & \textbf{38.0} & \textbf{60.0} & \textbf{40.8} & \textbf{23.1} & \textbf{40.7} & \textbf{50.0} & \textbf{34.9} & \textbf{56.6} & \textbf{37.0} & \textbf{19.4} & \textbf{37.9} & \textbf{47.8} \\
\cline{2-14}
& ConvNeXt-Tiny & 43.4 & 65.8 & 47.7 & 27.6 & 46.8 & 55.9 & 39.7 & 62.6 & 42.4 & 23.1 & 43.1 & 53.7 \\
& + KW ($1\times$) & \textbf{44.7} & \textbf{67.5} & \textbf{48.9} & \textbf{29.6} & \textbf{48.2} & \textbf{57.2} & \textbf{40.6} & \textbf{64.3} & \textbf{43.6} & \textbf{24.6} & \textbf{44.1} & \textbf{54.8} \\
\hline
\end{tabular}
}
\end{center}
\vskip -0.2 in
\end{table}

\subsection{Object Detection on MS-COCO}
To evaluate the generalization capacity of our method on different tasks, we further conduct comparative experiments for object detection and instance segmentation on the MS-COCO 2017 dataset~\citep{Dataset_MS_COCO}, which contains 118,000 training images and 5,000 validation images with 80 object categories.

\textbf{Experimental Setup.}
We adopt Mask R-CNN~\citep{DL_Instance_MaskRCNN} as the detection framework, ResNet50 and MobileNetV2 ($1.0\times$) built with different attention methods as the backbones which are pre-trained on ImageNet dataset. All the models are trained with standard $1\times$ schedule 
on MS-COCO dataset.
For a fair comparison, we adopt the same settings including data processing pipeline and hyperparameters for all the models. Experimental details are described in the Appendix.

\textbf{Results Comparison.}
The comparison results on Mask R-CNN with different backbones are shown in Table~\ref{table:detection}.
For Mask R-CNN with ResNet50 backbone models, we observe a similar trend to the main experiments on ImageNet dataset: KW ($4\times$) outperforms DY-Conv ($4\times$) and ODConv ($4\times$) on both object detection and instance segmentation tasks.
Our KW ($1\times$) brings an AP improvement of 2.2\%|2.0\% on object detection and instance segmentation tasks, which is on par with ODConv ($4\times$).
With the MobileNetV2 ($1.0\times$) backbone, KernelWarehouse yields consistent high accuracy improvements to the baseline. Our KW ($4\times$) achieves the best results.
With the ConvNeXt-Tiny backbone, KW ($1\times$) brings more obvious performance gain to the baseline model on MS-COCO dataset compared to that on ImageNet dataset, showing its higher capacity and good generalization ability to the downstream tasks.

\subsection{Ablation Studies}
For a deeper understanding of KernelWarehouse, we further provide a lot of ablative experiments on the ImageNet dataset. All the models are trained with the training strategy proposed in ConvNeXt~\citep{CNN_ConvNeXt}.
\begin{table}[t]
\vskip -0.1 in
\begin{minipage}[t]{0.49\linewidth}
\caption{Ablation of KernelWarehouse with warehouse sharing between kernel cells within each stage, within each layer and without sharing.}
\vskip -0.0 in
\label{table:ablation_sharingscale}
\begin{center}
\resizebox{1.0\linewidth}{!}{
\begin{tabular}{l|c|c|c|c}
\hline
Models & Sharing Strategies & Params & Top-1 Acc (\%) & Top-5 Acc (\%)\\
\hline
ResNet18 & - & 11.69M & 70.44 & 89.72 \\
\hline
\multirow{3}{*}{+ KW ($1\times$)} & Within each stage & 11.93M & \textbf{74.77} ($\uparrow$\textbf{4.33}) & \textbf{92.13} ($\uparrow$\textbf{2.41}) \\
 & Within each layer & 11.81M & 74.34 ($\uparrow$3.90) & 91.82 ($\uparrow$2.10) \\
 & Without sharing & 11.78M & 72.49 ($\uparrow$2.05) & 90.81 ($\uparrow$1.09) \\
\hline

\end{tabular}
}
\end{center}
\vskip -0.1 in
\end{minipage}
\hfill
\begin{minipage}[t]{0.49\linewidth}

\caption{Ablation of KernelWarehouse with or without warehouse sharing between kernels having different dimensions in convolutional blocks.}
\vskip -0.0 in
\label{table:ablation_sharingshape}
\begin{center}
\resizebox{1.0\linewidth}{!}{
\begin{tabular}{l|c|c|c|c}
\hline
Models & Sharing Strategies & Params & Top-1 Acc (\%) & Top-5 Acc (\%)\\
\hline
ResNet50 & - & 25.56M & 78.44 & 94.24 \\
\hline
\multirow{2}{*}{+ KW ($1\times$)} & With different dimensions & 28.05M & \textbf{80.38} ($\uparrow$\textbf{1.94}) & \textbf{95.27} ($\uparrow$\textbf{1.03}) \\
& Only with the same dimensions & 26.95M & 79.80 ($\uparrow$1.36) & 95.01 ($\uparrow$0.77) \\
\hline

\end{tabular}
}
\end{center}
\end{minipage}
\vskip -0.2 in
\begin{minipage}[b]{0.49\linewidth}
\caption{Ablation of KernelWarehouse with or without kernel partition.}
\vskip -0.0 in
\label{table:ablation_partition}
\begin{center}
\resizebox{1.0\linewidth}{!}{
\begin{tabular}{l|c|c|c|c}
\hline
Models & Kernel Partition & Params & Top-1 Acc (\%) & Top-5 Acc (\%)\\
\hline
ResNet18 & - & 11.69M & 70.44 & 89.72 \\
\hline
\multirow{2}{*}{+ KW ($1\times$)} & \checkmark & 11.93M & \textbf{74.77} ($\uparrow$\textbf{4.33}) & \textbf{92.13} ($\uparrow$\textbf{2.41}) \\
 & - & 11.78M & 70.49 ($\uparrow$0.05) & 89.84 ($\uparrow$0.12) \\
\hline
\end{tabular}
}
\end{center}
\vskip -0.2 in
\end{minipage}
\hfill
\begin{minipage}[b]{0.49\linewidth}
\caption{Ablation of KernelWarehouse with different attention functions.}
\vskip -0.0 in
\label{table:ablation_function}
\begin{center}
\resizebox{1.0\linewidth}{!}{
\begin{tabular}{l|c|c|c|c}
\hline
Models & Attention Functions & Params & Top-1 Acc (\%) & Top-5 Acc (\%)\\
\hline
ResNet18 & - & 11.69M & 70.44 & 89.72 \\
\hline
\multirow{4}{*}{+ KW ($1\times$)}
& $z_{ij}/\sum^{n}_{p=1}{|z_{ip}|}$ & 11.93M & \textbf{74.77} ($\uparrow$\textbf{4.33}) & \textbf{92.13} ($\uparrow$\textbf{2.41}) \\
& Softmax & 11.93M & 72.67 ($\uparrow$2.23) & 90.82 ($\uparrow$1.10) \\
& Sigmoid & 11.93M & 72.09 ($\uparrow$1.65) & 90.70 ($\uparrow$0.98) \\
& $max(z_{ij},0)/\sum^{n}_{p=1}{|z_{ip}|}$ & 11.93M & 72.74 ($\uparrow$2.30) & 90.86 ($\uparrow$1.14) \\
\hline
\end{tabular}
}
\end{center}
\vskip -0.1 in
\end{minipage}
\vskip -0.1 in
\end{table}

\begin{table}[t]
\vskip -0.1 in
\caption{Ablation of KernelWarehouse with or without our attentions initialization strategy.}
\vskip -0.0 in
\label{table:ablation_temperature}
\begin{center}
\resizebox{.6\linewidth}{!}{
\begin{tabular}{l|c|c|c|c}
\hline
Models & Attentions Initialization Strategy & Params & Top-1 Acc (\%) & Top-5 Acc (\%)\\
\hline
ResNet18 & - & 11.69M & 70.44 & 89.72 \\
\hline
\multirow{2}{*}{+ KW ($1\times$)} & \checkmark & 11.93M & \textbf{74.77} ($\uparrow$\textbf{4.33}) & \textbf{92.13} ($\uparrow$\textbf{2.41}) \\
 & - & 11.93M & 73.39 ($\uparrow$2.95) & 91.24 ($\uparrow$1.52) \\
\hline
\end{tabular}
}
\end{center}
\vskip -0.15 in
\end{table}

\textbf{Warehouse Sharing with Different Ranges.}
To validate the effectiveness of warehouse sharing,
we first perform ablative experiments on the ResNet18 backbone with different ranges of warehouse sharing.
From the results shown in Table~\ref{table:ablation_sharingscale}, 
we can see that when sharing warehouse in wider range, KernelWarehouse brings larger performance improvement to the baseline model. It indicates that explicitly enhancing parameter dependencies within the same convolutional layer and across successive layers both can strengthen the network capacity.


\textbf{Warehouse Sharing between Kernels with Different Dimensions.}
In the mainstream ConvNet designs, a convolutional block mostly contains several kernels having different dimensions ($k\times k\times c\times f$).
We next perform ablative experiments on the ResNet50 backbone to explore the effect of warehouse sharing between kernels having different dimensions in convolutional blocks.
Results are summarized in Table~\ref{table:ablation_sharingshape}, showing that warehouse sharing between kernels having different dimensions performs better compared to warehouse sharing only between kernels having the same dimensions.
Combining the results in Table~\ref{table:ablation_sharingscale} and Table~\ref{table:ablation_sharingshape}, we can conclude that enhancing the warehouse sharing between more kernel cells in KernelWarehouse mostly leads to better performance.
%

\textbf{Effect of Kernel Partition.} Using kernel partition, KernelWarehouse can apply denser kernel assembling with more kernel cells. In Table~\ref{table:ablation_partition}, we present the comparison results on the ResNet18 backbone about kernel partition. Without kernel partition, the top-1 gain for KernelWarehouse to the baseline sharply decreases from 4.33\% to 0.05\%, demonstrating its great importance to KernelWarehouse.

\textbf{Attention Function.}
Recall that KernelWarehouse uses a new attention function. 
In the experiments, we compare the performance of KernelWarehouse with different attention functions on the ResNet18 backbone. As shown in Table~\ref{table:ablation_function}, the attention function plays an important role in KernelWarehouse, where the performance gap between our design $z_{ij}/\sum^{n}_{p=1}{|z_{ip}|}$ and Softmax|Sigmoid reaches 2.10\%|2.68\%. Our design also outperforms the function $max(z_{ij},0)/\sum^{n}_{p=1}{|z_{ip}|}$ by 2.03\%, validating the importance of introducing negative values for scalar attentions to encourage the network to learn adversarial attention relationships.
%

\textbf{Attentions Initialization Strategy.}
To help the optimization of KernelWarehouse in the early training stage, $\beta_{ij}$ with temperature $\gamma$ is used for initializing the scalar attentions. In the experiments, we use ResNet18 as the backbone to study the effect of our attentions initialization strategy.
As shown in Table~\ref{table:ablation_temperature}, a proper initialization strategy for scalar attentions is beneficial for a network to learn relationships between linear mixtures and kernel cells, which leads to 1.38\% top-1 improvement to the ResNet18 backbone based on KW ($1\times$).

\begin{figure}[t!]
    \begin{subfigure}[b]{0.245\linewidth}
        \begin{center}
            \includegraphics[width=\textwidth]{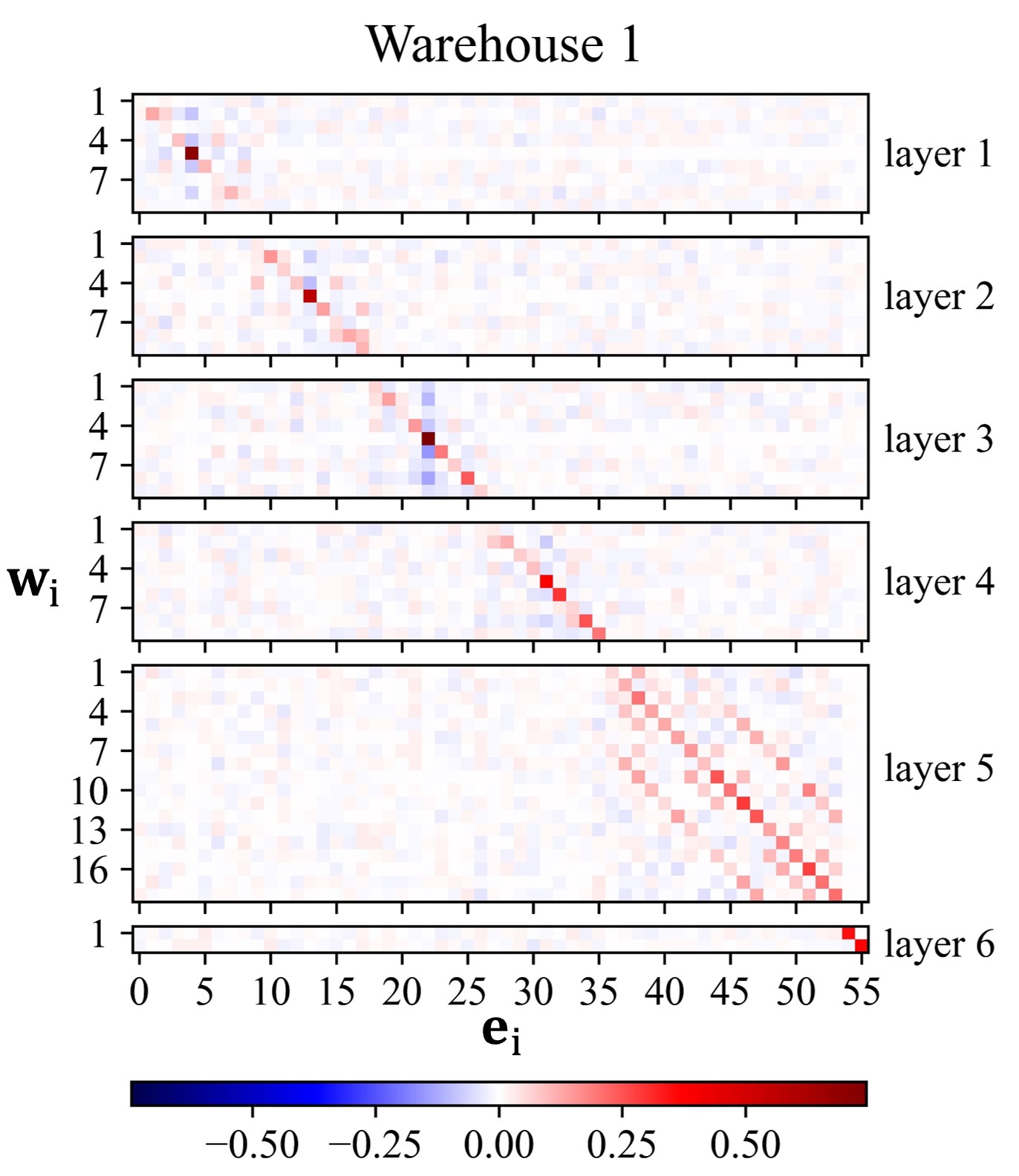}
        \end{center}
    \end{subfigure}
    \hfill
    \begin{subfigure}[b]{0.245\linewidth}
        \begin{center}
            \includegraphics[width=\textwidth]{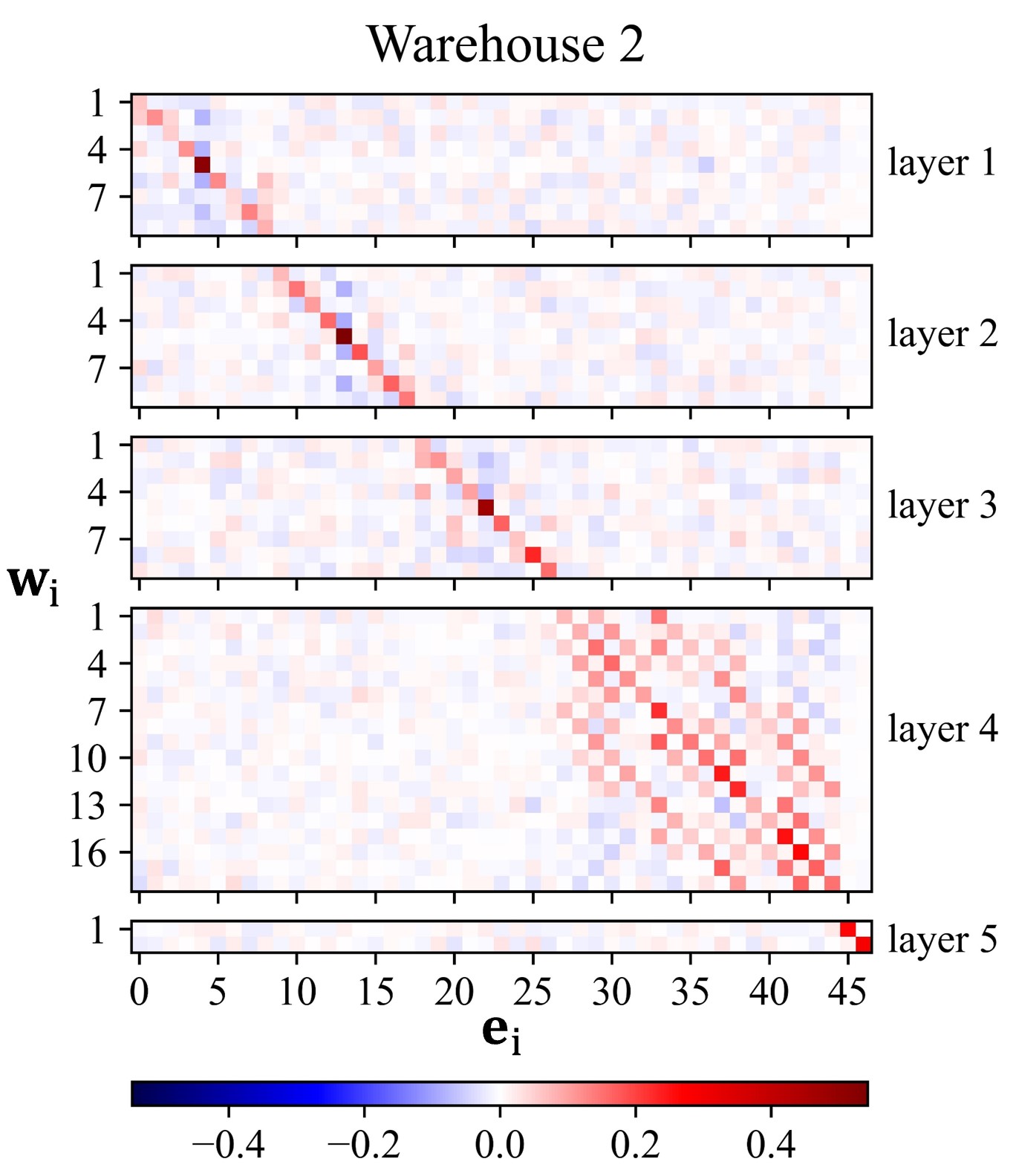}
        \end{center}
    \end{subfigure}
     \hfill
    \begin{subfigure}[b]{0.245\linewidth}
        \begin{center}
            \includegraphics[width=\textwidth]{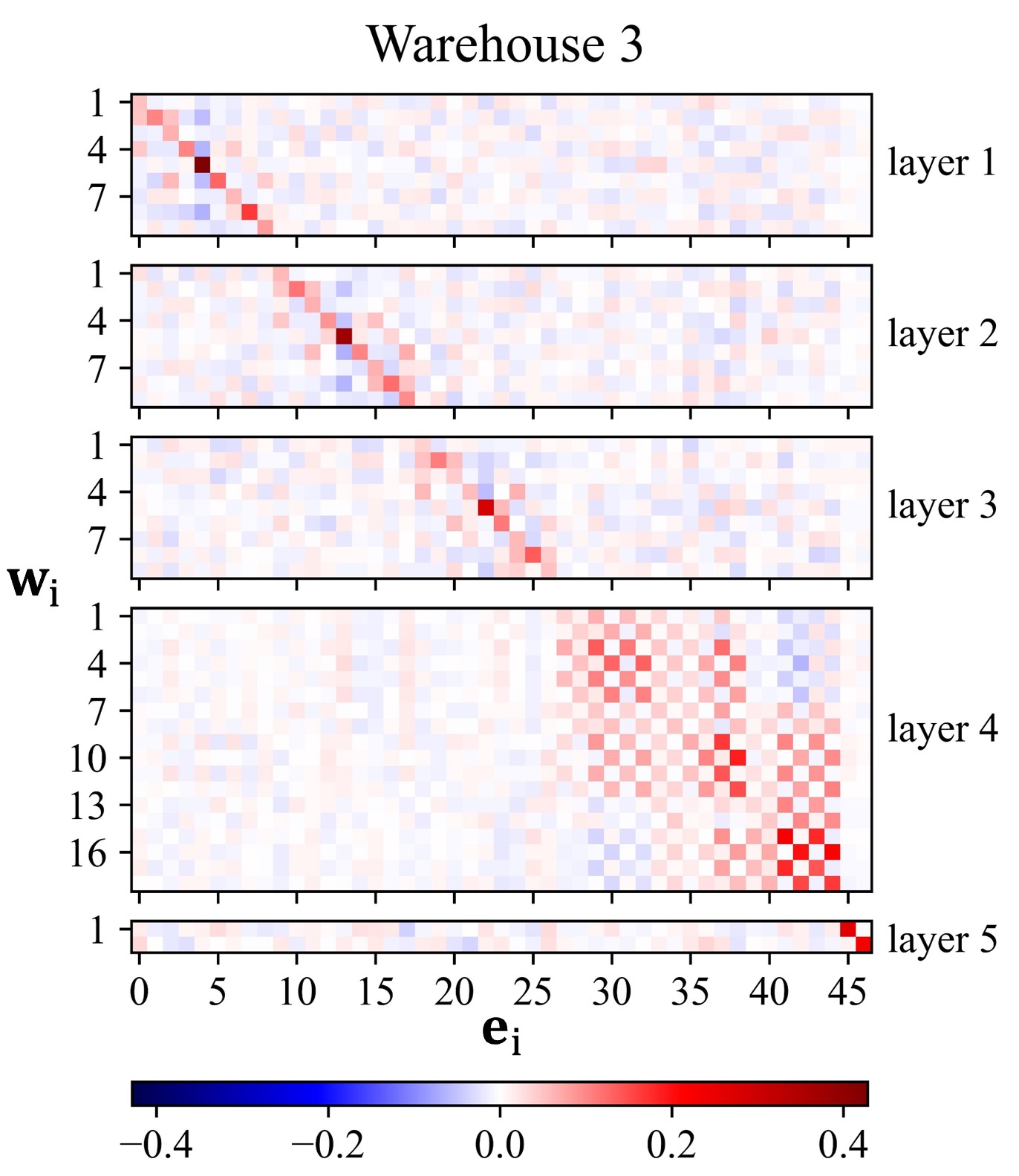}
        \end{center}
    \end{subfigure}
    \hfill
    \begin{subfigure}[b]{0.245\linewidth}
        \begin{center}
            \includegraphics[width=\textwidth]{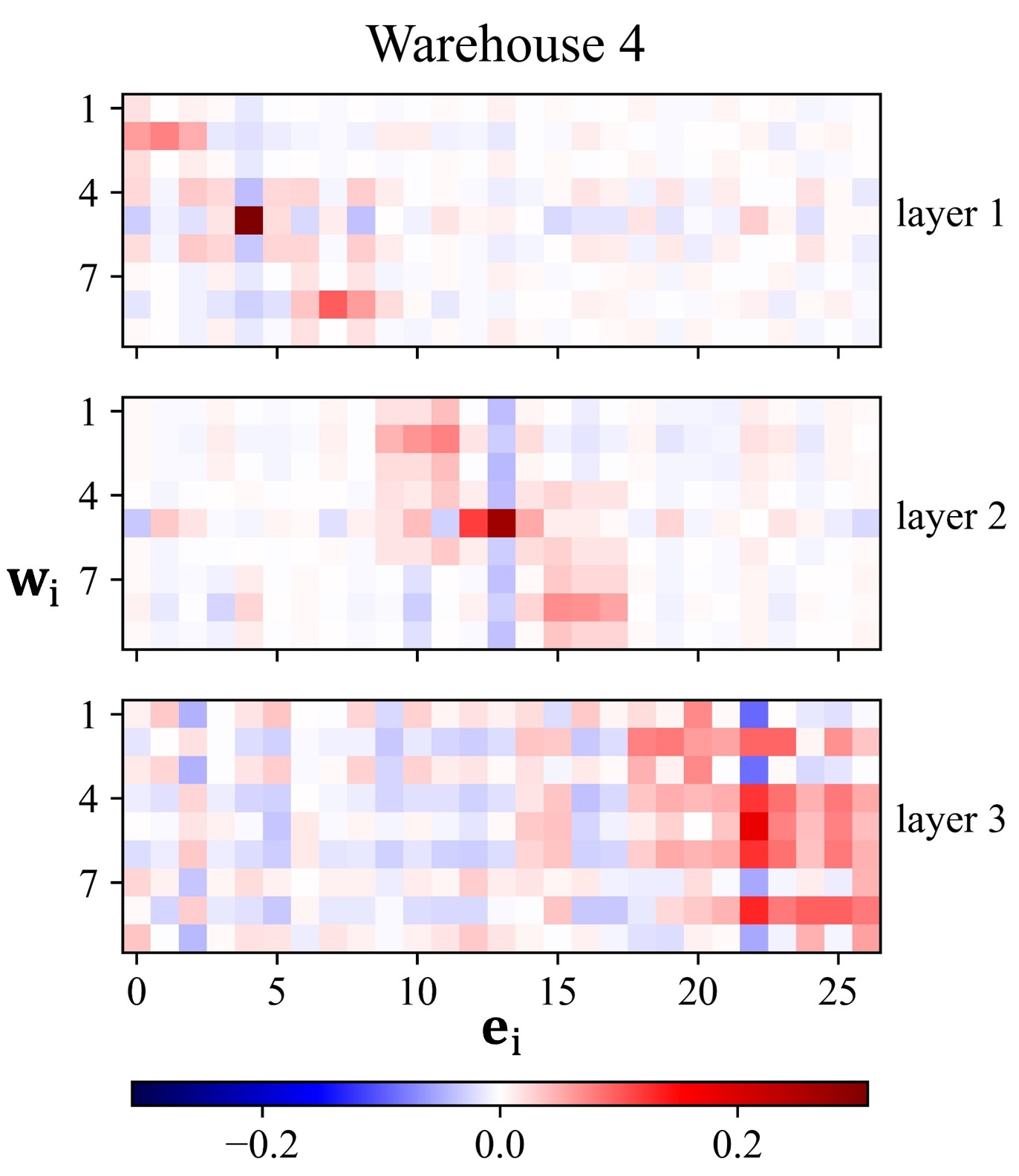}
        \end{center}
    \end{subfigure}

    \vskip -0.00in
    \caption{Visualization of statistical mean values of learnt attention $\alpha_{ij}$ in each warehouse. The results are obtained from the pre-trained ResNet18 backbone with KW ($1\times$) for all of the 50,000 images on the ImageNet validation dataset. Best viewed with zoom-in.
    }

    \label{figure:attention_resnet18_heatmap}
    \vskip -0.17 in
\end{figure}

\textbf{Visualization.}
To have a better understanding of the warehouse sharing mechanism of KernelWarehouse, we further analyze the statistical mean values of $\alpha_{ij}$ to study its learnt attention values. The results are obtained from the pre-trained ResNet18 backbone with KW ($1\times$). The visualization results are shown in Figure~\ref{figure:attention_resnet18_heatmap}, from which we can observe:
(1) each linear mixture can learn its own distribution of scalar attentions for different kernel cells;
(2) in each warehouse, the maximum value of $\alpha_{ij}$ in each row mostly appears in the diagonal line throughout the whole warehouse. It indicates that our attentions initialization strategy can help KernelWarehouse build one-to-one relationship between linear mixtures and kernel cells according to our setting of $\beta_{ij}$;
(3) compared to linear mixtures in different layers, the attentions $\alpha_{ij}$ with higher absolute values for linear mixtures in the same layer have more overlap. It indicates that parameter dependencies within the same kernel are stronger than that across layers, which can be learned by KernelWarehouse.

%

\begin{table}[h]
\centering
\vskip -0.2 in
\caption{Results comparison of inference speed (frames per second) on the ResNet50 and MobileNetV2 ($1.0\times$) with different dynamic convolution methods. All the models are tested on an NVIDIA TITAN X GPU (with batch size 100) and a single core of Intel E5-2683 v3 CPU (with batch size 1), separately. The input image size is 224$\times$224.}
\vskip 0.1 in
\label{table:ablation_speed}
\begin{minipage}[b]{0.48\linewidth}
\centering
\begin{center}
\resizebox{1.0\linewidth}{!}{
\begin{tabular}{l|c|c|c|c}
\hline
Models & Params & Top-1 Acc (\%) & Speed on GPU & Speed on CPU \\
\hline
ResNet50 & 25.56M & 78.44 & 647.0 & 6.4 \\
+ DY-Conv ($4\times$) & 100.88M & 79.00 ($\uparrow$0.56) & 322.7 & 4.1 \\
+ ODConv ($4\times$) & 90.67M & 80.62 ($\uparrow$2.18) &142.3 & 2.5 \\
+ KW ($1/2\times$) & 17.64M & 79.30 ($\uparrow$0.86) & 201.0 & 1.5 \\
+ KW ($1\times$) & 28.05M & 80.38 ($\uparrow$1.94) & 246.1 & 1.6 \\
+ KW ($4\times$) & 102.02M & 81.05 ($\uparrow$2.61) & 178.5 & 0.6 \\
\hline
\end{tabular}
}
\end{center}
\vskip -0.0 in
\hfill
\end{minipage}
\hfill
\begin{minipage}[b]{0.49\linewidth}
\centering
\vskip -0.0 in
\begin{center}
\resizebox{1.0\linewidth}{!}{
\begin{tabular}{l|c|c|c|c}
\hline
Models & Params & Top-1 Acc (\%) & Speed on GPU & Speed on CPU \\
\hline
MobileNetV2 ($1.0\times$) & 3.50M & 72.02 & 1410.8 & 17.0 \\
+ DY-Conv ($4\times$) & 12.40M & 74.94 ($\uparrow$2.92) & 862.4 & 11.8\\
+ ODConv ($4\times$) & 11.52M & 75.42 ($\uparrow$3.40) & 536.5 & 11.0 \\
+ KW ($1/2\times$) & 2.65M & 72.57 ($\uparrow$0.55) & 825.9 & 11.6 \\
+ KW ($1\times$) & 5.17M & 74.68 ($\uparrow$2.66) & 575.3 & 10.7 \\
+ KW ($4\times$) & 11.38M & 75.92 ($\uparrow$3.90) & 567.2 & 8.4 \\
\hline
\end{tabular}
}
\end{center}
\hfill
\end{minipage}
\vskip -0.15 in
\end{table}

\textbf{Limitations.} 
Table~\ref{table:ablation_speed} provides two sets of experiments for inference speed analysis, from which we can observe that the runtime speed of the models trained with KernelWarehouse is usually slower than its dynamic convolution counterparts under the similar model size budget, especially for large ConvNets. The main reason for this limitation is the dense computation of linear mixtures in KernelWarehouse. Besides, we believe KernelWarehouse could be applied to more deep and large ConvNets, yet we are unable to explore this constrained by our computational resources.

\section{Conclusion}

In this paper, we present KernelWarehouse, a more general form of dynamic convolution, which could be used to improve the performance of modern ConvNets. Experiments on ImageNet and MS-COCO datasets show its great potential. We hope our work would inspire future research in dynamic convolution.

\section{Acknowledgement}
We thank Intel Data Center \& AI group's great support of their DGX-2 and DGX-A100 servers for training large models in this project.

\bibliography{neurips_2023}
\bibliographystyle{unsrt}

\clearpage

\appendix
\section{Appendix}

\subsection{Datasets and Implementation Details}

\subsubsection{Image Classification on ImageNet}

Recall that we use ResNet~\citep{CNN_ResNet}, MobileNetV2~\citep{CNN_MobileNetV2} and ConvNeXt~\citep{CNN_ConvNeXt} families for the main experiments on ImageNet dataset~\citep{Dataset_ImageNet}, which consists of over 1.2 million training images and 50,000 validation images with 1,000 object categories.
We use an input image resolution of 224$\times$224 for both training and evaluation. All the input images are standardized with mean and standard deviation per channel.
For evaluation, we report top-1 and top-5 recognition rates of a single 224$\times$224 center crop on the ImageNet validation set.
All the experiments are performed on the servers having 8 GPUs. Specifically, the models of ResNet18, MobileNetV2 ($1.0\times$), MobileNetV2 ($0.5\times$) are trained on the servers with 8 NVIDIA Titan X GPUs. The models of ResNet50, ConvNeXt-Tiny are trained on the servers with 8 NVIDIA Tesla V100-SXM3 or A100 GPUs. The training setups for different models are as follows.

\textbf{Training setup for ResNet models with the traditional training strategy.}
All the models are trained by the stochastic gradient descent (SGD) optimizer for 100 epochs, with a batch size of 256, a momentum of 0.9 and a weight decay of 0.0001. The initial learning rate is set to 0.1 and decayed by a factor of 10 for every 30 epoch. Horizontal flipping and random resized cropping are used for data augmentation. For KernelWarehouse, the temperature $\tau$ linearly reduces from 1 to 0 in the first 10 epochs.

\textbf{Training setup for ResNet and ConvNeXt models with the advanced training strategy.}
Following the settings of ConvNeXt~\citep{CNN_ConvNeXt}, all the models are trained by the AdamW optimizer with $\beta_{1}=0.9, \beta_{2}=0.999$ for 300 epochs, with a batch size of 4096, a momentum of 0.9 and a weight decay of 0.05. The initial learning rate is set to 0.004 and annealed down to zero following a cosine schedule. Randaugment~\citep{augmentation_randaugment}, mixup~\citep{augmentation_mixup}, cutmix~\citep{augmentation_cutmix}, random erasing~\citep{augmentation_randerazing} and label smoothing~\citep{augmentation_labelsmoothing} are used for augmentation. For KernelWarehouse, the temperature $\tau$ linearly reduces from 1 to 0 in the first 20 epochs.

\textbf{Training setup for MobileNetV2 models.}
All the models are trained by the SGD optimizer for 150 epochs, with a batch size of 256, a momentum of 0.9 and a weight decay of 0.00004. The initial learning rate is set to 0.1 and annealed down to zero following a cosine schedule. Horizontal flipping and random resized cropping are used for data augmentation.
For KernelWarehouse, the temperature $\tau$ linearly reduces from 1 to 0 in the first 10 epochs.

\subsubsection{Object Detection and Semantic Segmentation on MS-COCO}

Recall that we conduct comparative experiments for object detection and instance segmentation on the MS-COCO 2017 dataset~\citep{Dataset_MS_COCO}, which contains 118,000 training images and 5,000 validation images with 80 object categories.
We adopt Mask R-CNN as the detection framework, ResNet50 and MobileNetV2 (1.0$\times$) built with different dynamic convolution methods as the backbones which are pre-trained on ImageNet dataset.
All the models are trained with a batch size of 16 and standard 1$\times$ schedule on the MS-COCO dataset using multi-scale training. The learning rate is decreased by a factor of 10 at the 8$^{th}$ and the 11$^{th}$ epoch of total 12 epochs.
For a fair comparison, we adopt the same settings including data processing pipeline and hyperparameters for all the models. All the experiments are performed on the servers with 8 NVIDIA Tesla V100 GPUs.
The attentions initialization strategy is not used for KernelWarehouse during fine-tuning to avoid disrupting the learnt relationships of the pre-trained models between kernel cells and linear mixtures.
For evaluation, we report both bounding box Average Precision (AP) and mask AP on the MS-COCO 2017 validation set, including $AP_{50}$, $AP_{75}$ (AP at different IoU thresholds) and $AP_{S}$, $AP_{M}$, $AP_{L}$ (AP at different scales).

\subsection{Visualization Examples of Attentions Initialization Strategy}

\begin{figure}[th]
    \begin{center}
        \includegraphics[width=\linewidth]{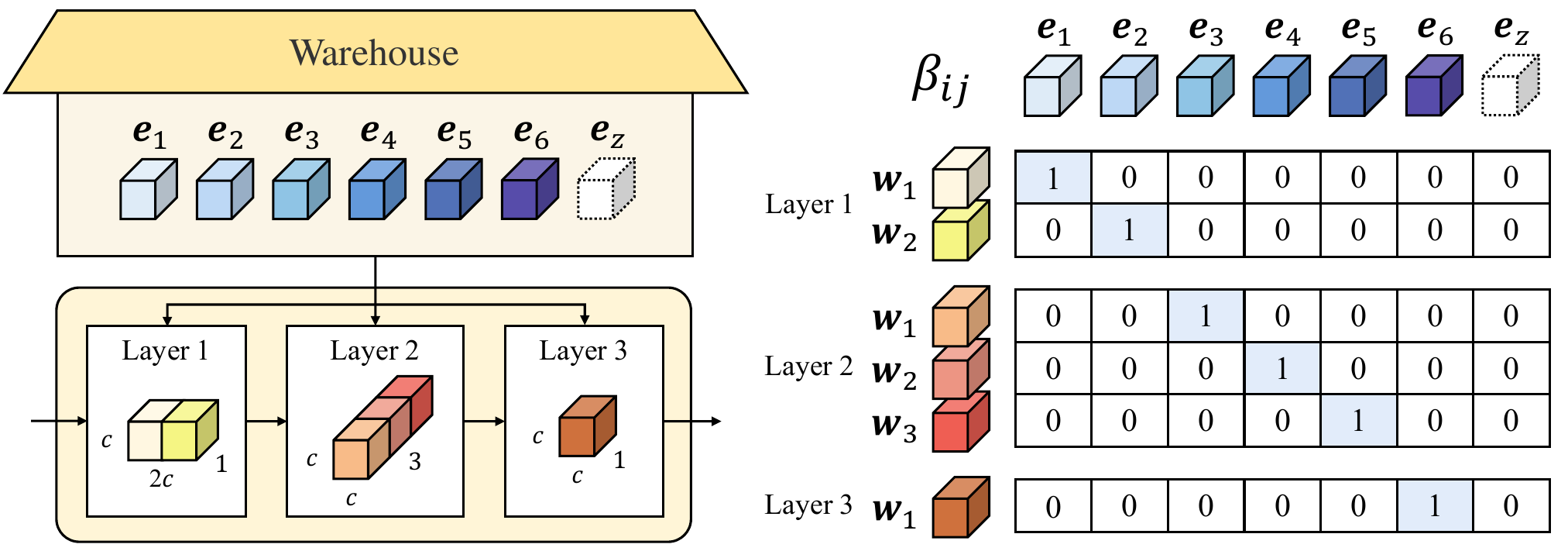}
    \end{center}
    \caption{A visualization example of attentions initialization strategy for KW ($1\times$), where both $n$ and $m_{t}$ equal to 6.
    It helps the ConvNet to build one-to-one relationships between kernel cells and linear mixtures in the early training stage according to our setting of $\beta_{ij}$.
    $\mathbf{e}_{z}$ is a kernel cell that doesn't really exist and it keeps as a zero matrix constantly. In the beginning of the training process when temperature $\tau$ is 1, a ConvNet built with KW ($1\times$) can be roughly seen as a ConvNet with standard convolutions.
    }
    \label{fig:attention_initialization_1x}
\end{figure}

\begin{figure}[th]
    \begin{minipage}[t]{0.49\linewidth}
        \begin{center}
            \includegraphics[width=\textwidth]{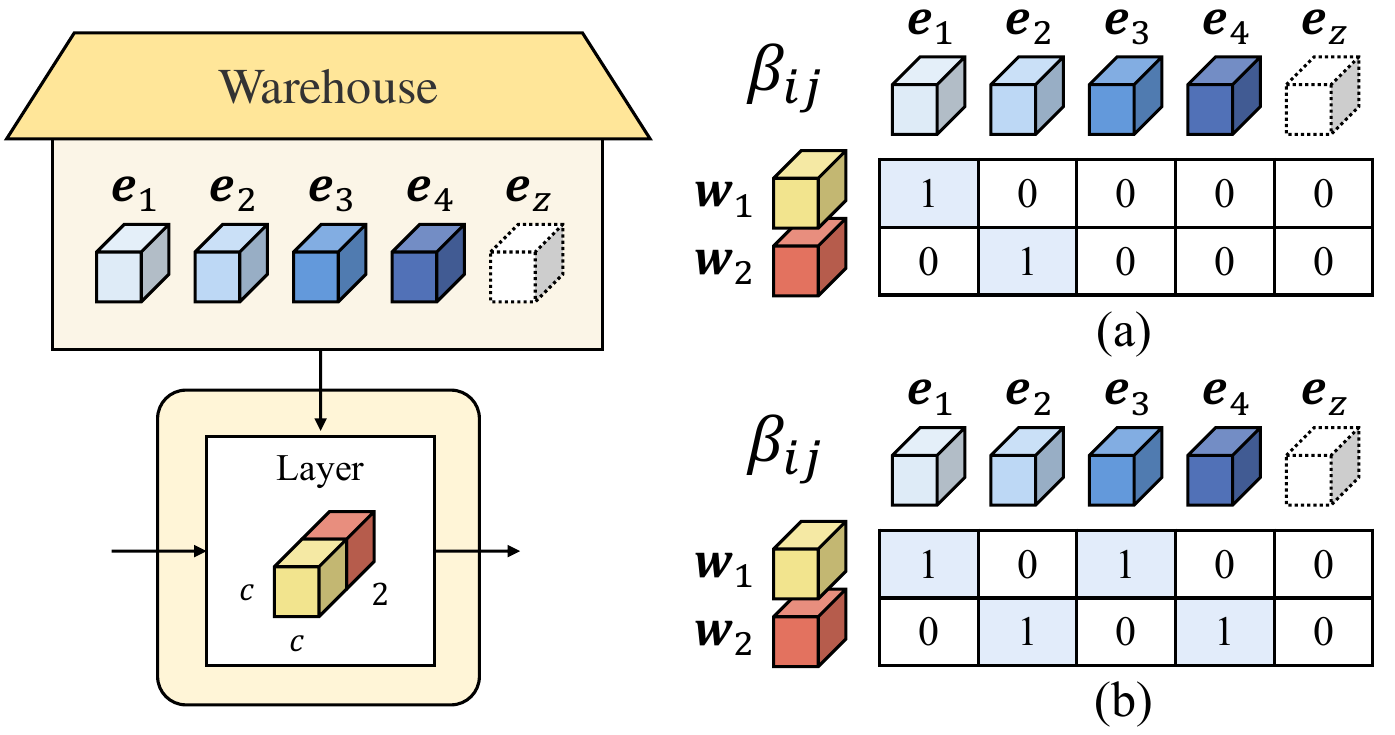}
        \end{center}
        \vskip -0.1in
        \caption{Visualization examples of attentions initialization strategies for KW ($2\times$), where $n=4$ and $m_{t}=2$. (a) our proposed strategy builds one-to-one relationships between kernel cells and linear mixtures; (b) an alternative strategy which builds two-to-one relationships between kernel cells and linear mixtures.}
        \label{fig:attention_initialization_2x}
    \end{minipage}
    \hfill
    \begin{minipage}[t]{0.49\linewidth}
        \begin{center}
            \includegraphics[width=\textwidth]{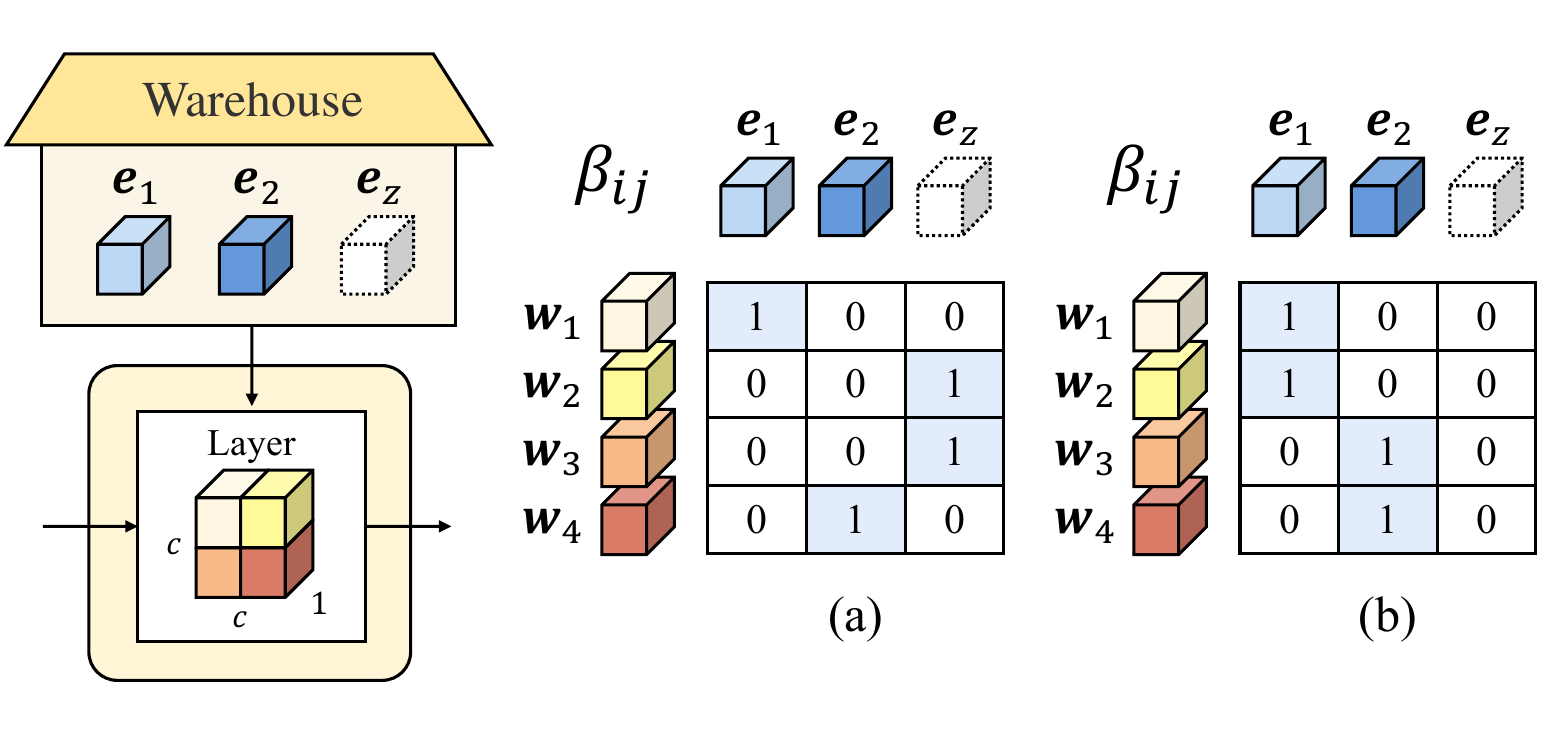}
        \end{center}
        \vskip -0.1in
        \caption{Visualization examples of attentions initialization strategies for KW ($1/2\times$), where $n=2$ and $m_{t}=4$. (a) our proposed strategy builds one-to-one relationships between kernel cells and linear mixtures; (b) an alternative strategy which builds one-to-two relationships between kernel cells and linear mixtures.}
        \label{fig:attention_initialization_1d2x}
    \end{minipage}

    \vskip -0.1in
\end{figure}

\begin{table}[tbh]
\vskip -0.15 in
\caption{Ablation of KernelWarehouse with different attentions initialization strategies.}
\vskip -0.0 in
\label{table:ablation_initialization}
\begin{center}
\resizebox{.85\linewidth}{!}{
\begin{tabular}{l|l|c|c|c}
\hline
Models & Attentions Initialization Strategies & Params & Top-1 Acc (\%) & Top-5 Acc (\%)\\
\hline
ResNet18 & - & 11.69M & 70.44 & 89.72 \\
\hline
\multirow{3}{*}{+ KW ($1\times$)} & 1 kernel cell to 1 linear mixture & 11.93M & \textbf{74.77} ($\uparrow$\textbf{4.33}) & \textbf{92.13} ($\uparrow$\textbf{2.41}) \\
 & all the kernel cells to 1 linear mixture & 11.93M & 73.36 ($\uparrow$2.92) & 91.41 ($\uparrow$1.69) \\
 & without attentions initialization & 11.93M & 73.39 ($\uparrow$2.95) & 91.24 ($\uparrow$1.52) \\
\hline
\multirow{2}{*}{+ KW ($4\times$)} &  1 kernel cell to 1 linear mixture & 45.86M & \textbf{76.05} ($\uparrow$\textbf{5.61}) & \textbf{92.68} ($\uparrow$\textbf{2.96}) \\
 &  4 kernel cells to 1 linear mixture & 45.86M & 76.03 ($\uparrow$5.59) & 92.53 ($\uparrow$2.81) \\
 \hline
 \multirow{2}{*}{+ KW ($1/2\times$)} &  1 kernel cell to 1 linear mixture & 7.43M & \textbf{73.33} ($\uparrow$\textbf{2.89}) & \textbf{91.42} ($\uparrow$\textbf{1.70}) \\
 &  1 kernel cell to 2 linear mixtures & 7.43M & 72.89 ($\uparrow$2.45) & 91.34 ($\uparrow$1.62) \\
 \hline
\end{tabular}
}
\end{center}
\vskip -0.1 in
\end{table}

Recall that we adopt an attentions initialization strategy for KernelWarehouse using $\tau$ and $\beta_{ij}$. It forces the scalar attentions to be one-hot in the early training stage for building one-to-one relationships between kernel cells and linear mixtures. To give a better understanding of this strategy, we provide visualization examples for KW ($1\times$), KW ($2\times$) and KW ($1/2\times$), respectively.
We also provide a set of ablative experiments to compare our proposed strategy with other alternatives.

\textbf{Attentions Initialization for KW ($1\times$).}
A visualization example of attentions initialization strategy for KW ($1\times$) is shown in Figure~\ref{fig:attention_initialization_1x}.
In this example, a warehouse $\mathbf{E}=\{\mathbf{e}_{1},\dots,\mathbf{e}_{6},\mathbf{e}_{z}\}$ is shared to 3 neighboring convolutional layers with kernel dimensions of $1\times 1 \times 2c \times c$, $1\times 3 \times c \times c$ and $1\times 1 \times c \times c$, respectively. The kernel dimensions are selected for simple illustration. The kernel cells have the same dimensions of $1\times 1 \times c \times c$.
Note that the kernel cell $\mathbf{e}_{z}$ doesn't really exist and it keeps as a zero matrix constantly. It is only used for attentions normalization but not assembling kernels.
This kernel is mainly designed for attentions initialization when $b<1$ and not counted in the number of kernel cells $n$.
In the early training stage, we adopt a strategy to explicitly force every linear mixture to build relationship with one specified kernel cell according to our setting of $\beta_{ij}$.
As shown in Figure~\ref{fig:attention_initialization_1x}, we assign one of $\mathbf{e}_{1},\dots,\mathbf{e}_{6}$ in the warehouse to each of the 6 linear mixtures at the 3 convolutional layers without repetition.
So that in the beginning of the training process when temperature $\tau$ is 1, a ConvNet built with KW ($1\times$) can be roughly seen as a ConvNet with standard convolutions.
The results of Table 9 in the main manuscript validate the effectiveness of our proposed attentions initialization strategy. Here, we compare it with another alternative. In this alternative strategy, we force every linear mixture to build relationships with all the kernel cells equally by setting all the $\beta_{ij}$ to be 1. The results are shown in Table~\ref{table:ablation_initialization}.
The all-to-one strategy demonstrates similar performance with KernelWarehouse without using any attentions initialization strategy, while our proposed strategy outperforms it by 1.41\% top-1 gain.

\textbf{Attentions Initialization for KW ($2\times$).}
For KernelWarehouse with $b>1$, we adopt the same strategy for initializing attentions used in KW ($1\times$). Figure~\ref{fig:attention_initialization_2x}(a) provides a visualization example of attentions initialization strategy for KW ($2\times$). For building one-to-one relationships, we assign $\mathbf{e}_{1}$ to $\mathbf{w}_{1}$ and $\mathbf{e}_{2}$ to $\mathbf{w}_{2}$, respectively.
When $b>1$, another reasonable strategy is to assign multiple kernel cells to every linear mixture without repetition, which is shown in Figure~\ref{fig:attention_initialization_2x}(b). We use the ResNet18 backbone based on KW ($4\times$) to compare the two strategies. From the results in Table~\ref{table:ablation_initialization}, we can see that our one-to-one strategy performs better.

\textbf{Attentions Initialization for KW ($1/2\times$).}
For KernelWarehouse with $b<1$, the number of kernel cells is less than that of linear mixtures, meaning that we cannot adopt the same strategy used for $b\geq1$.
Therefore, we only assign one of the total $n$ kernel cells in the warehouse to $n$ linear mixtures respectively without repetition. And we assign $\mathbf{e}_{z}$ to all of the remaining linear mixtures. The visualization example for KW ($1/2\times$) is shown in Figure~\ref{fig:attention_initialization_1d2x}(a).
When temperature $\tau$ is 1, a ConvNet built with KW ($1/2\times$) can be roughly seen as a ConvNet with group convolutions (groups=2).
We also provide comparison results between our proposed strategy and another alternative strategy which assigns one of the $n$ kernel cells to every 2 linear mixtures without repetition.
As shown in Table~\ref{table:ablation_initialization}, our one-to-one strategy achieves better result again, showing that introducing an extra kernel $\mathbf{e}_{z}$ for $b<1$ can help the ConvNet learn more appropriate relationships between kernel cells and linear mixtures.
When assigning one kernel cell to multiple linear mixtures, a ConvNet could not balance the relationships between them well.

\subsection{Design Details of KernelWarehouse.}

In this section, we describe the design details of our KernelWarehouse.
The corresponding values of $m$ and $n$ for each of our trained models are provided in the Table~\ref{table:m_and_n}.
Note that the values of $m$ and $n$ are naturally determined according to our setting of the dimensions of the kernel cells, the layers to share warehouses and $b$.

\begin{table}[tbh]
\vskip -0.15 in
\caption{The values of $m$ and $n$ for the ResNet18, ResNet50, ConvNeXt-Tiny, MobileNetV2 (1.0$\times$) and MobileNetV2 (0.5$\times$) backbones based on KernelWarehouse.}
\vskip -0.0 in
\label{table:m_and_n}
\begin{center}
\resizebox{.95\linewidth}{!}{
\begin{tabular}{c|c|l|l}
\hline
Backbones & $b$ & m & n \\
\hline
\multirow{5}{*}{ResNet18} & 1/4 & 224, 188, 188, 108 & 56, 47, 47, 27 \\
& 1/2 & 224, 188, 188, 108 & 112, 94, 94, 54 \\
& 1 & 56, 47, 47, 27 & 56, 47, 47, 27 \\
& 2 & 56, 47, 47, 27 & 112, 94, 94, 54 \\
& 4 & 56, 47, 47, 27 & 224, 188, 188, 108 \\
\hline
\multirow{3}{*}{ResNet50} & 1/2 & 348, 416, 552, 188 & 174, 208, 276, 94 \\
& 1 & 87, 104, 138, 47 & 87, 104, 138, 47 \\
& 4 & 87, 104, 138, 47 & 348, 416, 552, 188 \\
\hline
ConvNeXt-Tiny & 1 & 16,4,4,4,147,24,147,24,441,72,147,24
 & 16,4,4,4,147,24,147,24,441,72,147,24 \\
\hline
\multirow{3}{*}{
\makecell[l]{MobileNetV2 (1.0$\times$) \\
MobileNetV2 (0.5$\times$)}
} & 1/2 & 9, 36, 18, 27, 36, 27, 12, 27, 80, 40 & 9, 36, 18, 27, 36, 27, 6, 27, 40, 20 \\
& 1 & 9, 36, 34, 78, 18, 42, 27, 102, 36, 120, 27, 58, 27 & 9, 36, 34, 78, 18, 42, 27, 102, 36, 120, 27, 58, 27 \\
& 4 & 9, 36, 11, 1, 2, 18, 7, 3, 27, 4, 4, 36, 9, 3, 27, 11, 3, 27, 20 &36, 144, 44, 4, 8, 72, 28, 12, 108, 16, 16, 144, 36, 12, 108, 44, 12, 108, 80 \\

\hline
\end{tabular}
}
\end{center}
\vskip -0.1 in
\end{table}

\begin{table}[tbh]
\vskip -0.15 in
\caption{The example of warehouse sharing for the ResNet18 backbone based on KW ($1\times$) according to the original stages and reassigned stages.}
\vskip -0.0 in
\label{table:stage}
\begin{center}
\resizebox{.8\linewidth}{!}{
\begin{tabular}{c|c|c|c|c}
\hline
Dimensions of Kernel Cells & Original Stages & Layers & Reassigned Stages & Dimensions of Kernel Cells \\
\hline
\multirow{4}{*}{1$\times$1$\times$64$\times$64} & \multirow{4}{*}{1} & 3$\times$3$\times$64$\times$64 & \multirow{5}{*}{1} & \multirow{5}{*}{1$\times$1$\times$64$\times$64} \\
\cline{3-3}
& & 3$\times$3$\times$64$\times$64 & & \\
\cline{3-3}
& & 3$\times$3$\times$64$\times$64 & & \\
\cline{3-3}
& & 3$\times$3$\times$64$\times$64 & & \\
\cline{3-3}
\cline{1-2}
\multirow{4}{*}{1$\times$1$\times$64$\times$128} & \multirow{4}{*}{2} & 3$\times$3$\times$64$\times$128 &  & \\
\cline{3-3}
\cline{4-5}
& & 3$\times$3$\times$128$\times$128 &  \multirow{4}{*}{2} & \multirow{4}{*}{1$\times$1$\times$128$\times$128}\\
\cline{3-3}
& & 3$\times$3$\times$128$\times$128 & & \\
\cline{3-3}
& & 3$\times$3$\times$128$\times$128 & & \\
\cline{1-2}
\cline{3-3}
\multirow{4}{*}{1$\times$1$\times$128$\times$256} & \multirow{4}{*}{3} & 3$\times$3$\times$128$\times$256 & & \\
\cline{4-5}
\cline{3-3}
& & 3$\times$3$\times$256$\times$256 & \multirow{4}{*}{3} & \multirow{4}{*}{1$\times$1$\times$256$\times$256} \\
\cline{3-3}
& & 3$\times$3$\times$256$\times$256 & & \\
\cline{3-3}
& & 3$\times$3$\times$256$\times$256 & &  \\
\cline{1-2}
\cline{3-3}
\multirow{4}{*}{1$\times$1$\times$256$\times$512} & \multirow{4}{*}{4} & 3$\times$3$\times$256$\times$512 & & \\
\cline{4-5}
\cline{3-3}
& & 3$\times$3$\times$512$\times$512 & \multirow{3}{*}{4} & \multirow{3}{*}{1$\times$1$\times$512$\times$512}\\
\cline{3-3}
& & 3$\times$3$\times$512$\times$512 & &  \\
\cline{3-3}
& & 3$\times$3$\times$512$\times$512 & & \\
\hline
\end{tabular}
}
\end{center}
\vskip -0.1 in
\end{table}

\textbf{Design details of KernelWarehouse on ResNet18.}
Recall that in KernelWarehouse, a warehouse is shared to all same-stage convolutional layers.
While the layers are originally divided into different stages according to the resolutions of their input feature maps, the layers are divided into different stages according to their kernel dimensions in our KernelWarehouse. In our implementation, we usually reassign the first layer (or the first two layers) in each stage to the previous stage. An example for the ResNet18 backbone based on KW ($1\times$) is given in Table~\ref{table:stage}.
By reassigning the layers, we can avoid the condition that all the other layers have to be partitioned according to a single layer because of the greatest common dimension divisors.
For the ResNet18 backbone, we apply KernelWarehouse to all the convolutional layers except the first one. In each stage, the corresponding warehouse is shared to all of its convolutional layers. For KW ($1\times$), KW ($2\times$) and KW ($4\times$), we use the greatest common dimension divisors for static kernels as the uniform kernel cell dimensions for kernel partition. For KW ($1/2\times$) and KW ($1/4\times$), we use half of the greatest common dimension divisors.

\textbf{Design details of KernelWarehouse on ResNet50.}
For the ResNet50 backbone, we apply KernelWarehouse to all the convolutional layers except the first two layers. In each stage, the corresponding warehouse is shared to all of its convolutional layers. For KW ($1\times$) and KW ($4\times$), we use the greatest common dimension divisors for static kernels as the uniform kernel cell dimensions for kernel partition. For KW ($1/2\times$), we use half of the greatest common dimension divisors.

\textbf{Design details of KernelWarehouse on ConvNeXt-Tiny.}
For the ConvNeXt backbone, we apply KernelWarehouse to all the convolutional layers.
In each stage, the corresponding three warehouses are shared to the point-wise convolutional layers, the depth-wise convolutional layers and the downsampling layer, respectively.
We use the greatest common dimension divisors for static kernels as the uniform kernel cell dimensions for kernel partition.

\textbf{Design details of KernelWarehouse on MobileNetV2.}
For the MobileNetV2 (1.0$\times$) and MobileNetV2 (0.5$\times$) backbones based on KW ($1\times$) and KW ($4\times$), we apply KernelWarehouse to all the convolutional layers. For MobileNetV2 (1.0$\times$, 0.5$\times$) based on KW ($1\times$), the corresponding two warehouses are shared to the point-wise convolutional layers and the depth-wise convolutional layers in each stage, respectively. For MobileNetV2 (1.0$\times$, 0.5$\times$) based on KW ($4\times$), the corresponding three warehouses are shared to the depth-wise convolutional layers, the point-wise convolutional layers for channel expansion and the point-wise convolutional layers for channel reduction in each stage, respectively. We use the greatest common dimension divisors for static kernels as the uniform kernel cell dimensions for kernel partition.
For the MobileNetV2 (1.0$\times$) and MobileNetV2 (0.5$\times$) backbones based on KW ($1/2\times$), we take the parameters in the attention modules and classifier layer into consideration in order to reduce the total number of parameters. We apply KernelWarehouse to all the depth-wise convolutional layers, the point-wise convolutional layers in the last two stages and the classifier layer. We set $b=1$ for the point-wise convolutional layers and $b=1/2$ for the other layers. For the depth-wise convolutional layers, we use the greatest common dimension divisors for static kernels as the uniform kernel cell dimensions for kernel partition. For the point-wise convolutional layers, we use half of the greatest common dimension divisors. For the classifier layer, we use the kernel cell dimensions of 1000$\times$32.

\subsection{More Visualization Results for Learnt Attentions of KernelWarehouse}
In the main manuscript, we provide visualization results of learnt attention values $\alpha_{ij}$ for the ResNet18 backbone based on KW ($1\times$) (see Figure~3 in the main manuscript). For a better understanding of KernelWarehouse, we provide more visualization results in this section, covering different alternative attention functions, alternative initialization strategies and values of $b$.
For all the results, the statistical mean values of learnt attention $\alpha_{ij}$ are obtained using all of the 50,000 images on the ImageNet validation dataset.

\begin{figure}[thp]
    \begin{minipage}[t]{1.0\linewidth}
        \begin{minipage}[t]{0.24\linewidth}
            \begin{center}
                \includegraphics[width=\textwidth]{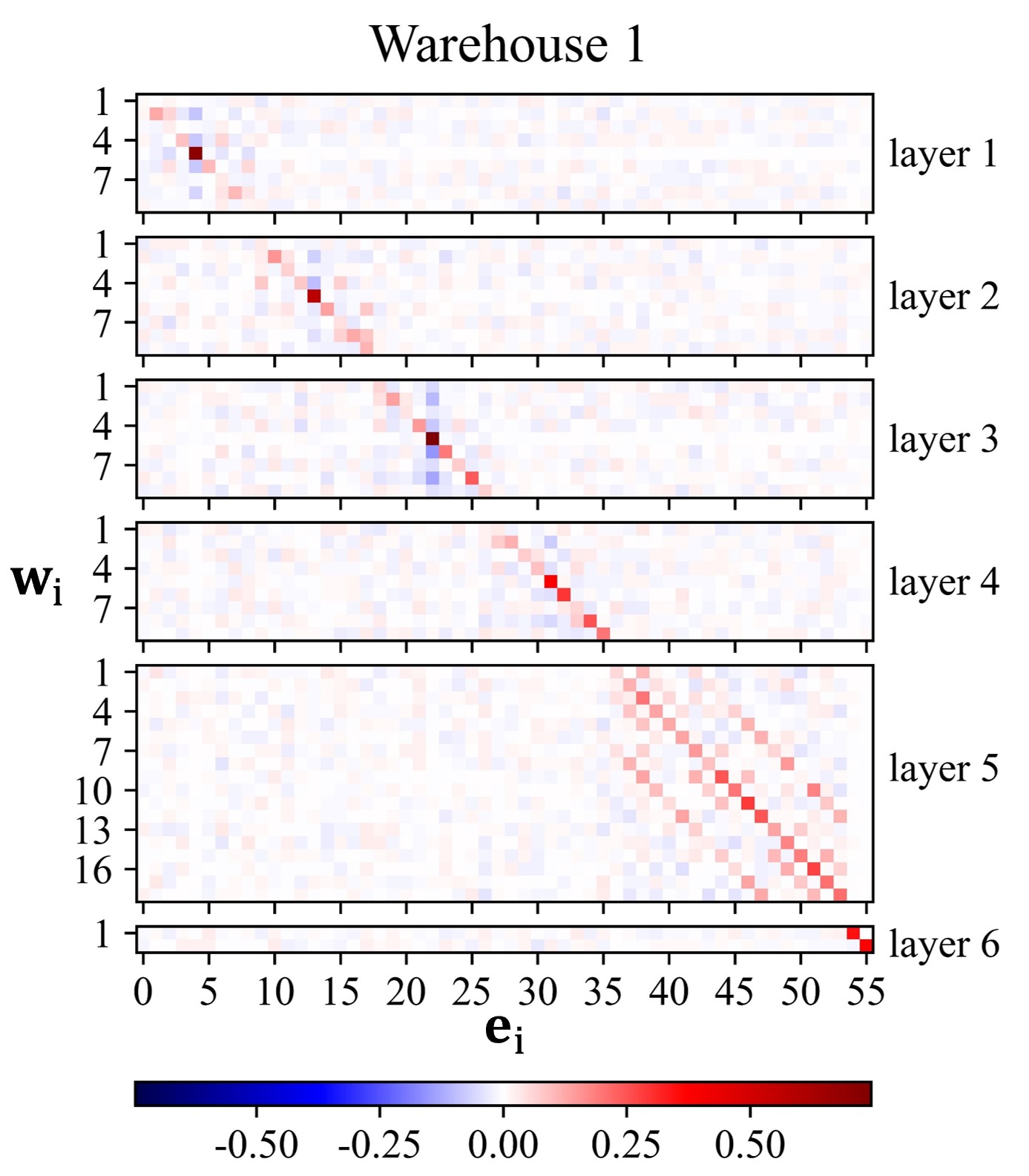}
            \end{center}
        \end{minipage}
        \hfill
        \begin{minipage}[t]{0.24\linewidth}
            \begin{center}
                \includegraphics[width=\textwidth]{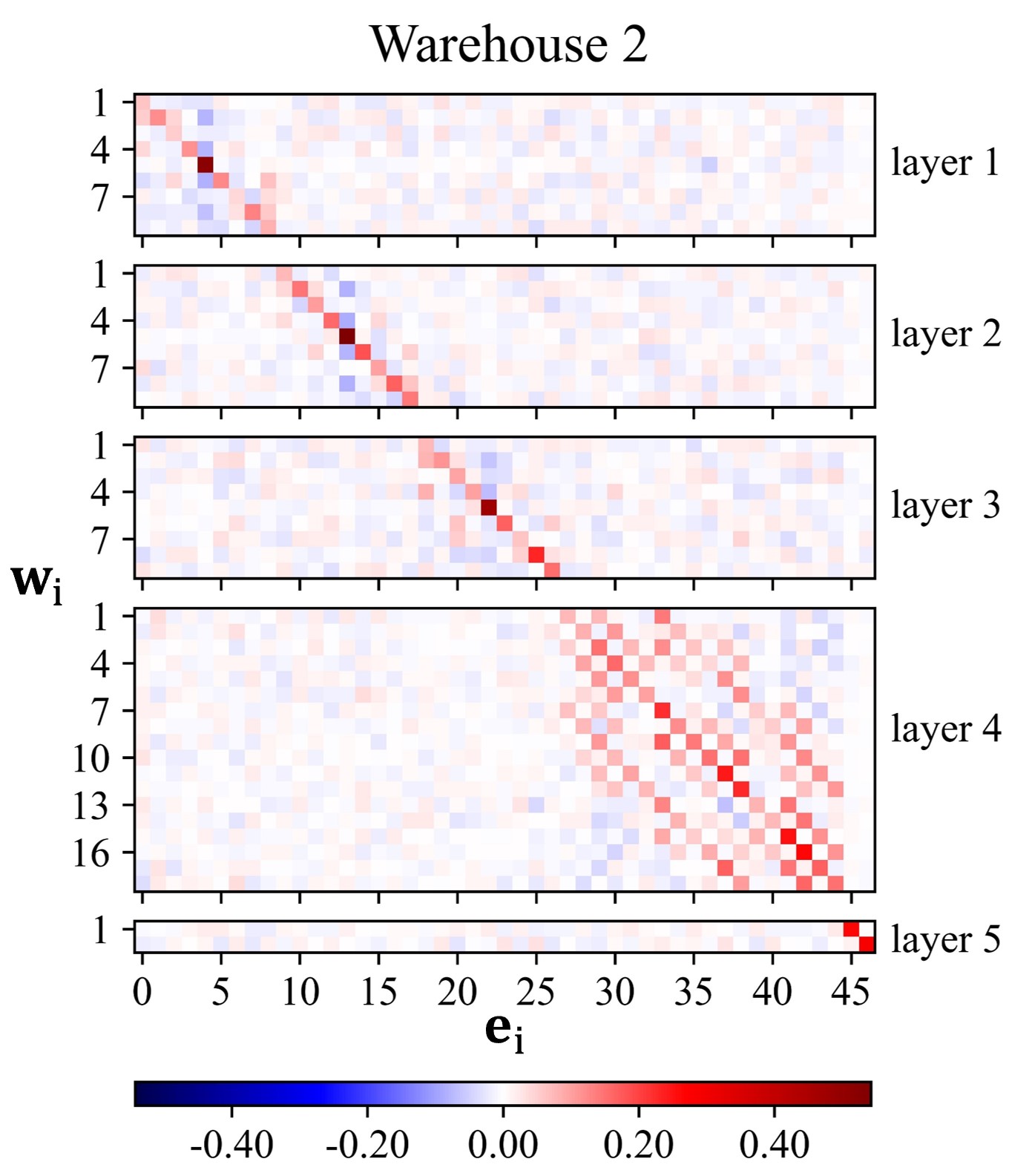}
            \end{center}
        \end{minipage}
        \hfill
        \begin{minipage}[t]{0.24\linewidth}
            \begin{center}
                \includegraphics[width=\textwidth]{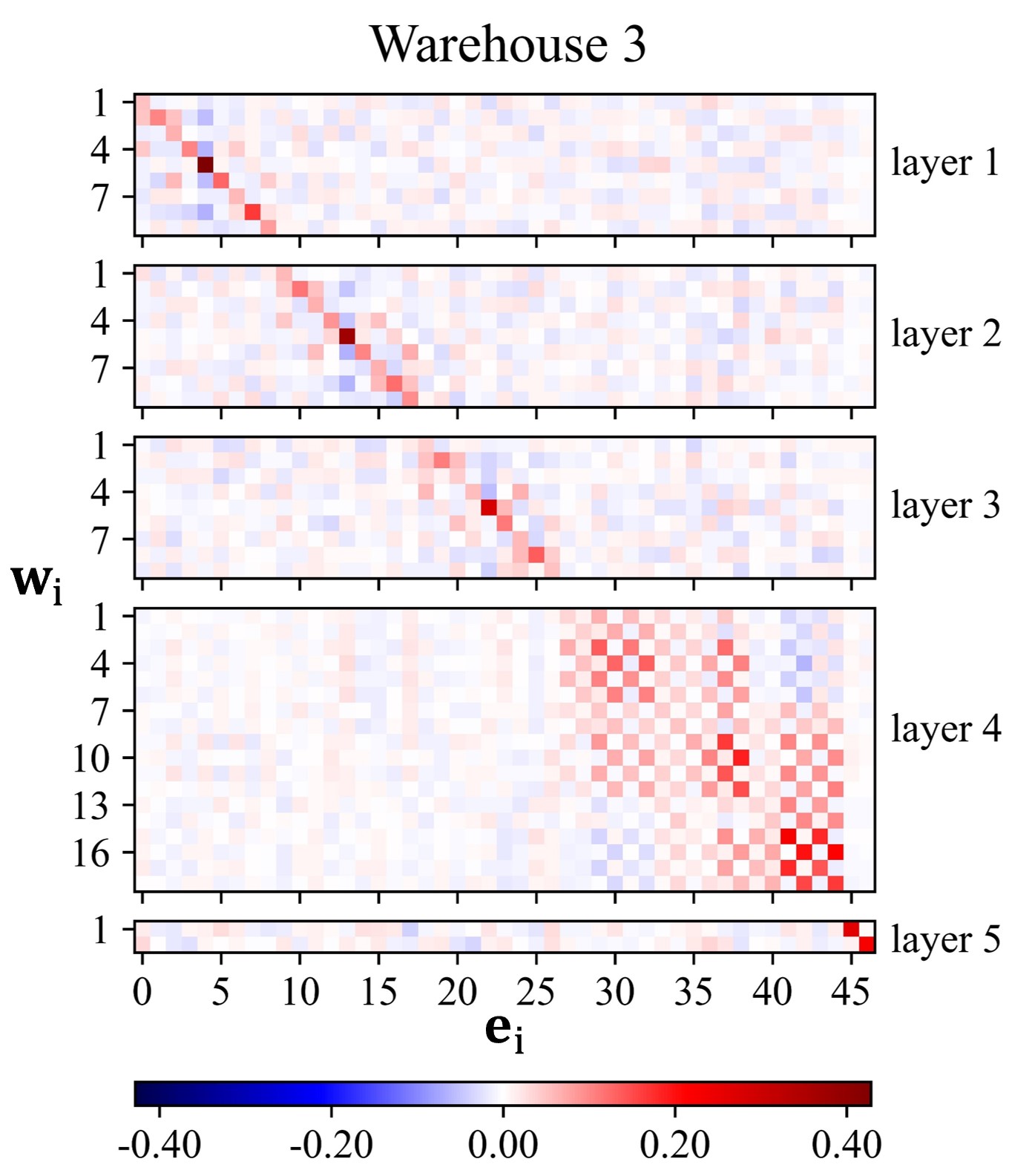}
            \end{center}
        \end{minipage}
            \hfill
        \begin{minipage}[t]{0.24\linewidth}
            \begin{center}
                \includegraphics[width=\textwidth]{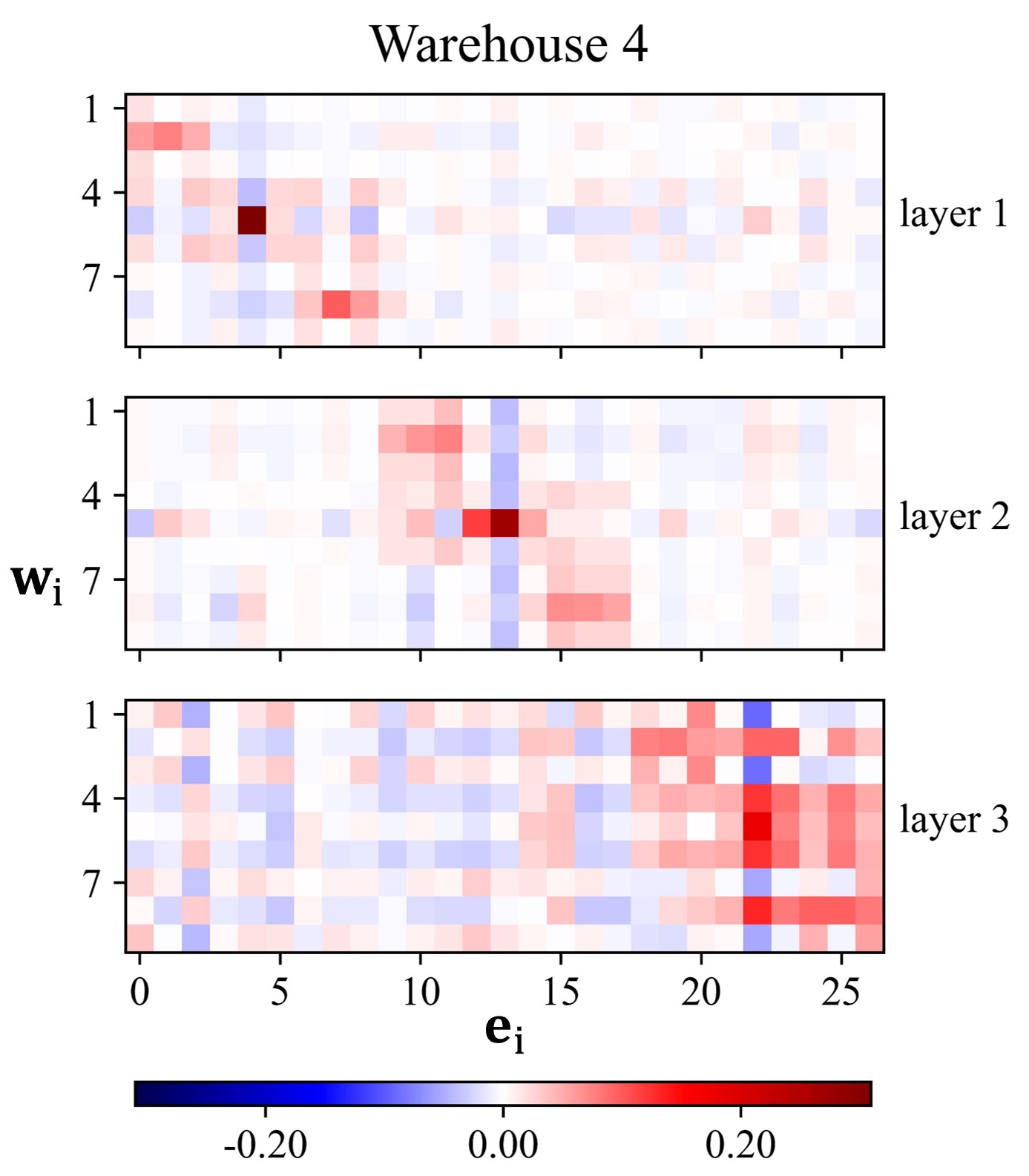}
            \end{center}
        \end{minipage}
    \subcaption{}
    \end{minipage}
    \begin{minipage}[t]{1.0\linewidth}
        \begin{minipage}[t]{0.24\linewidth}
            \begin{center}
                \includegraphics[width=\textwidth]{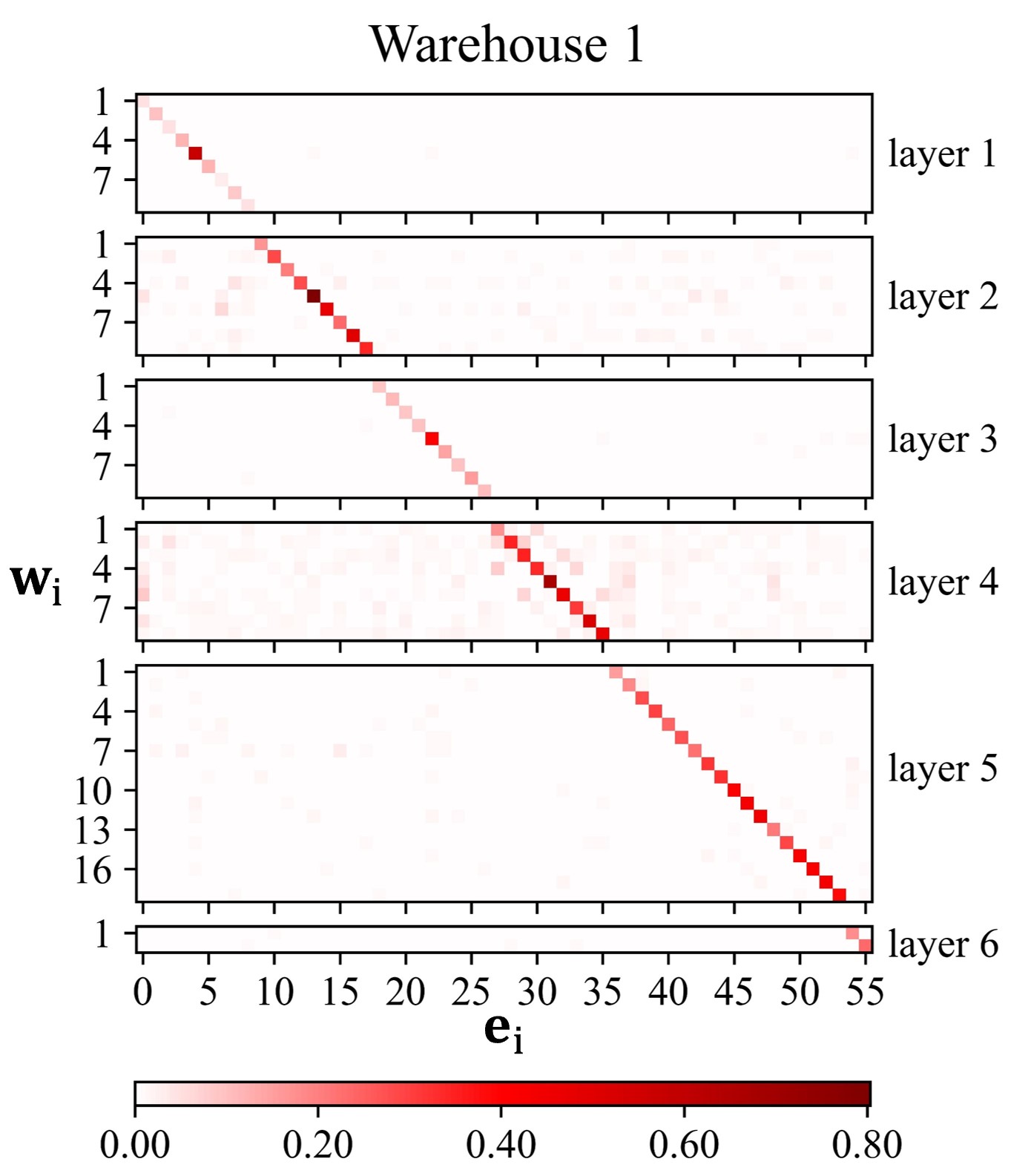}
            \end{center}
        \end{minipage}
        \hfill
        \begin{minipage}[t]{0.24\linewidth}
            \begin{center}
                \includegraphics[width=\textwidth]{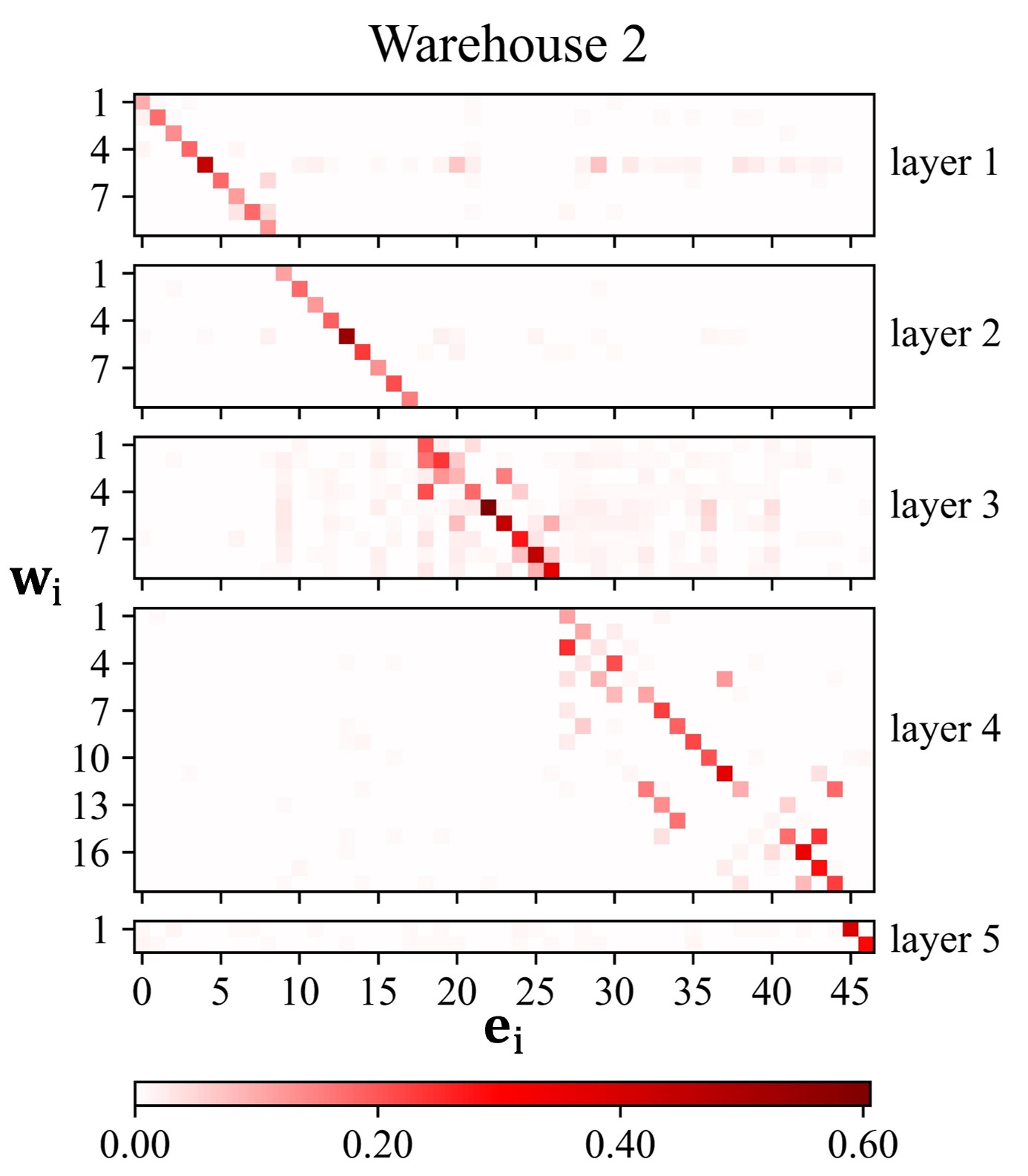}
            \end{center}
        \end{minipage}
        \hfill
        \begin{minipage}[t]{0.24\linewidth}
            \begin{center}
                \includegraphics[width=\textwidth]{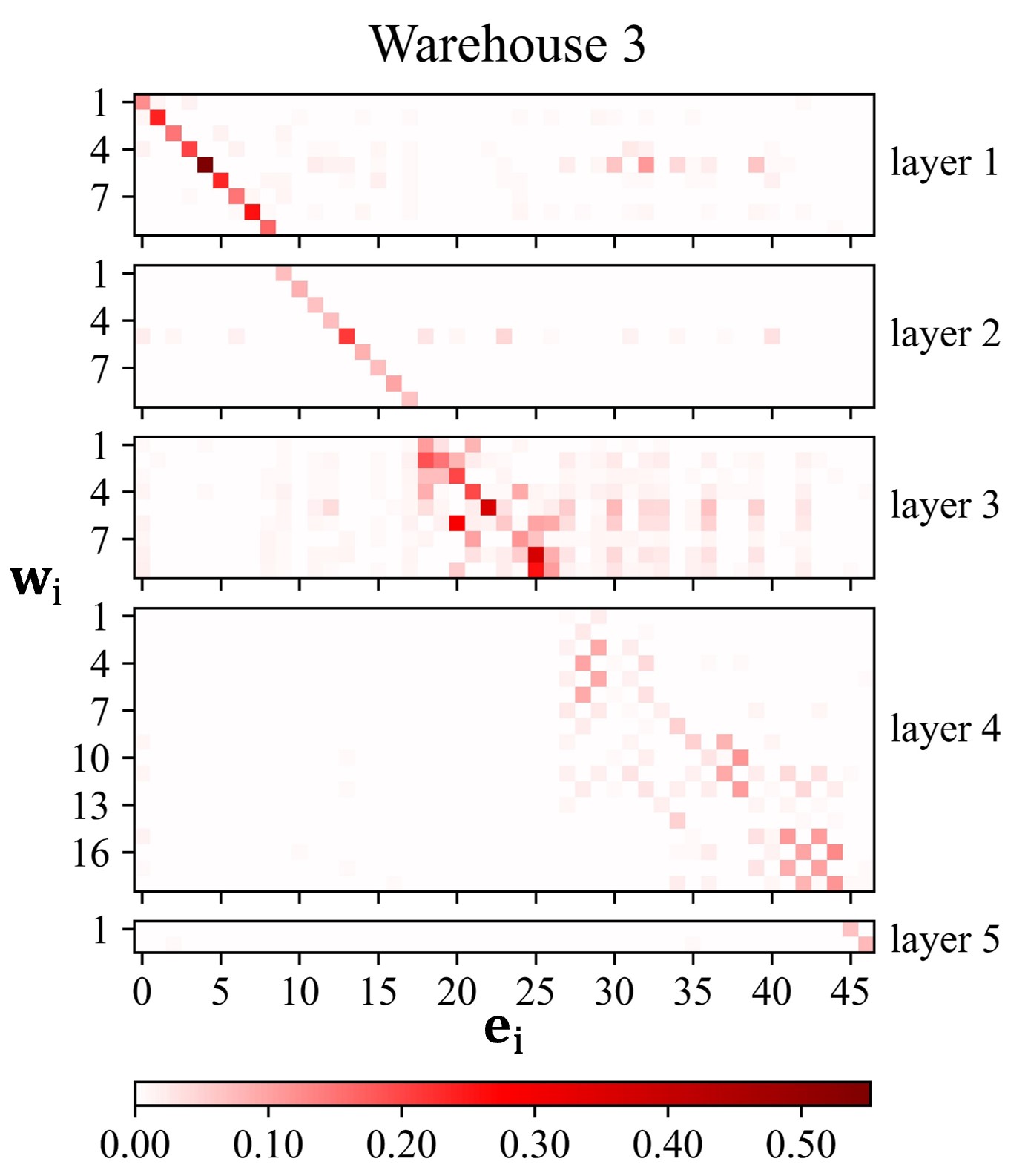}
            \end{center}
        \end{minipage}
            \hfill
        \begin{minipage}[t]{0.24\linewidth}
            \begin{center}
                \includegraphics[width=\textwidth]{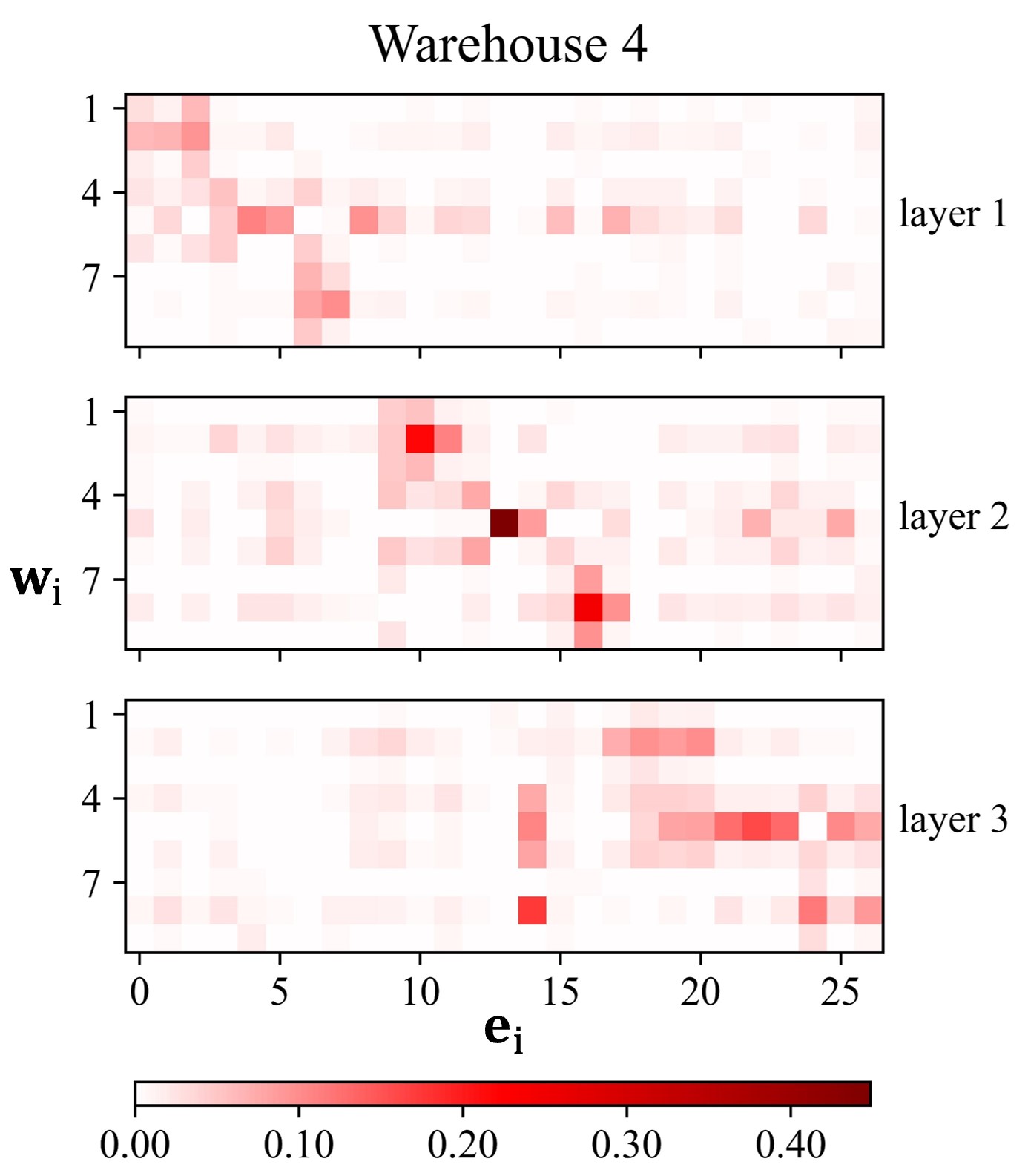}
            \end{center}
        \end{minipage}
    \subcaption{}
    \end{minipage}
    \begin{minipage}[t]{1.0\linewidth}
        \begin{minipage}[t]{0.24\linewidth}
            \begin{center}
                \includegraphics[width=\textwidth]{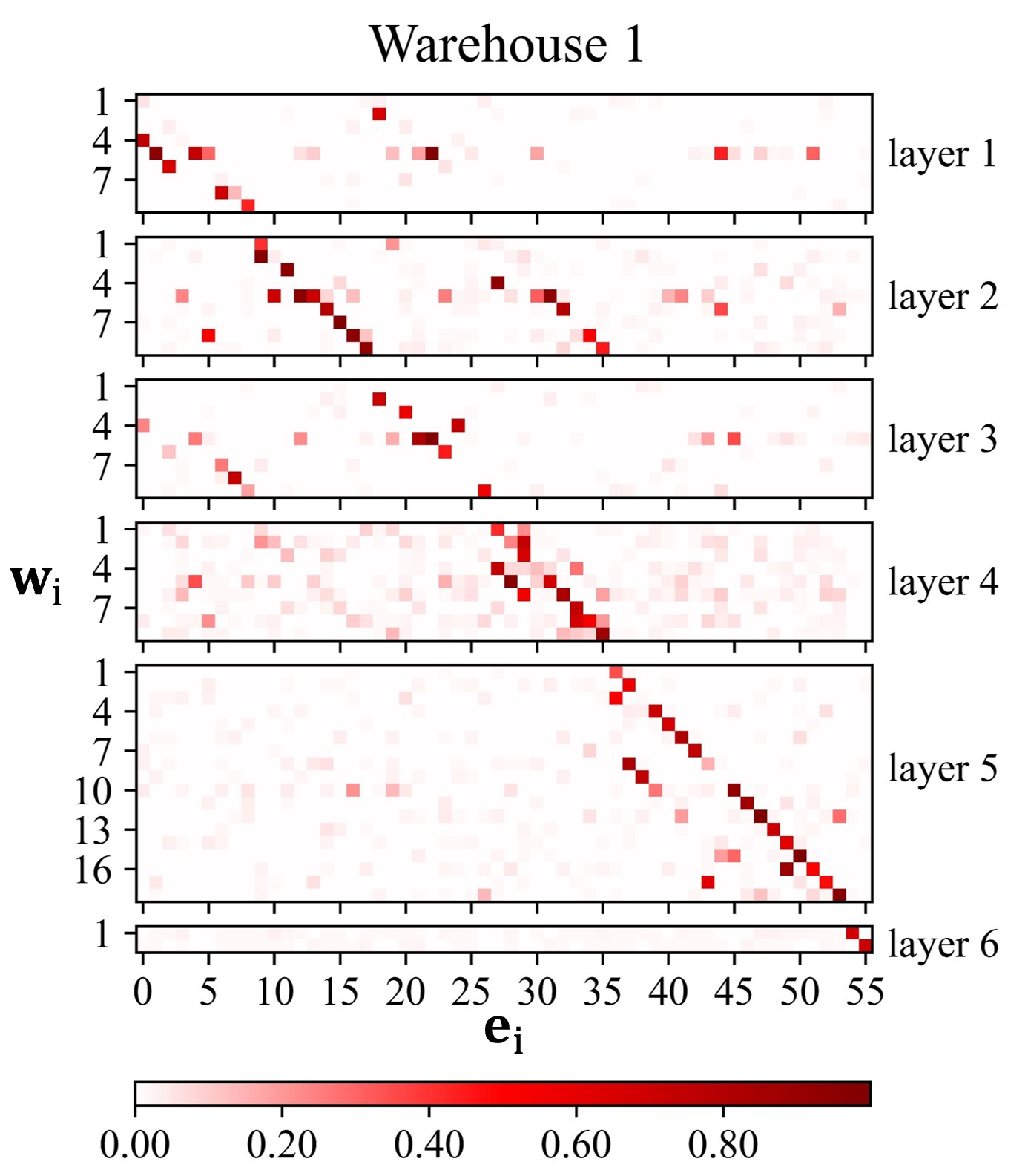}
            \end{center}
        \end{minipage}
        \hfill
        \begin{minipage}[t]{0.24\linewidth}
            \begin{center}
                \includegraphics[width=\textwidth]{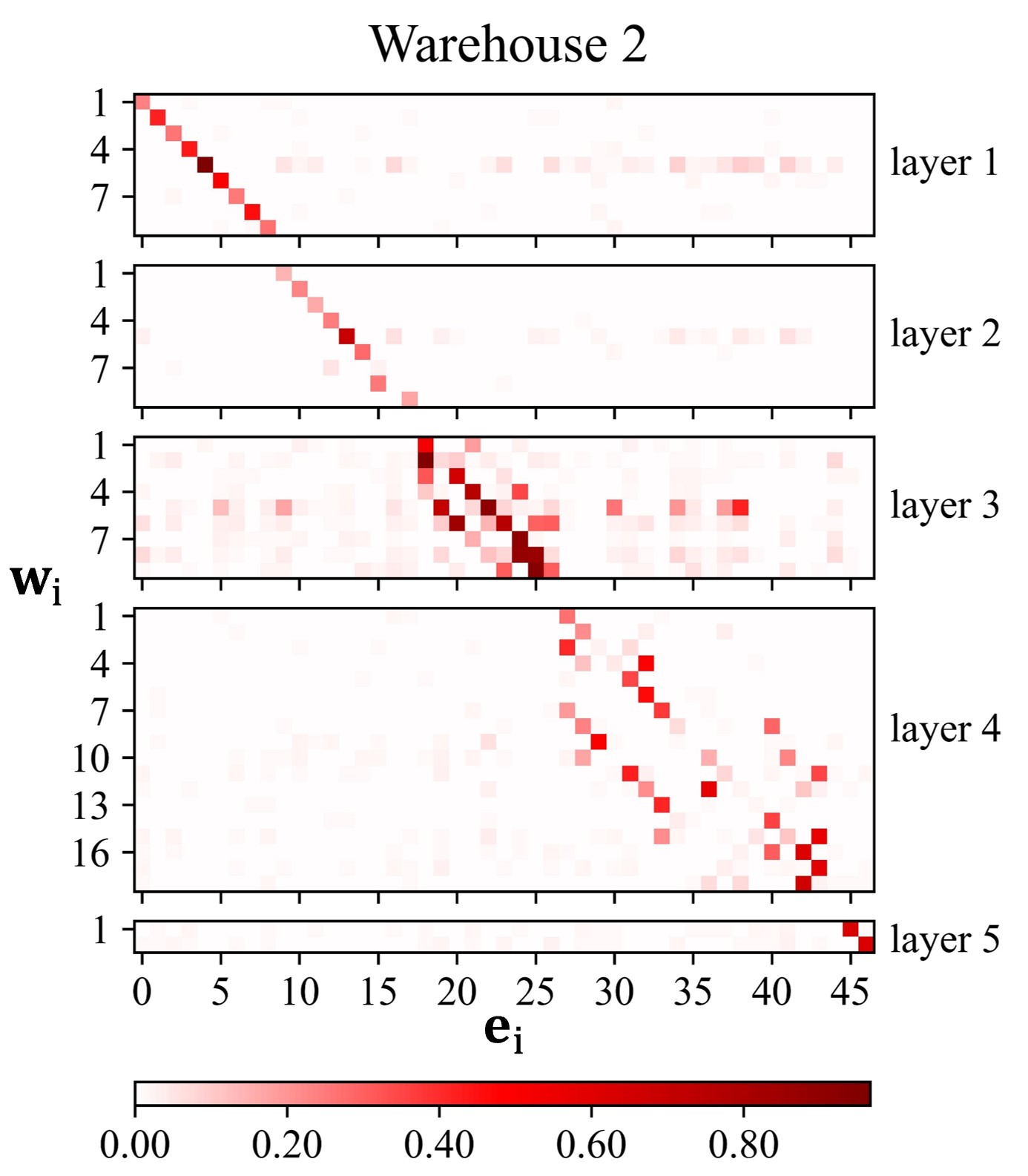}
            \end{center}
        \end{minipage}
        \hfill
        \begin{minipage}[t]{0.24\linewidth}
            \begin{center}
                \includegraphics[width=\textwidth]{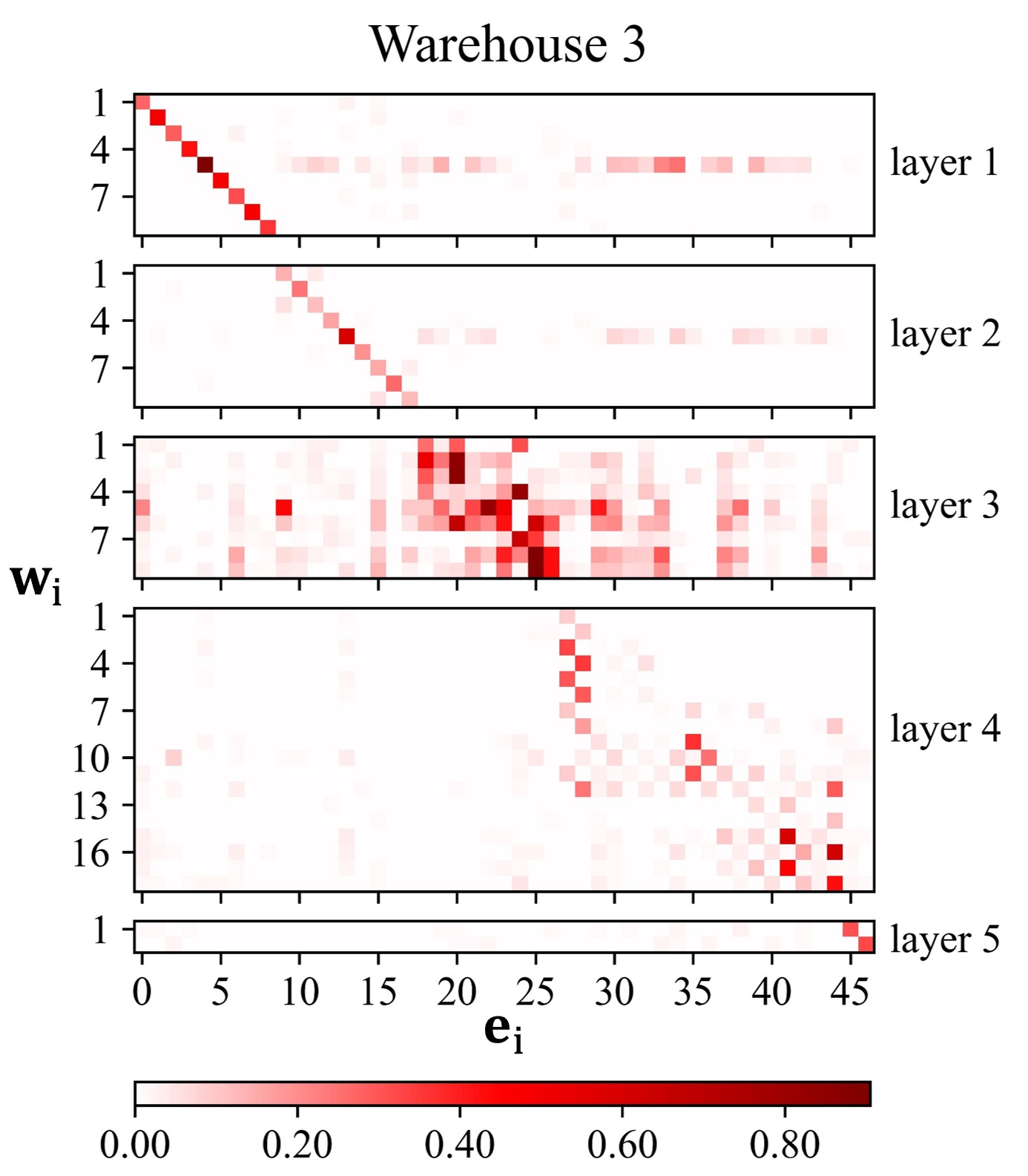}
            \end{center}
        \end{minipage}
            \hfill
        \begin{minipage}[t]{0.24\linewidth}
            \begin{center}
                \includegraphics[width=\textwidth]{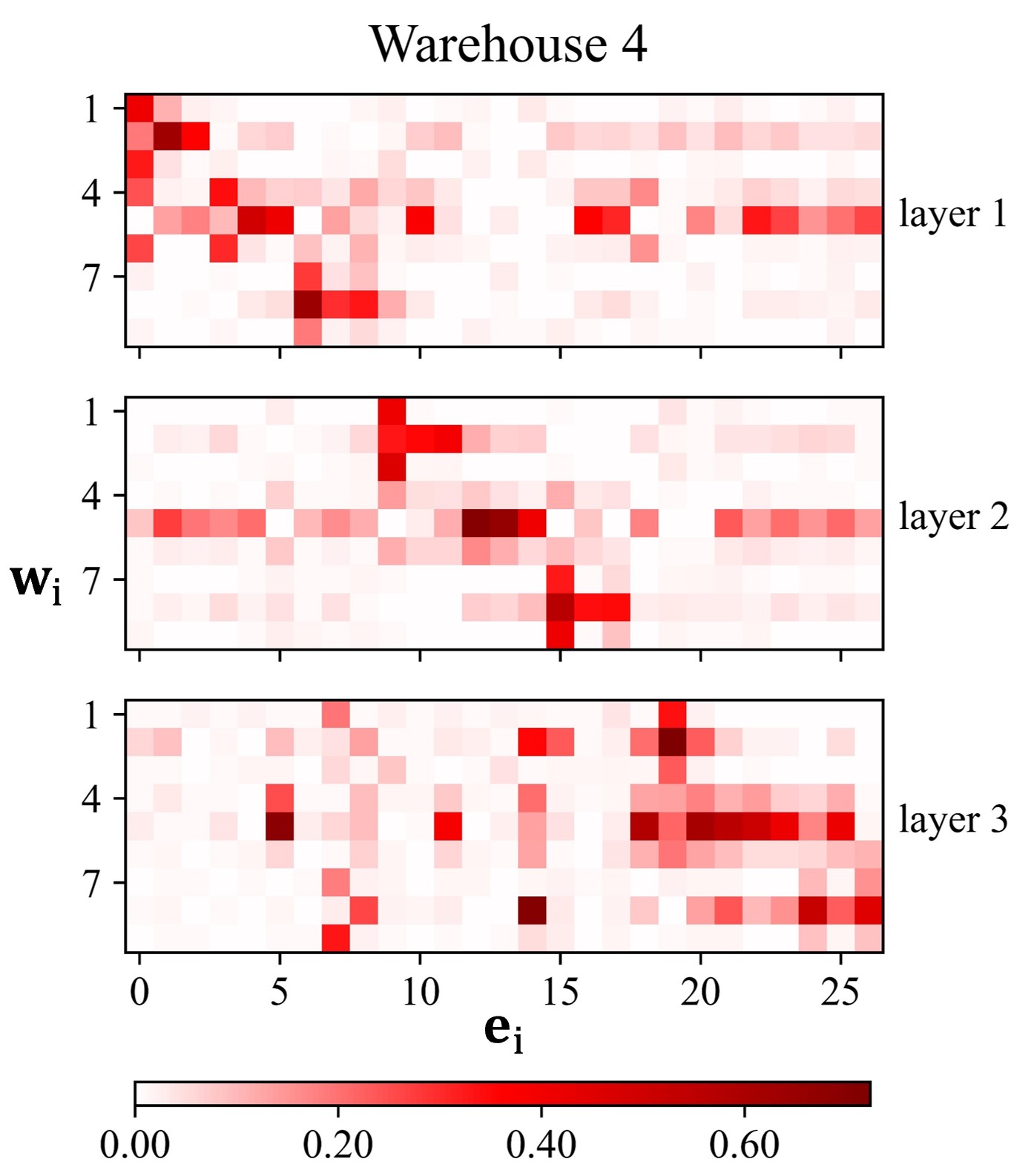}
            \end{center}
        \end{minipage}
    \subcaption{}
    \end{minipage}
    \begin{minipage}[t]{1.0\linewidth}
        \begin{minipage}[t]{0.24\linewidth}
            \begin{center}
                \includegraphics[width=\textwidth]{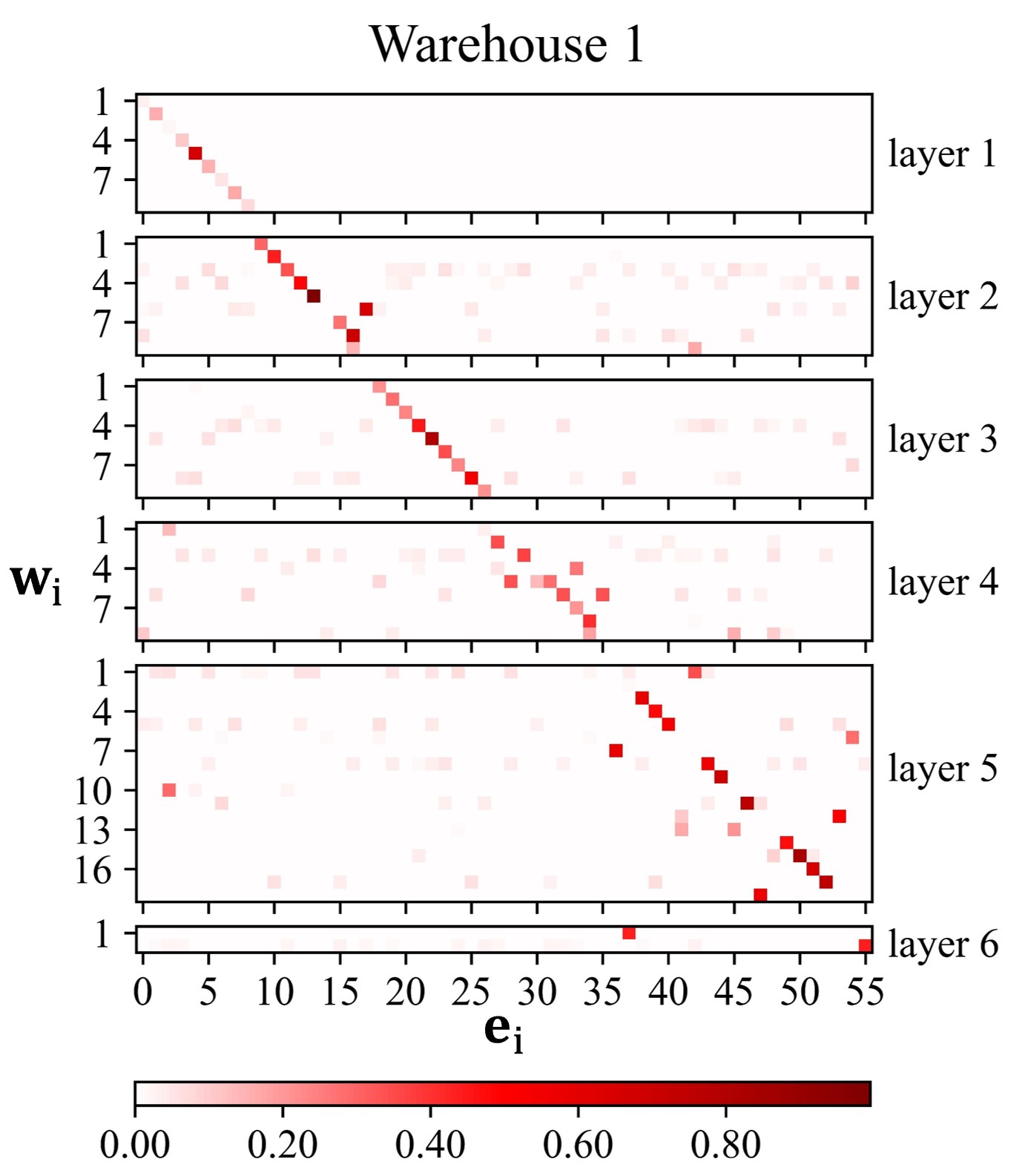}
            \end{center}
        \end{minipage}
        \hfill
        \begin{minipage}[t]{0.24\linewidth}
            \begin{center}
                \includegraphics[width=\textwidth]{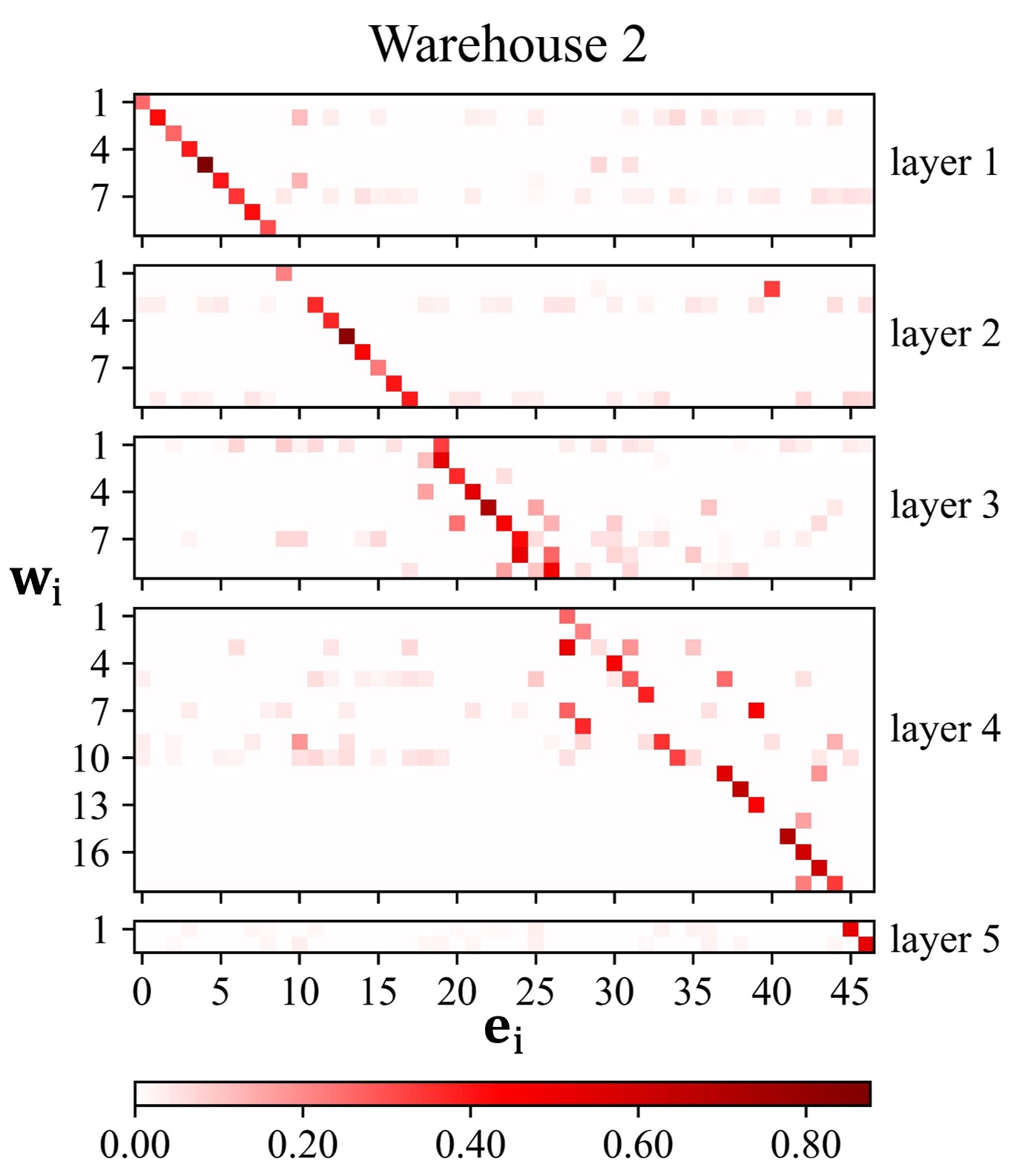}
            \end{center}
        \end{minipage}
        \hfill
        \begin{minipage}[t]{0.24\linewidth}
            \begin{center}
                \includegraphics[width=\textwidth]{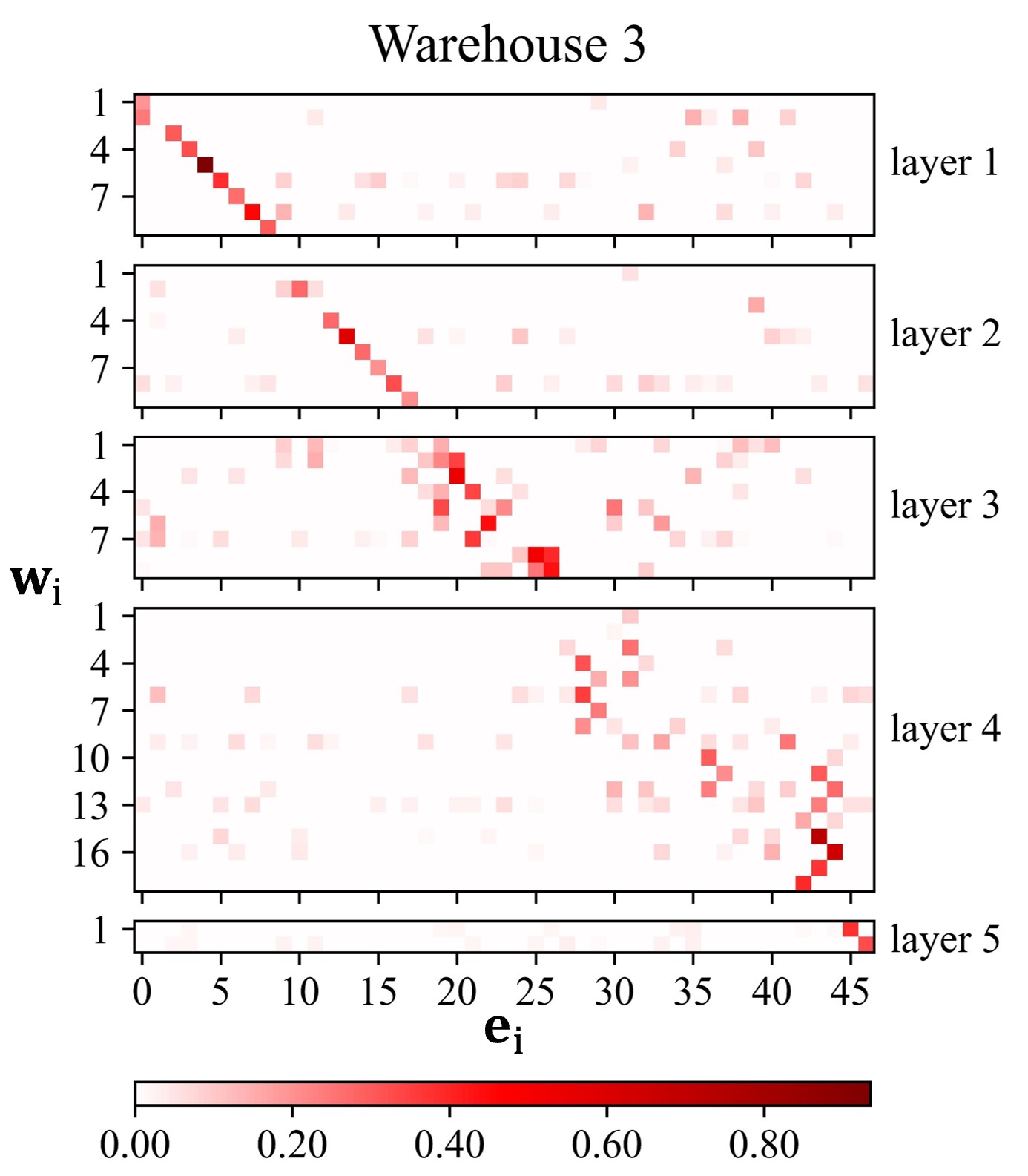}
            \end{center}
        \end{minipage}
            \hfill
        \begin{minipage}[t]{0.24\linewidth}
            \begin{center}
                \includegraphics[width=\textwidth]{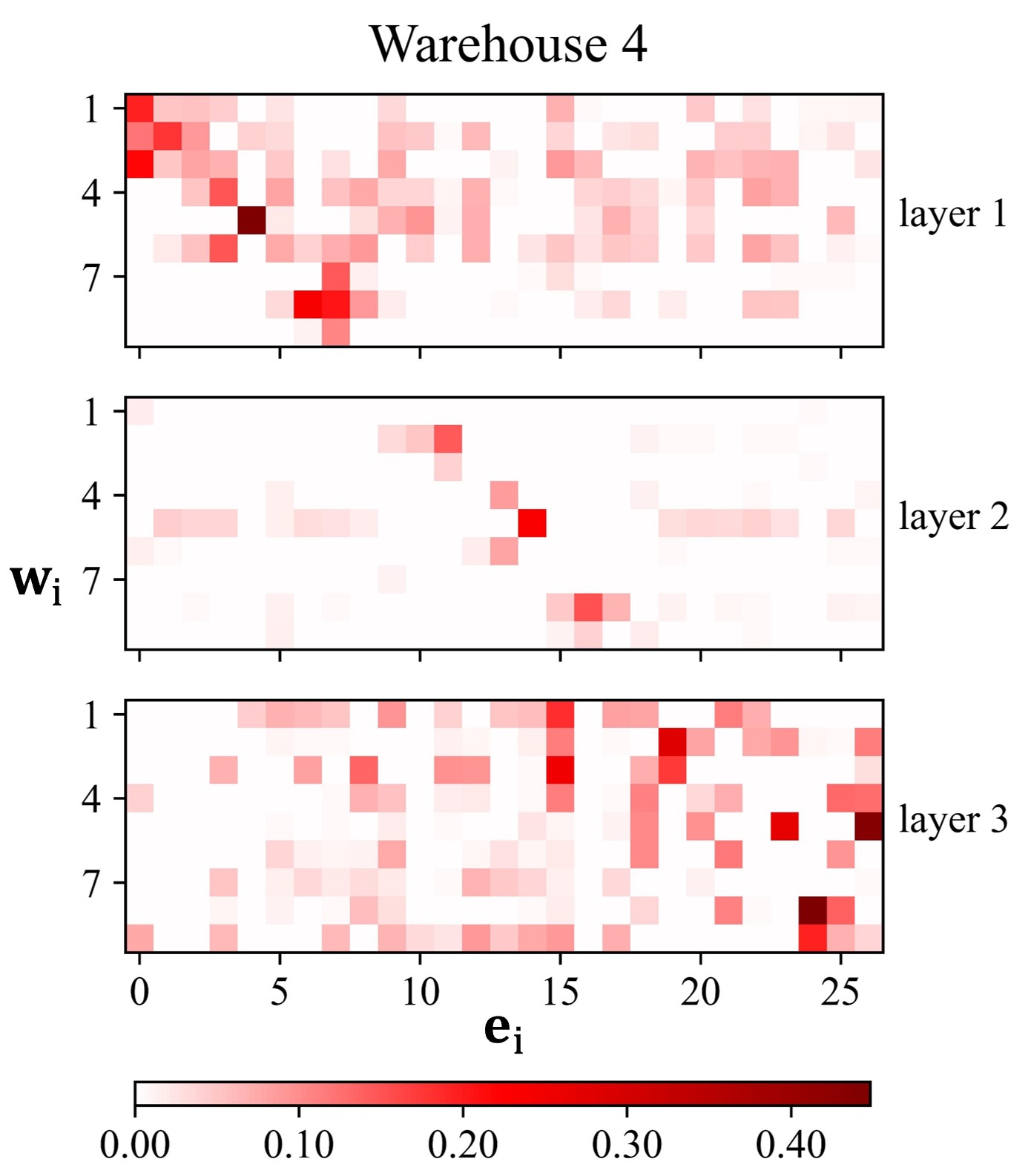}
            \end{center}
        \end{minipage}
    \subcaption{}
    \end{minipage}
    \caption{Visualization of statistical mean values of learnt attention $\alpha_{ij}$ in each warehouse for KernelWarehouse with different attention functions. The results are obtained from the pre-trained ResNet18 backbone with KW ($1\times$) for all of the 50,000 images on the ImageNet validation dataset. Best viewed with zoom-in.
    The attention functions for the groups of visualization results are as follows:
    (a) $z_{ij}/\sum^{n}_{p=1}|z_{ip}|$ (our design); (b) softmax; (c) sigmoid; (d) $max(z_{ij},0)/\sum^{n}_{p=1}|z_{ip}|$.}
    \label{fig:visualization_attention_function}

\end{figure}

\textbf{Visualization Results for KernelWarehouse with Different Attention Functions.}
The visualization results for KernelWarehouse with different attention functions are shown in Figure~\ref{fig:visualization_attention_function}, which are corresponding to the comparison results of Table~8 in the main manuscript. From which we can observe that: (1) for all of the attention functions, the maximum value of $\alpha_{ij}$ in each row mostly appears in the diagonal line throughout the whole warehouse. It indicates that our proposed attentions initialization strategy also works for the other three attention functions, which helps our KernelWarehouse to build one-to-one relationships between kernel cells and linear mixtures; (2) with different attention functions, the scalar attentions learnt by KernelWarehouse are obviously different, showing that the attention function plays an important role in our design; (3) compared to the other three functions, the maximum value of $\alpha_{ij}$ in each row tends to be relatively lower for our design (shown in Figure~\ref{fig:visualization_attention_function}(a)). It indicates that the introduction of negative values for scalar attentions can help the ConvNet to enhance warehouse sharing, where each linear mixture not only focuses on the kernel cell assigned to it.

\begin{figure}[thp]
    \begin{minipage}[t]{1.0\linewidth}
        \begin{minipage}[t]{0.24\linewidth}
            \begin{center}
                \includegraphics[width=\textwidth]{Figures/Attentions/resnet18_1x/Fig_Warehouse1.jpg}
            \end{center}
        \end{minipage}
        \hfill
        \begin{minipage}[t]{0.24\linewidth}
            \begin{center}
                \includegraphics[width=\textwidth]{Figures/Attentions/resnet18_1x/Fig_Warehouse2.jpg}
            \end{center}
        \end{minipage}
        \hfill
        \begin{minipage}[t]{0.24\linewidth}
            \begin{center}
                \includegraphics[width=\textwidth]{Figures/Attentions/resnet18_1x/Fig_Warehouse3.jpg}
            \end{center}
        \end{minipage}
            \hfill
        \begin{minipage}[t]{0.24\linewidth}
            \begin{center}
                \includegraphics[width=\textwidth]{Figures/Attentions/resnet18_1x/Fig_Warehouse4.jpg}
            \end{center}
        \end{minipage}
    \subcaption{}
    \end{minipage}
    \begin{minipage}[t]{1.0\linewidth}
        \begin{minipage}[t]{0.24\linewidth}
            \begin{center}
                \includegraphics[width=\textwidth]{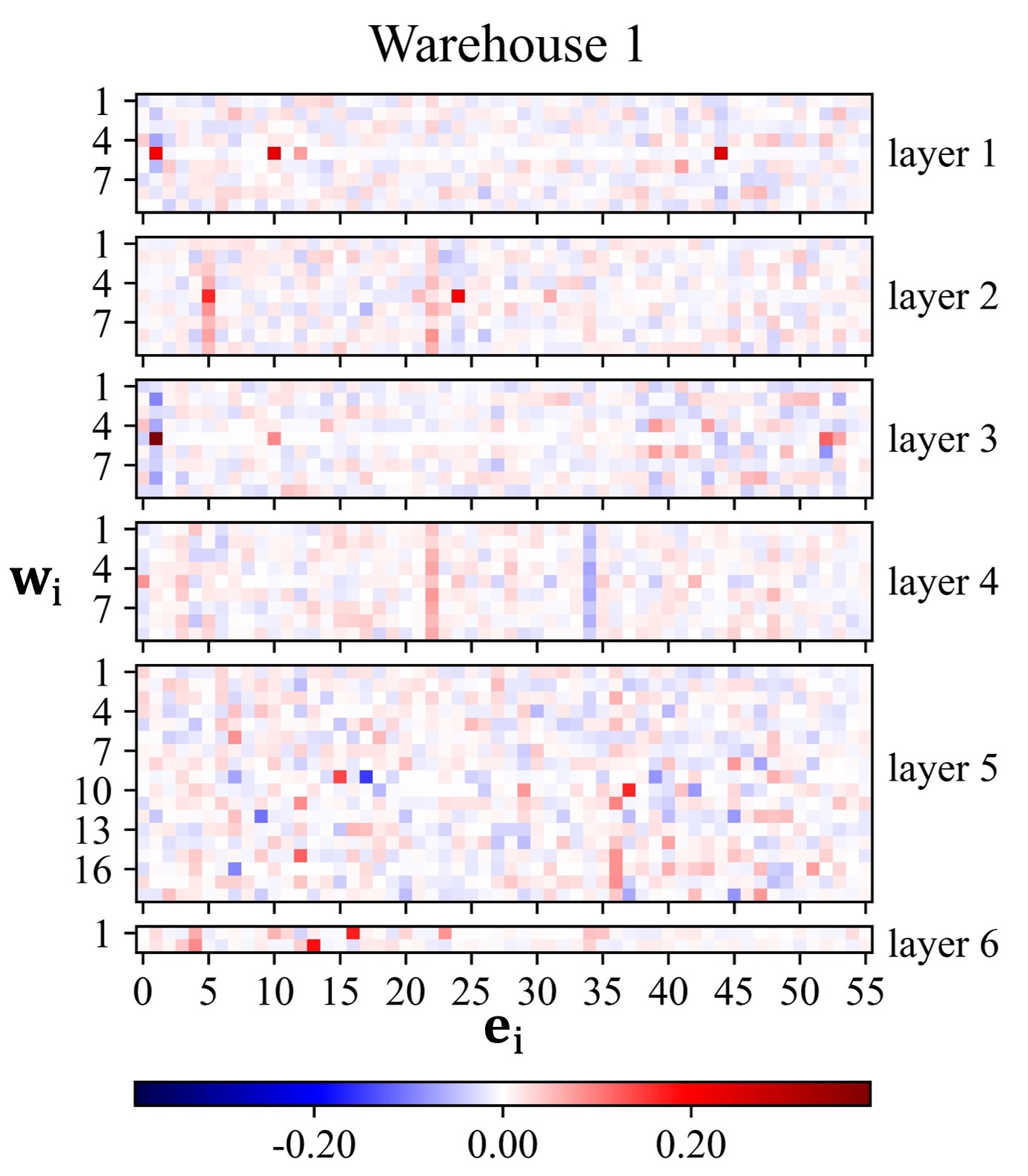}
            \end{center}
        \end{minipage}
        \hfill
        \begin{minipage}[t]{0.24\linewidth}
            \begin{center}
                \includegraphics[width=\textwidth]{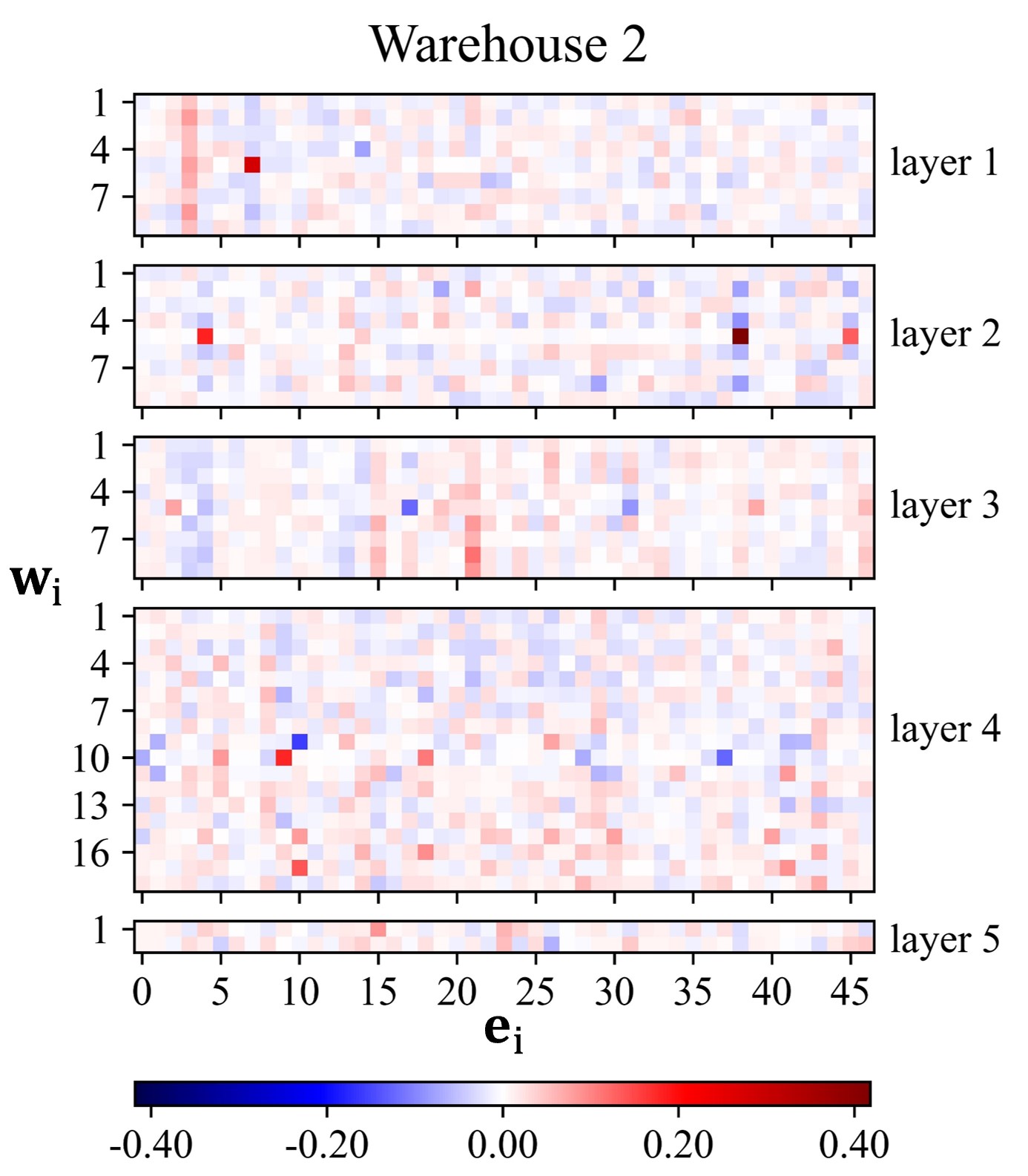}
            \end{center}
        \end{minipage}
        \hfill
        \begin{minipage}[t]{0.24\linewidth}
            \begin{center}
                \includegraphics[width=\textwidth]{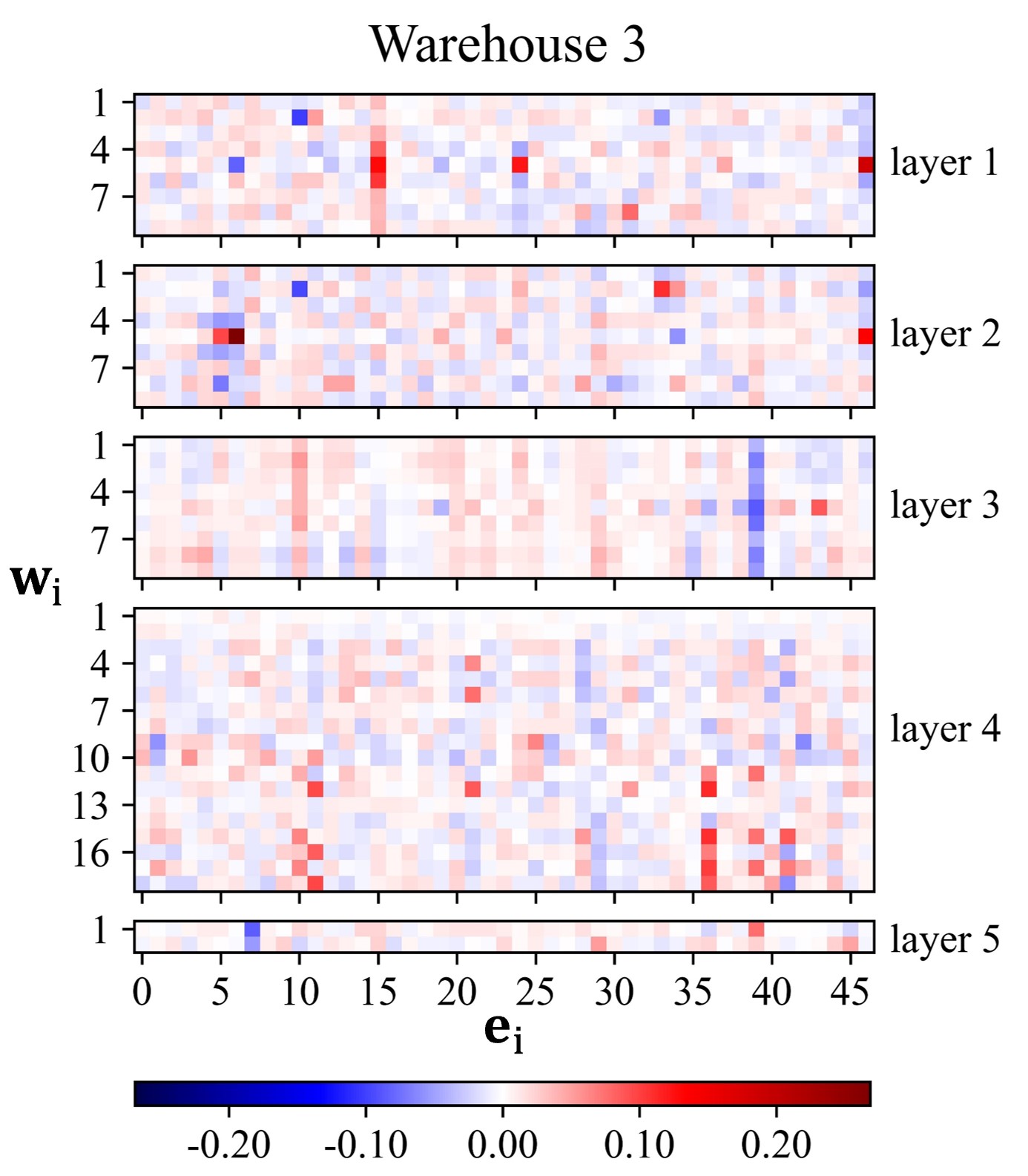}
            \end{center}
        \end{minipage}
            \hfill
        \begin{minipage}[t]{0.24\linewidth}
            \begin{center}
                \includegraphics[width=\textwidth]{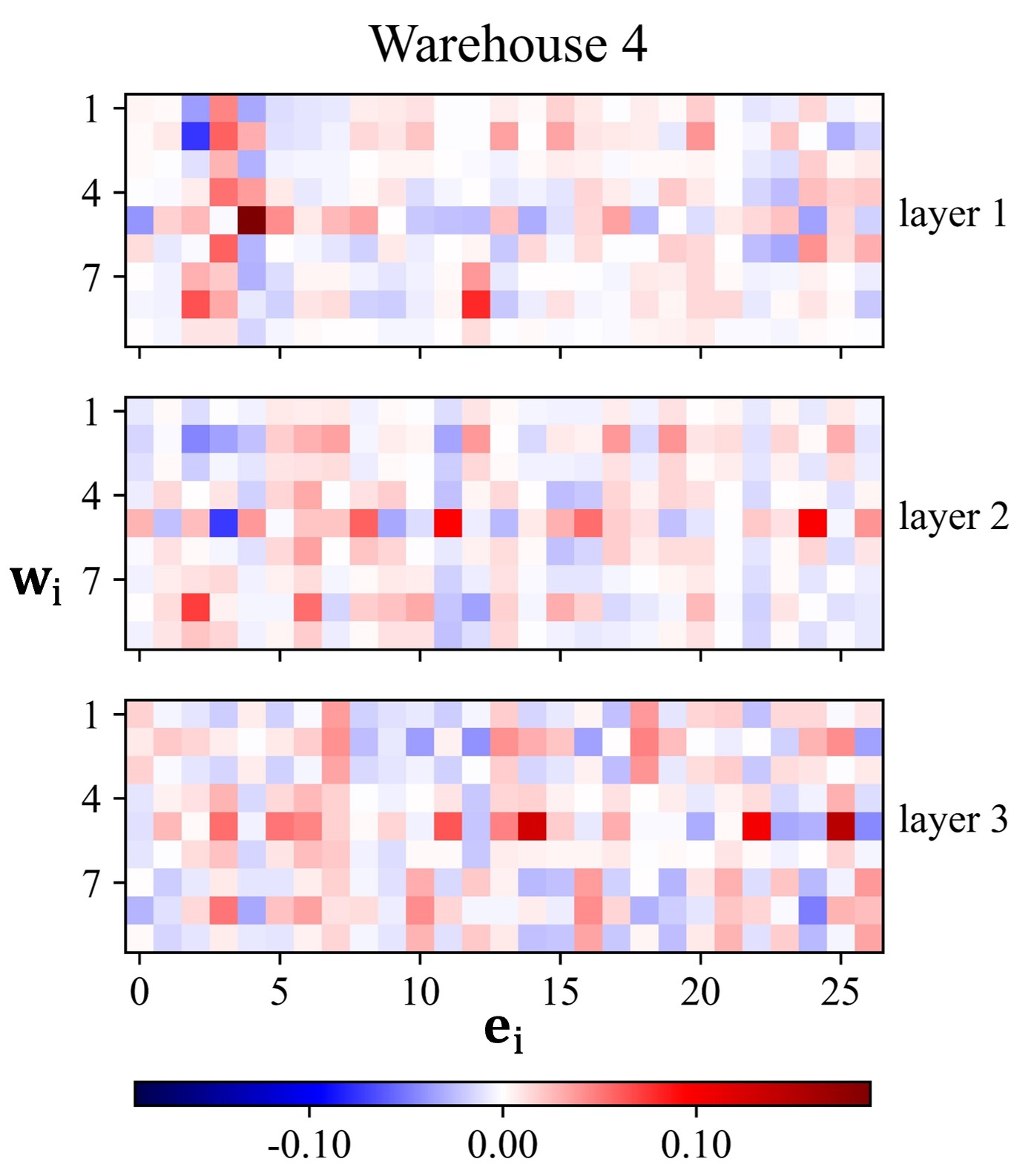}
            \end{center}
        \end{minipage}
    \subcaption{}
    \end{minipage}
    \begin{minipage}[t]{1.0\linewidth}
        \begin{minipage}[t]{0.24\linewidth}
            \begin{center}
                \includegraphics[width=\textwidth]{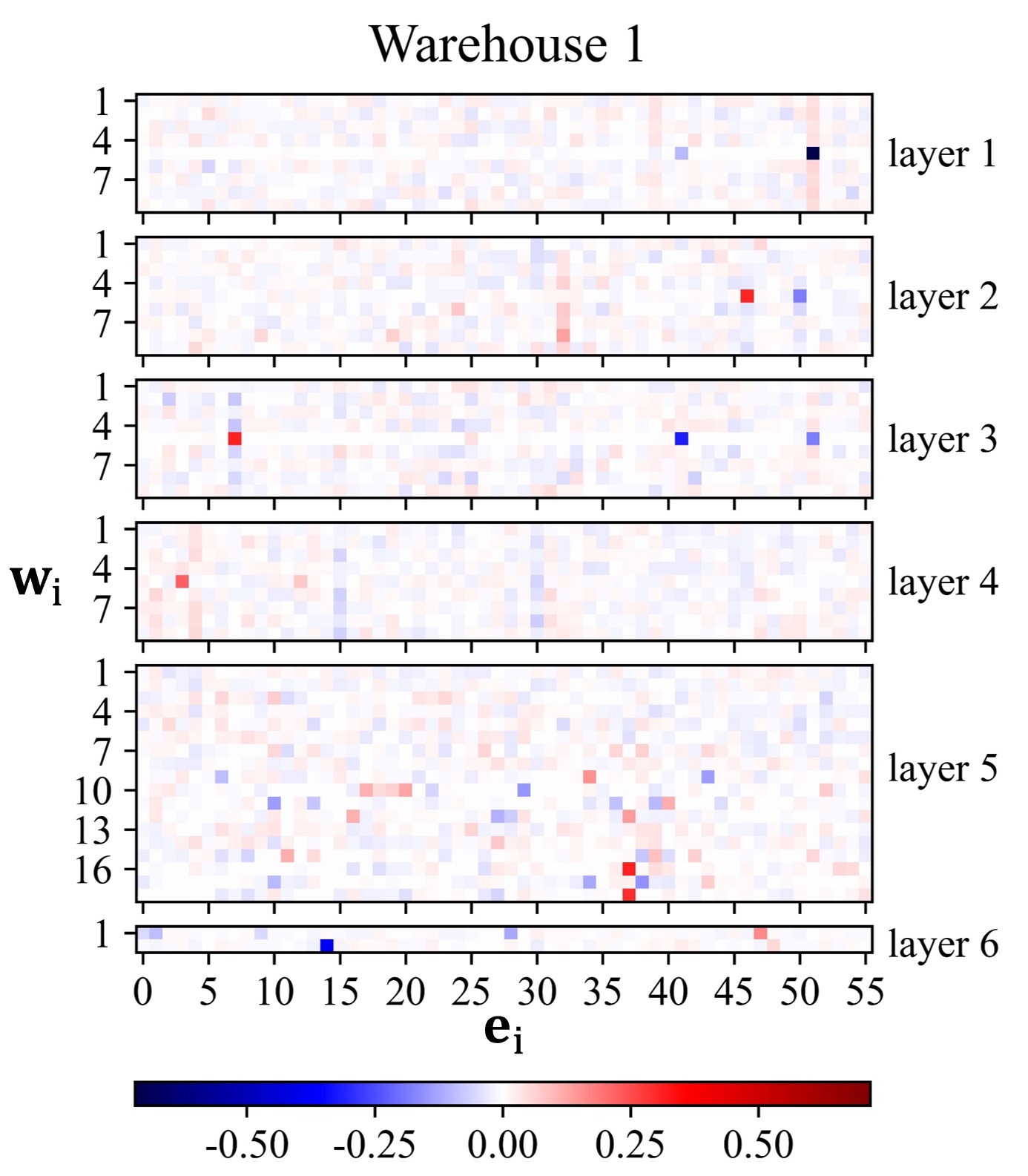}
            \end{center}
        \end{minipage}
        \hfill
        \begin{minipage}[t]{0.24\linewidth}
            \begin{center}
                \includegraphics[width=\textwidth]{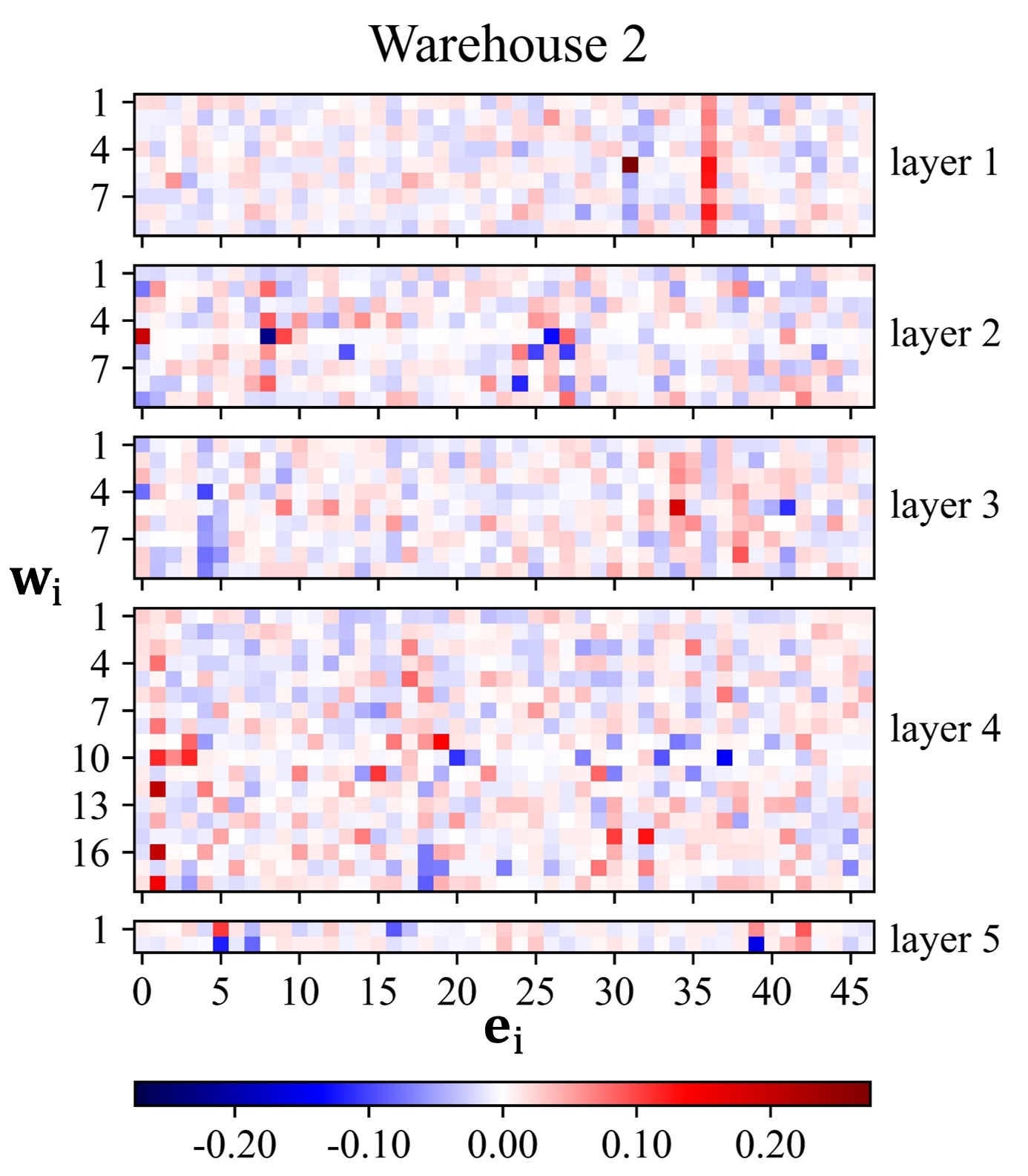}
            \end{center}
        \end{minipage}
        \hfill
        \begin{minipage}[t]{0.24\linewidth}
            \begin{center}
                \includegraphics[width=\textwidth]{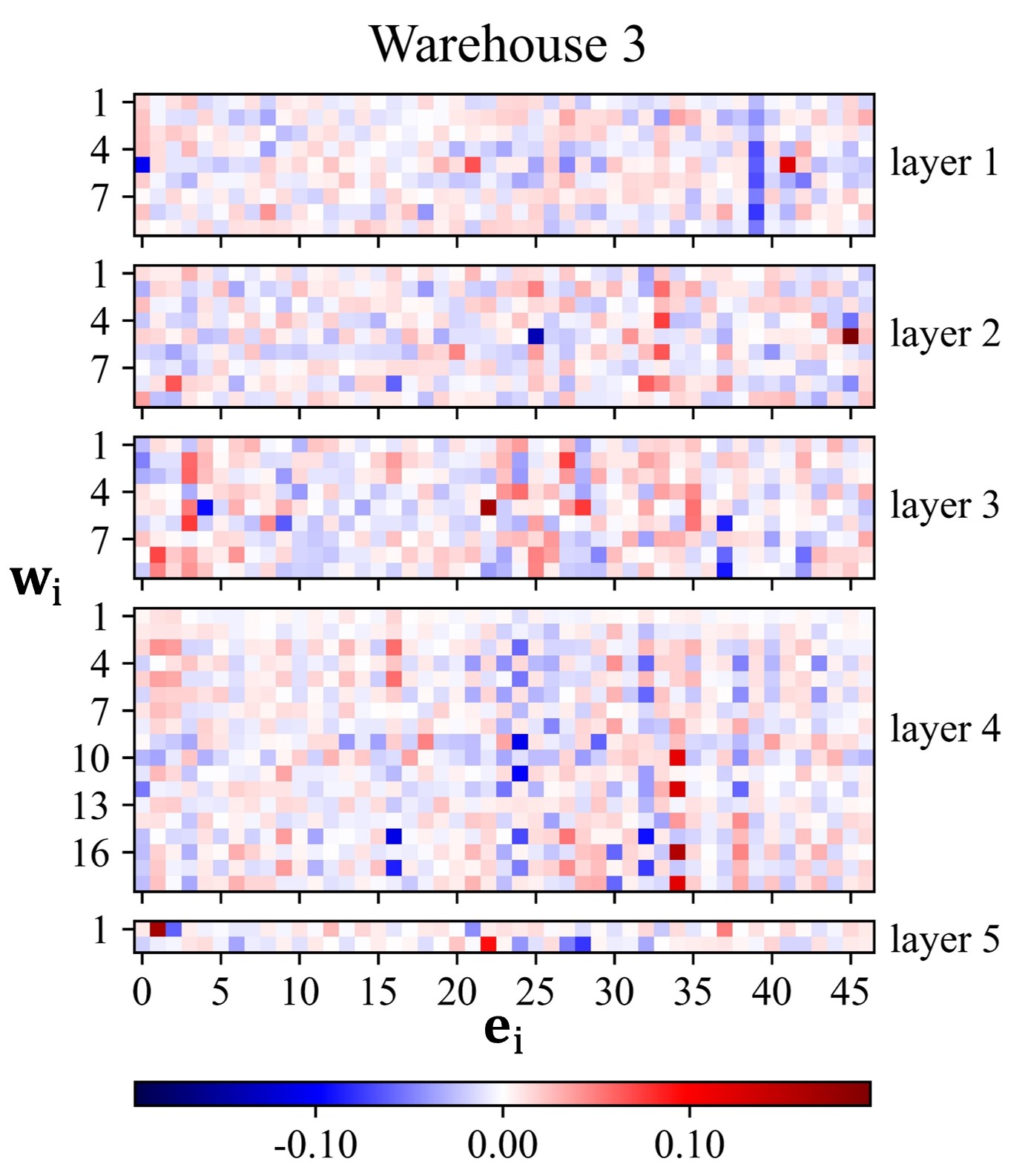}
            \end{center}
        \end{minipage}
            \hfill
        \begin{minipage}[t]{0.24\linewidth}
            \begin{center}
                \includegraphics[width=\textwidth]{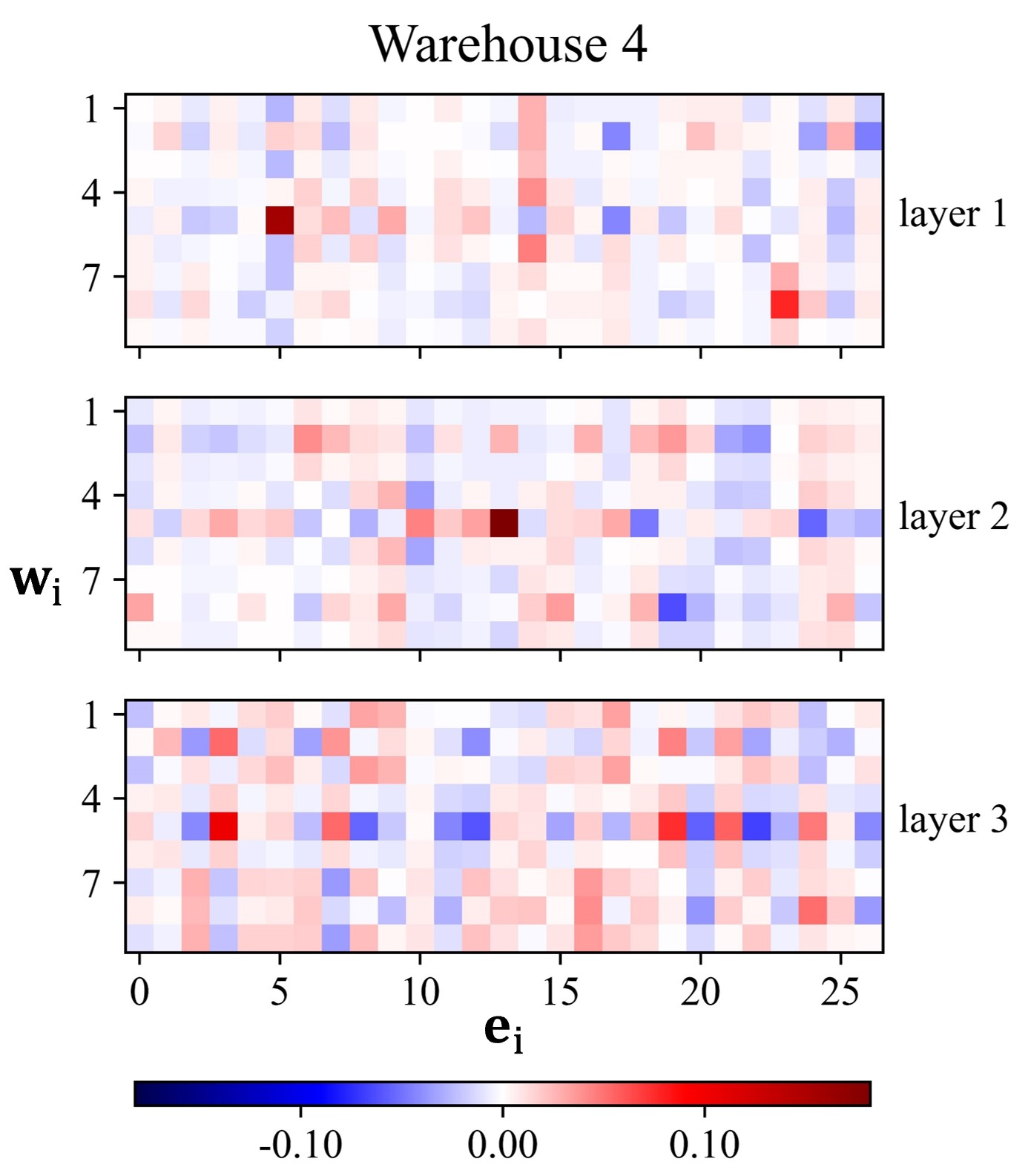}
            \end{center}
        \end{minipage}
    \subcaption{}
    \end{minipage}
    \caption{Visualization of statistical mean values of learnt attention $\alpha_{ij}$ in each warehouse for KernelWarehouse with different attentions initialization strategies. The results are obtained from the pre-trained ResNet18 backbone with KW ($1\times$) for all of the 50,000 images on the ImageNet validation dataset. Best viewed with zoom-in.
    The attentions initialization strategies for the groups of visualization results are as follows:
    (a) building one-to-one relationships between kernel cells and linear mixtures; (b) building all-to-one relationships between kernel cells and linear mixtures; (c) without initialization.}
    \label{fig:visualization_initialization_strategy_1x}

\end{figure}

\begin{figure}[thp]
    \begin{minipage}[t]{1.0\linewidth}
        \begin{minipage}[t]{0.49\linewidth}
            \begin{center}
                \includegraphics[width=\textwidth]{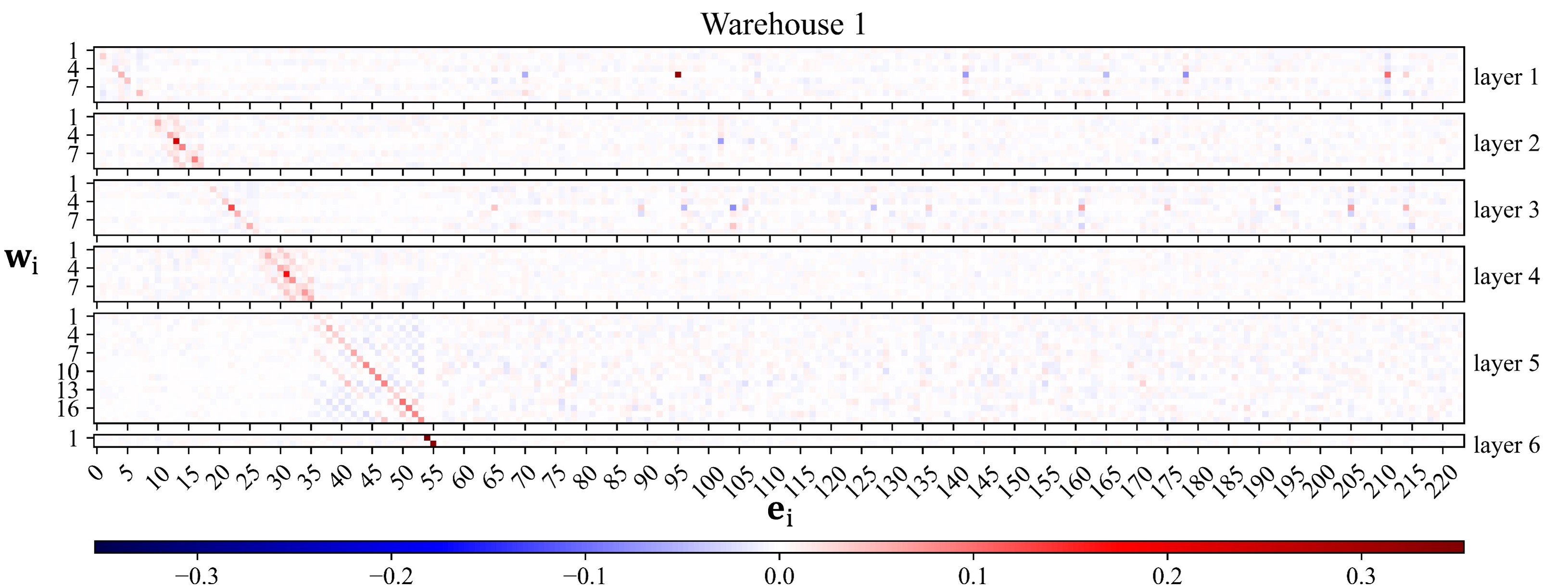}
            \end{center}
        \end{minipage}
        \hfill
        \begin{minipage}[t]{0.49\linewidth}
            \begin{center}
                \includegraphics[width=\textwidth]{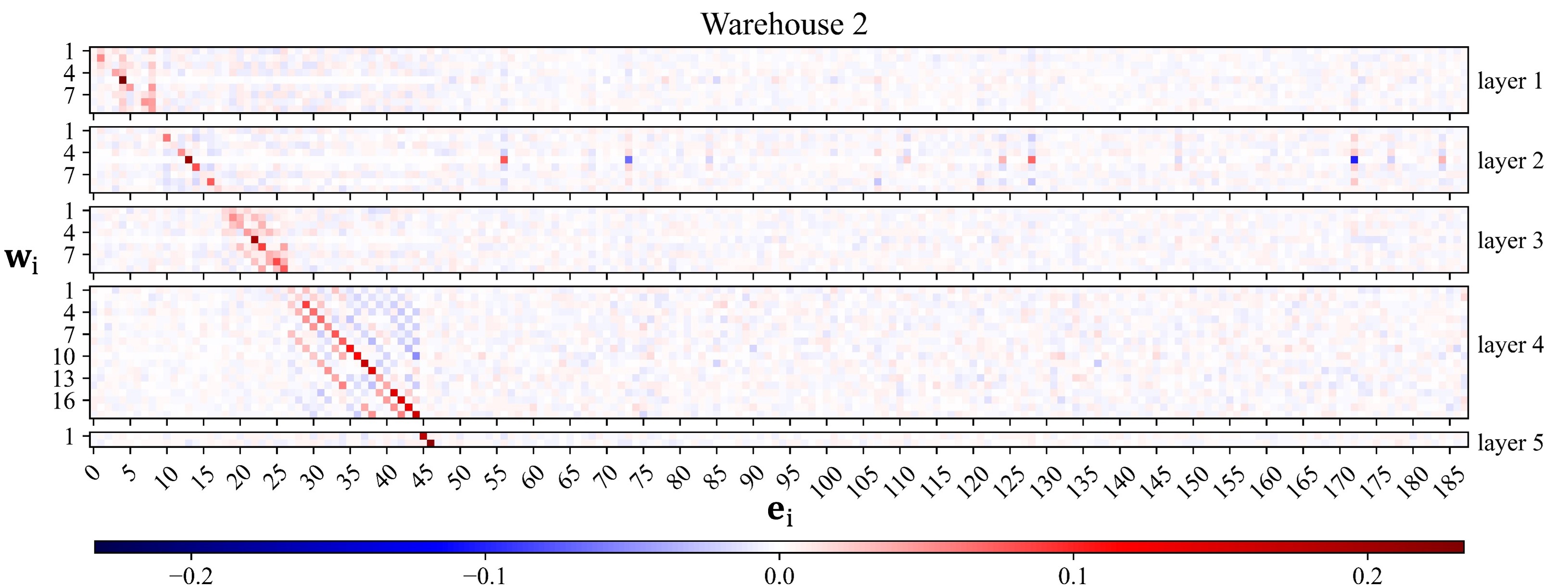}
            \end{center}
        \end{minipage}
        \hfill
        \begin{minipage}[t]{0.49\linewidth}
            \begin{center}
                \includegraphics[width=\textwidth]{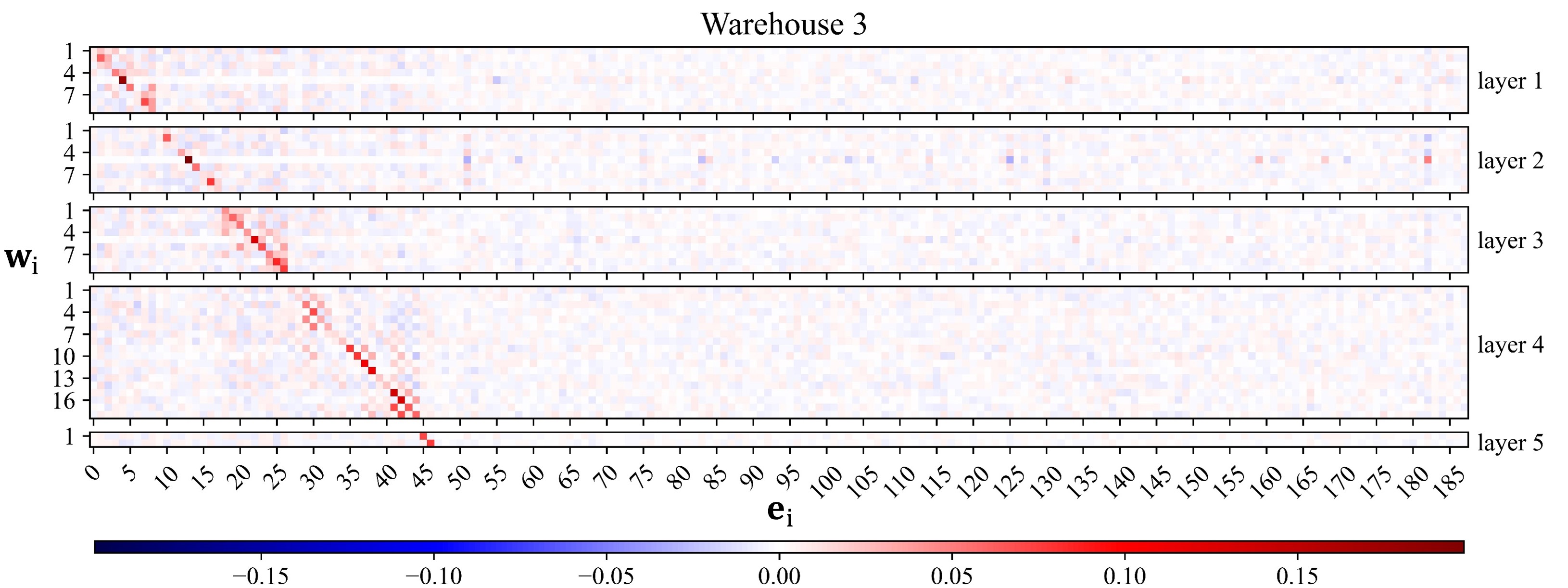}
            \end{center}
        \end{minipage}
            \hfill
        \begin{minipage}[t]{0.49\linewidth}
            \begin{center}
                \includegraphics[width=\textwidth]{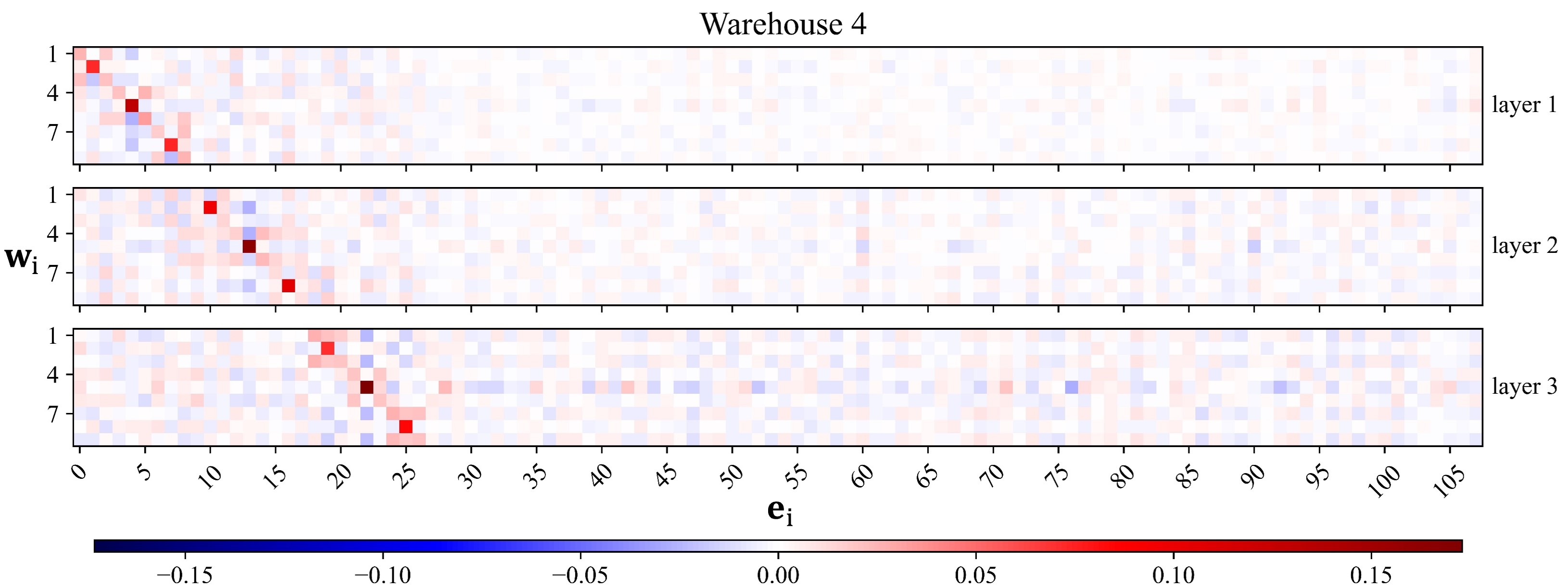}
            \end{center}
        \end{minipage}
    \subcaption{}
    \end{minipage}

    \begin{minipage}[t]{1.0\linewidth}
        \begin{minipage}[t]{0.49\linewidth}
            \begin{center}
                \includegraphics[width=\textwidth]{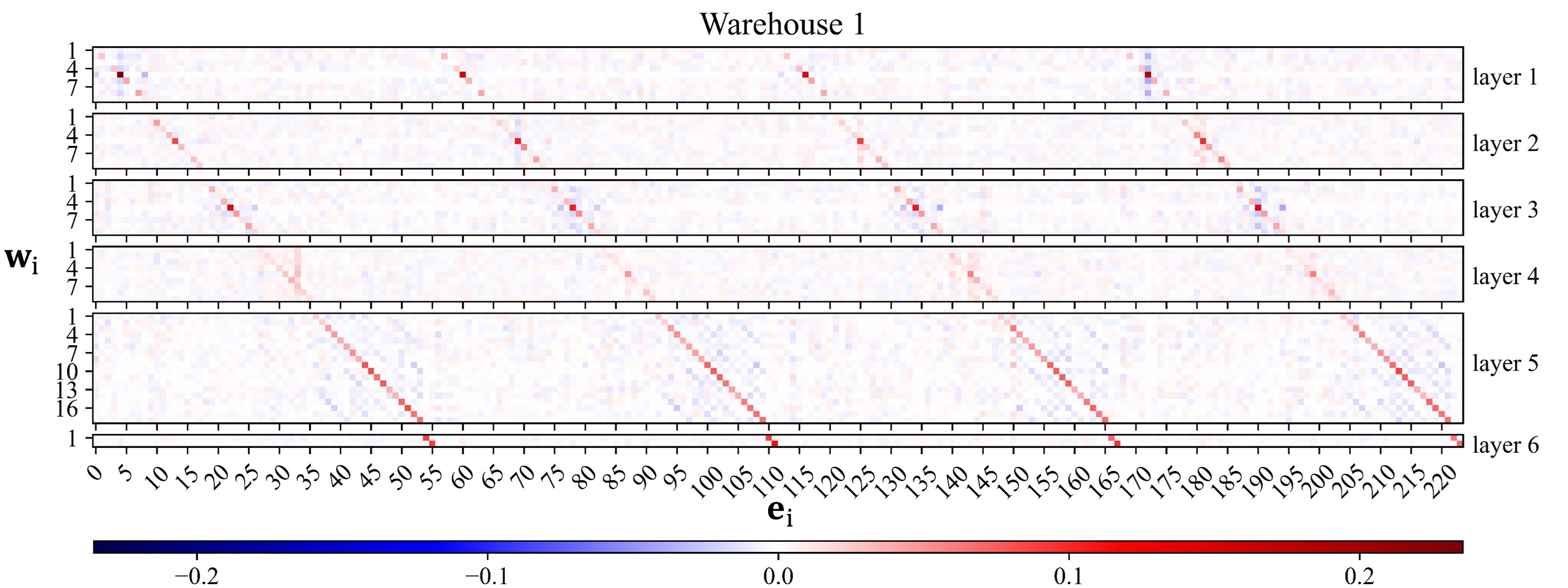}
            \end{center}
        \end{minipage}
        \hfill
        \begin{minipage}[t]{0.49\linewidth}
            \begin{center}
                \includegraphics[width=\textwidth]{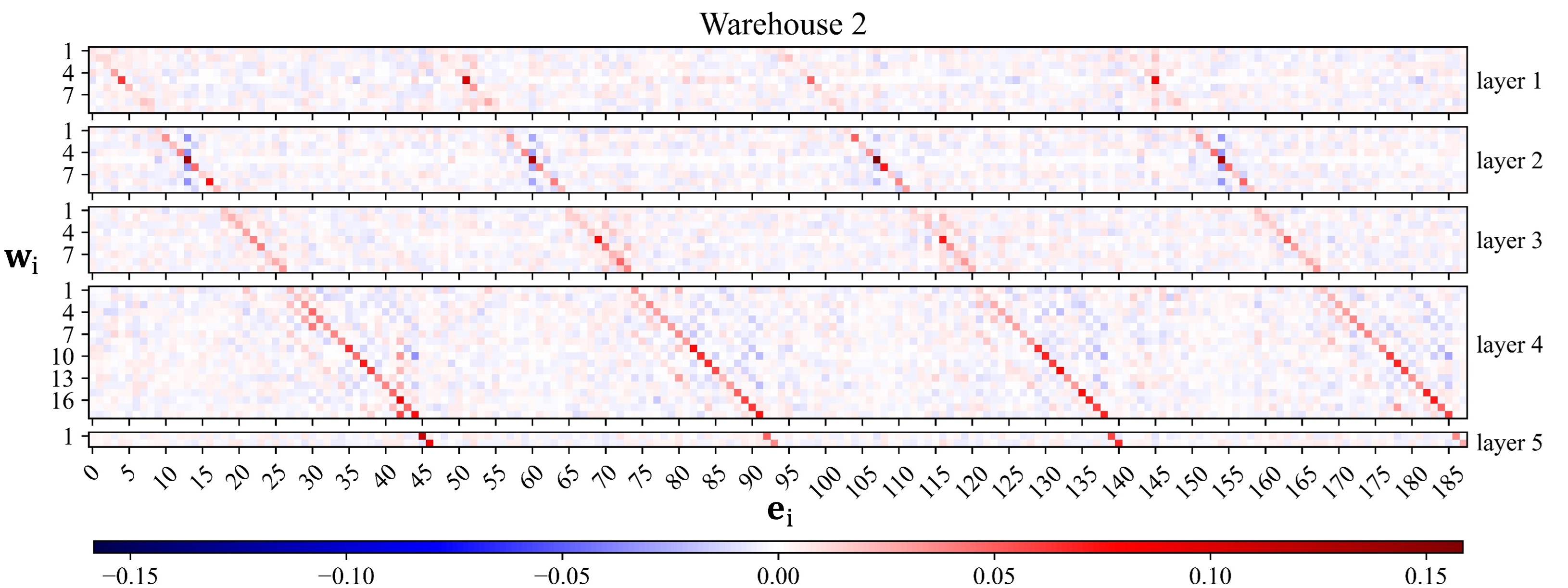}
            \end{center}
        \end{minipage}
        \hfill
        \begin{minipage}[t]{0.49\linewidth}
            \begin{center}
                \includegraphics[width=\textwidth]{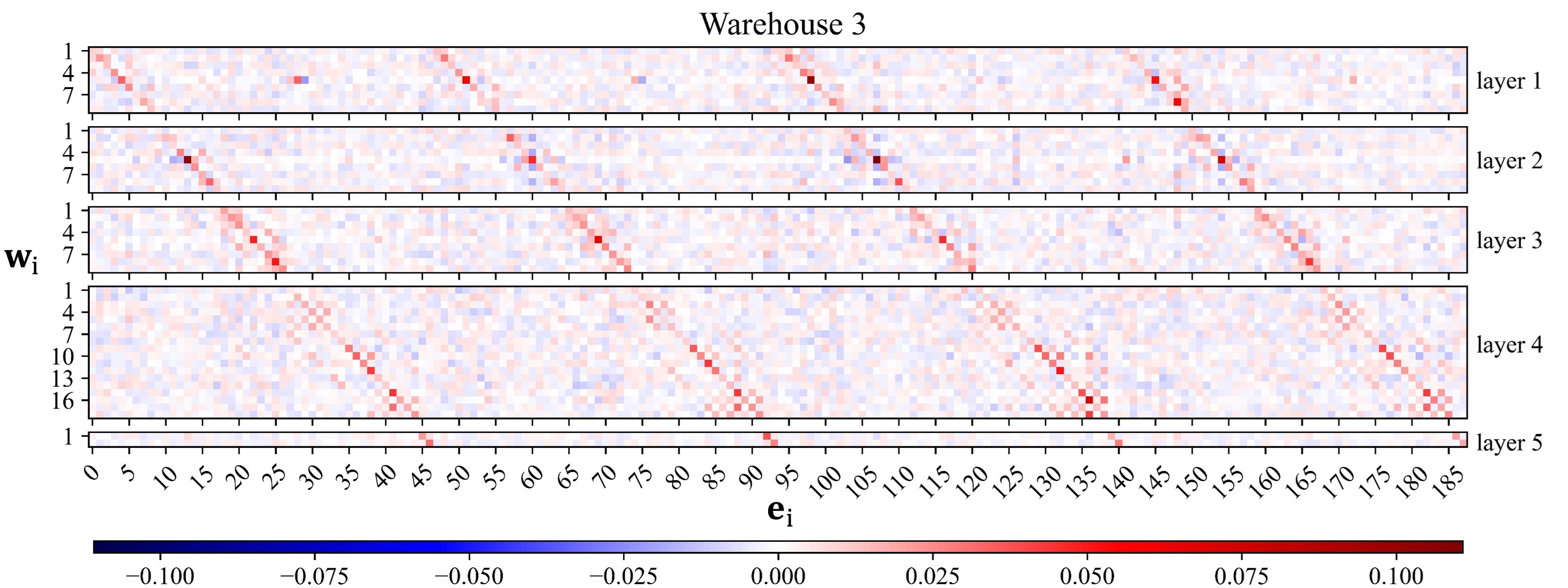}
            \end{center}
        \end{minipage}
            \hfill
        \begin{minipage}[t]{0.49\linewidth}
            \begin{center}
                \includegraphics[width=\textwidth]{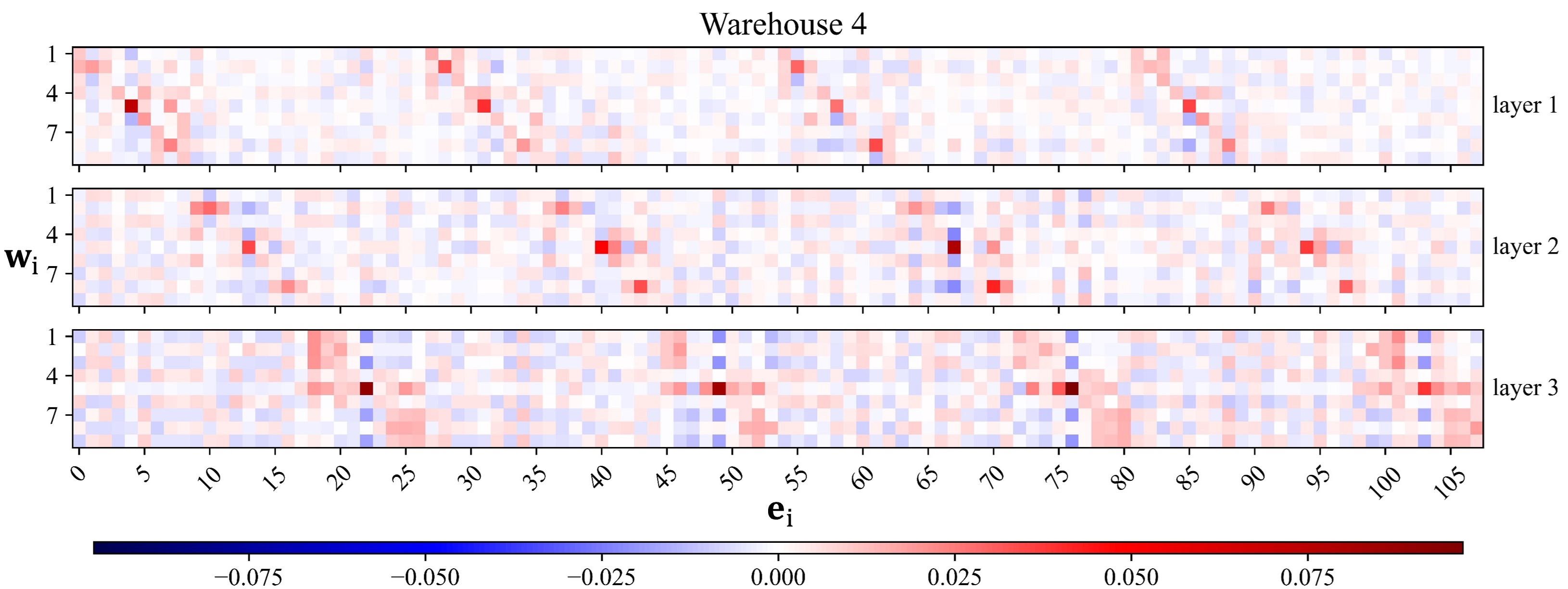}
            \end{center}
        \end{minipage}
    \subcaption{}
    \end{minipage}
    \caption{Visualization of statistical mean values of learnt attention $\alpha_{ij}$ in each warehouse for KernelWarehouse with different attentions initialization strategies. The results are obtained from the pre-trained ResNet18 backbone with KW ($4\times$) for all of the 50,000 images on the ImageNet validation dataset. Best viewed with zoom-in.
    The attentions initialization strategies for the groups of visualization results are as follows:
    (a) building one-to-one relationships between kernel cells and linear mixtures; (b) building four-to-one relationships between kernel cells and linear mixtures.}
    \label{fig:visualization_initialization_strategy_4x}

\end{figure}

\begin{figure}[thp]
    \begin{minipage}[t]{1.0\linewidth}
        \begin{minipage}[t]{0.24\linewidth}
            \begin{center}
                \includegraphics[width=\textwidth]{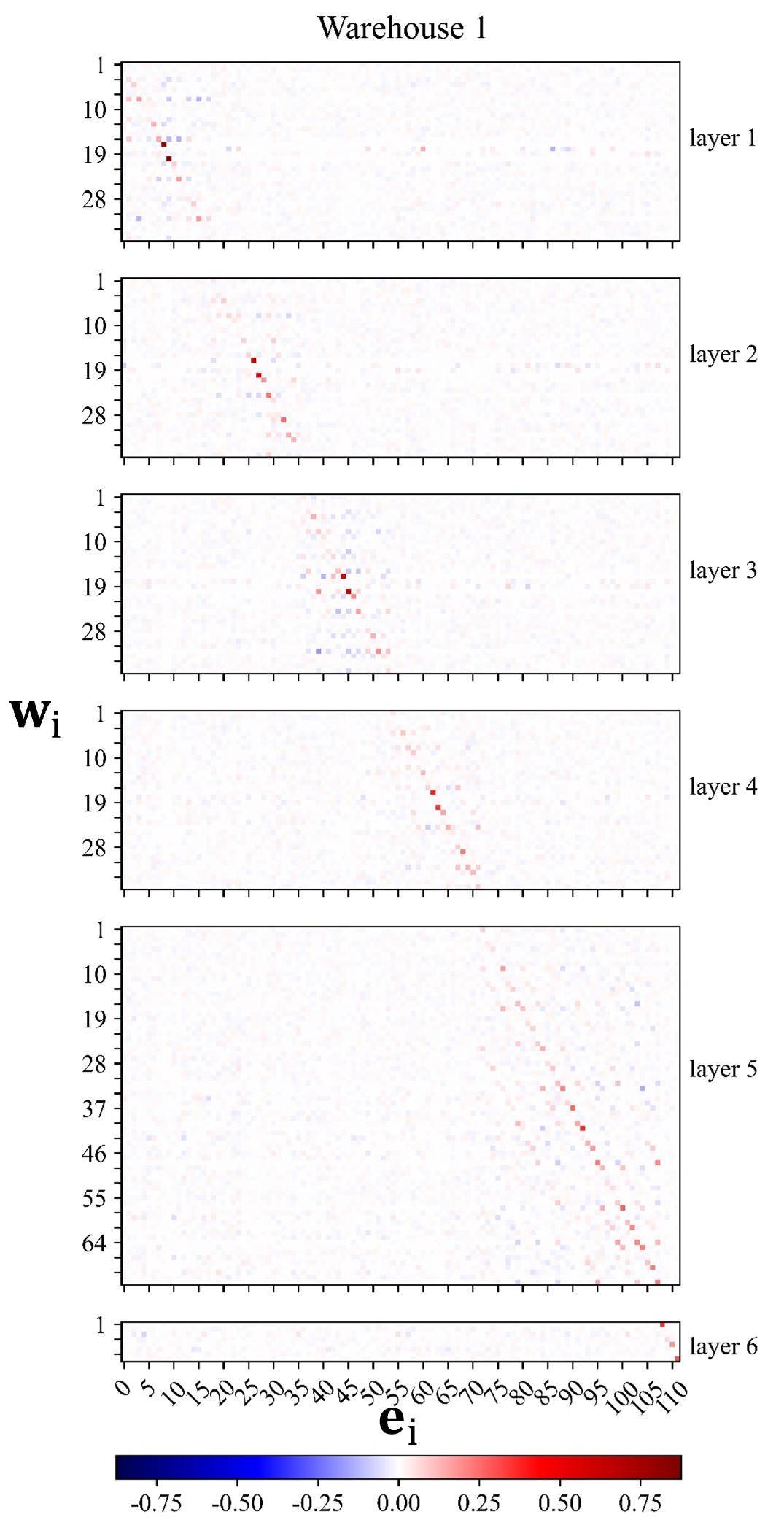}
            \end{center}
        \end{minipage}
        \hfill
        \begin{minipage}[t]{0.24\linewidth}
            \begin{center}
                \includegraphics[width=\textwidth]{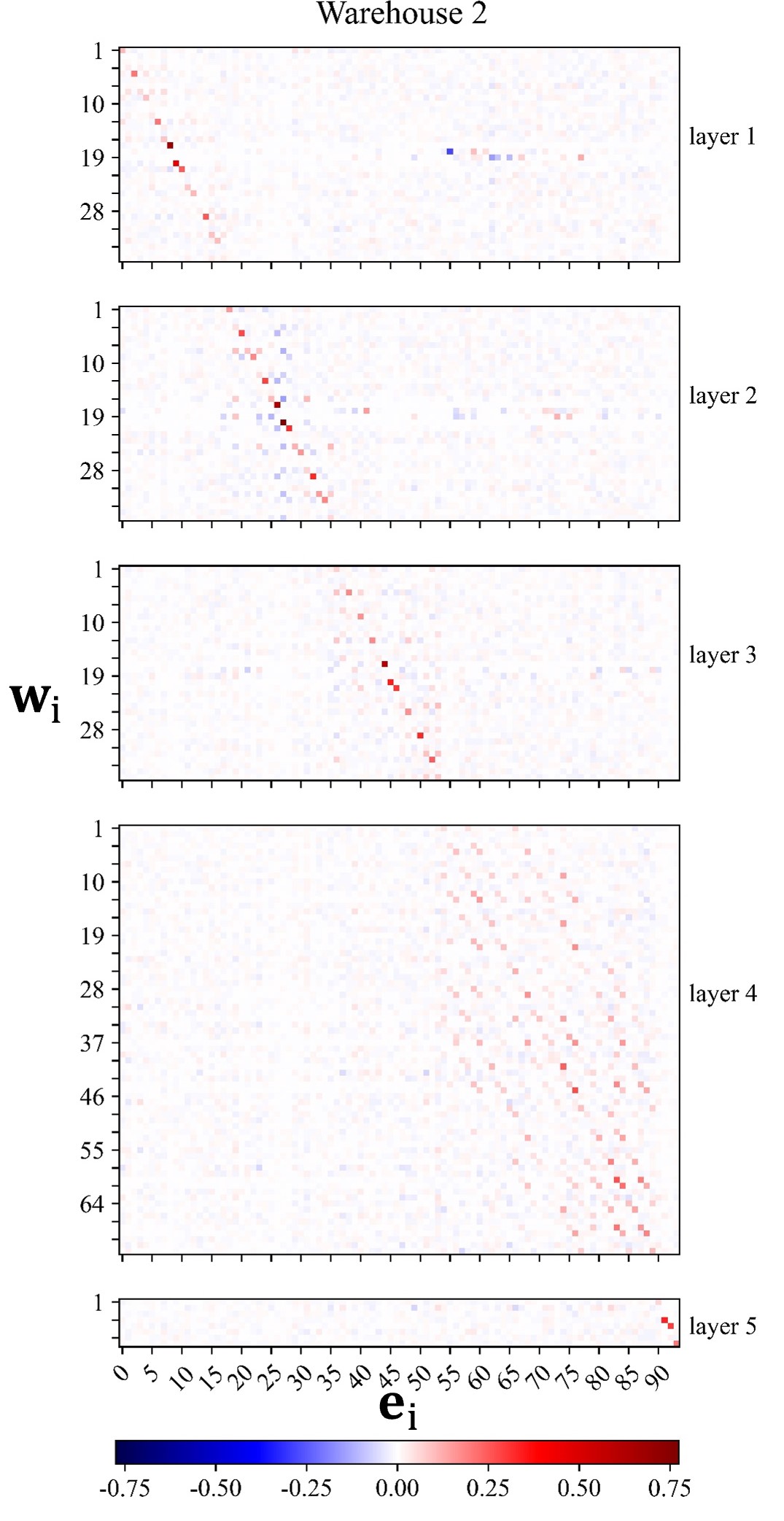}
            \end{center}
        \end{minipage}
        \hfill
        \begin{minipage}[t]{0.24\linewidth}
            \begin{center}
                \includegraphics[width=\textwidth]{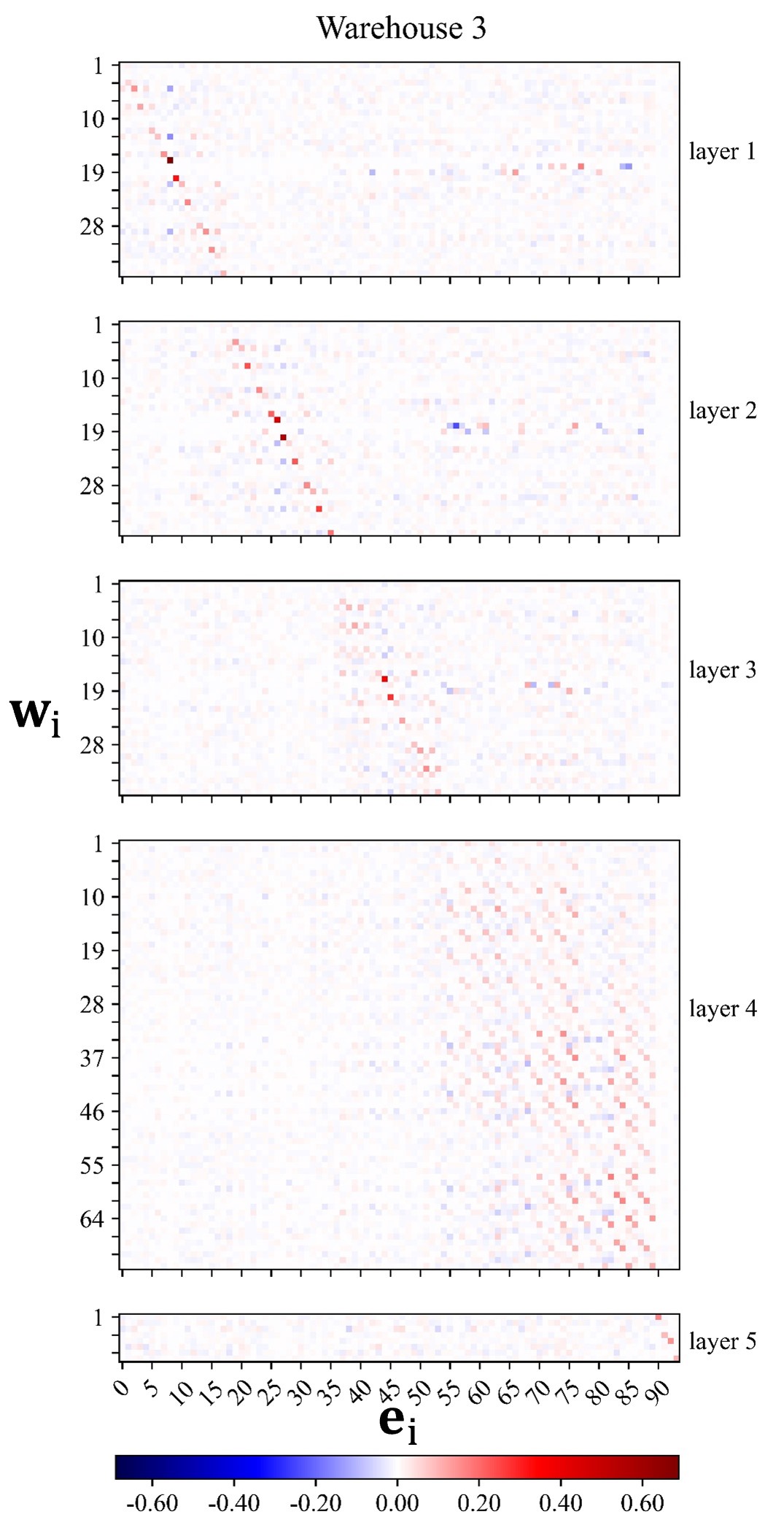}
            \end{center}
        \end{minipage}
            \hfill
        \begin{minipage}[t]{0.24\linewidth}
            \begin{center}
                \includegraphics[width=\textwidth]{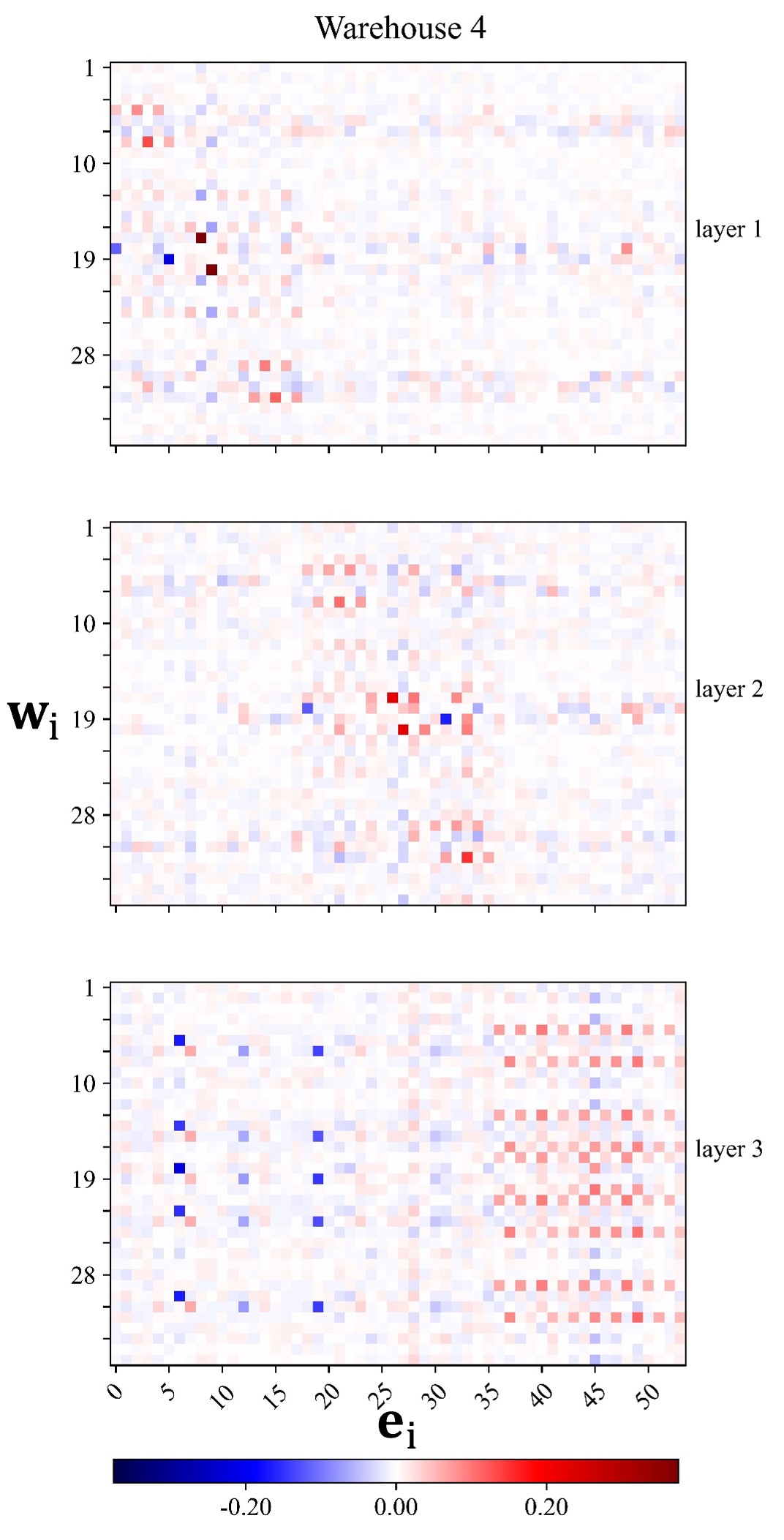}
            \end{center}
        \end{minipage}
    \subcaption{}
    \end{minipage}

    \begin{minipage}[t]{1.0\linewidth}
        \begin{minipage}[t]{0.24\linewidth}
            \begin{center}
                \includegraphics[width=\textwidth]{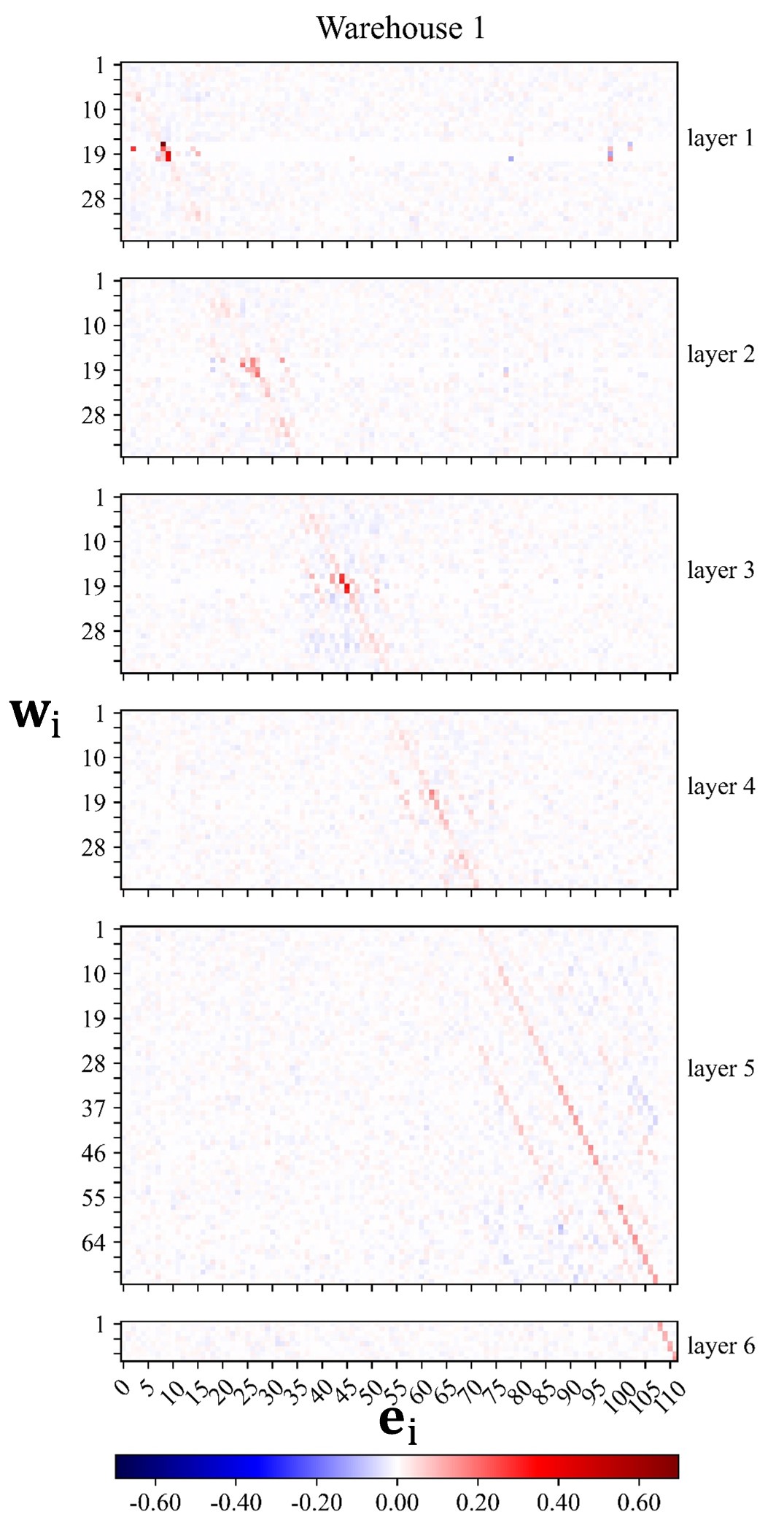}
            \end{center}
        \end{minipage}
        \hfill
        \begin{minipage}[t]{0.24\linewidth}
            \begin{center}
                \includegraphics[width=\textwidth]{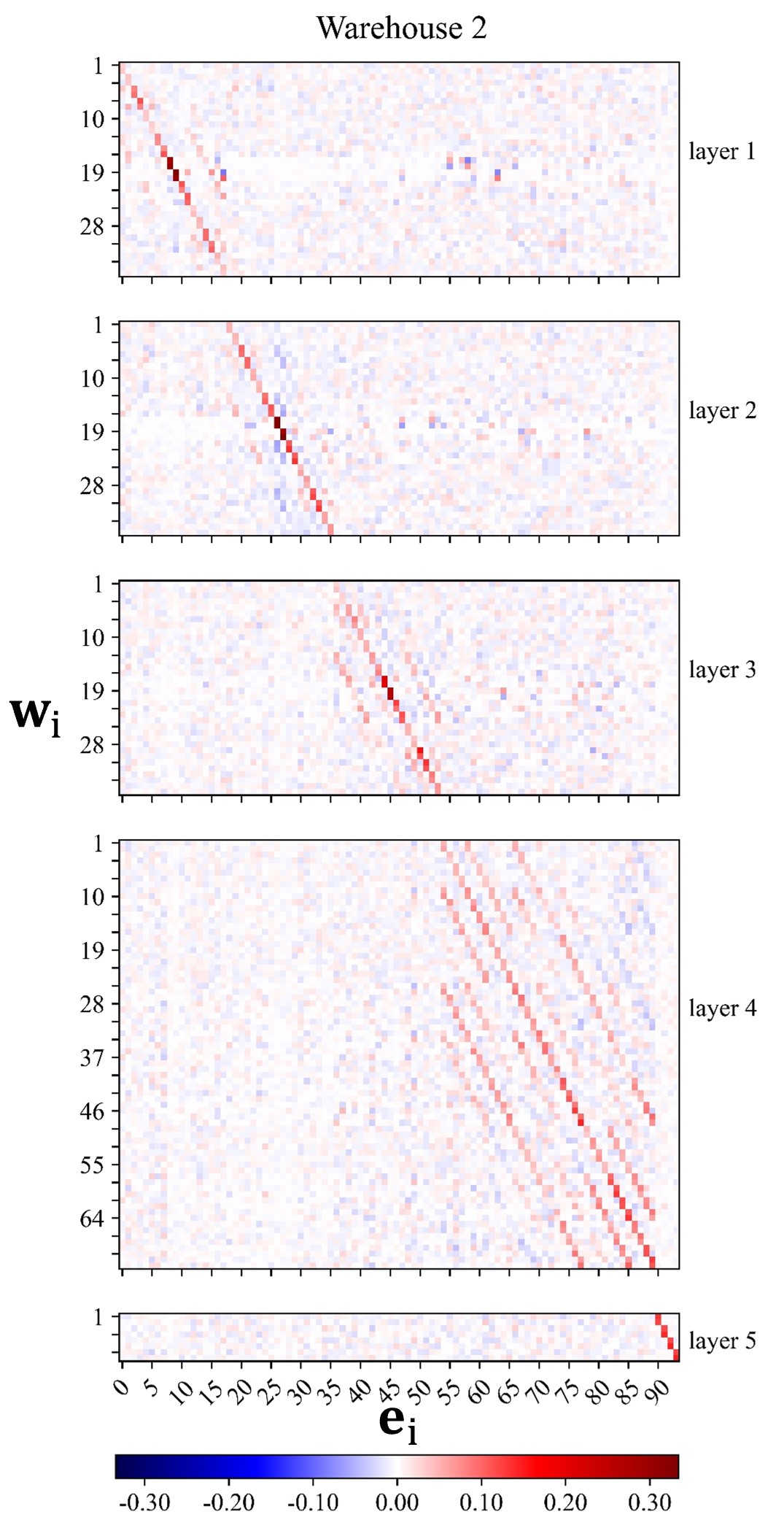}
            \end{center}
        \end{minipage}
        \hfill
        \begin{minipage}[t]{0.24\linewidth}
            \begin{center}
                \includegraphics[width=\textwidth]{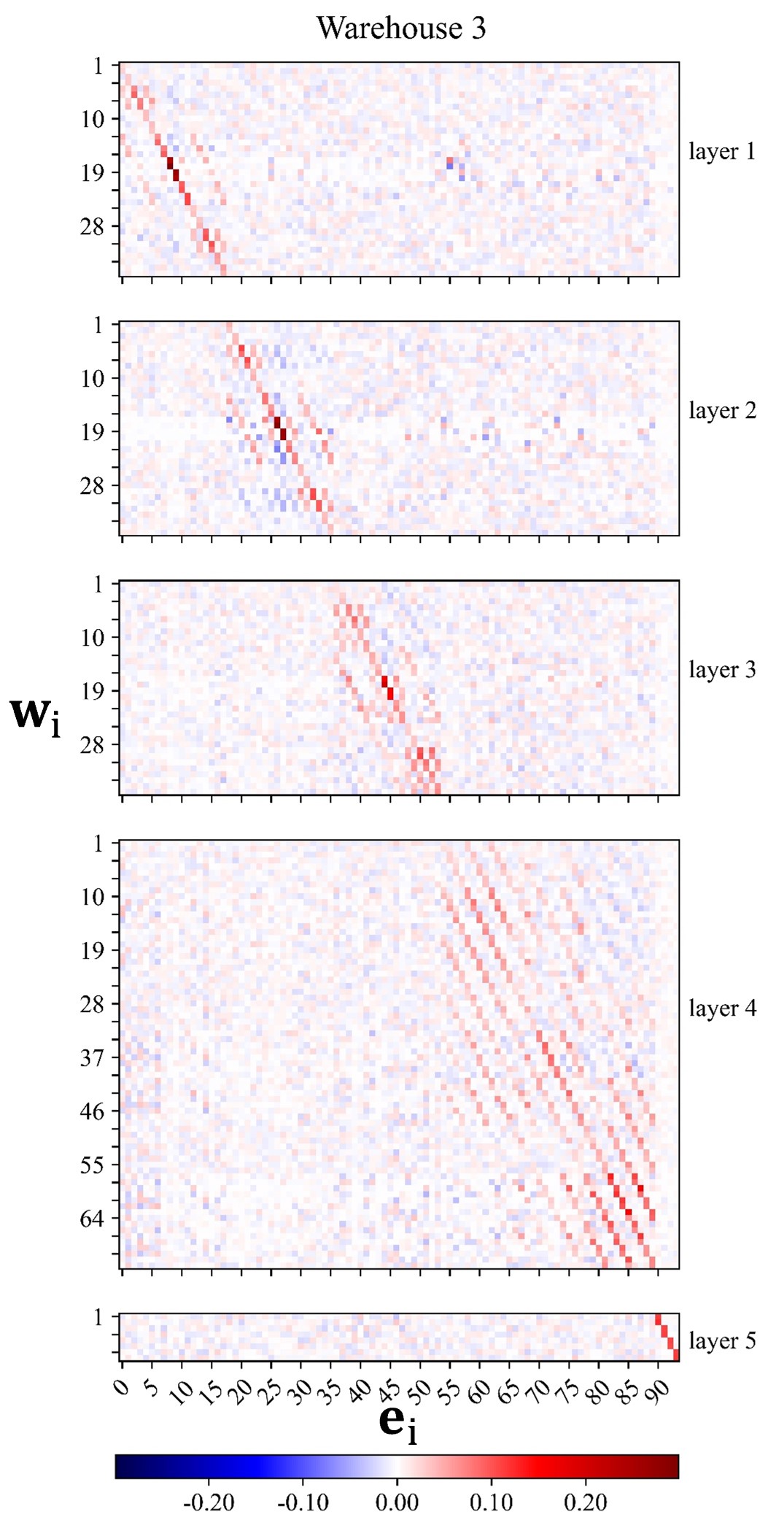}
            \end{center}
        \end{minipage}
            \hfill
        \begin{minipage}[t]{0.24\linewidth}
            \begin{center}
                \includegraphics[width=\textwidth]{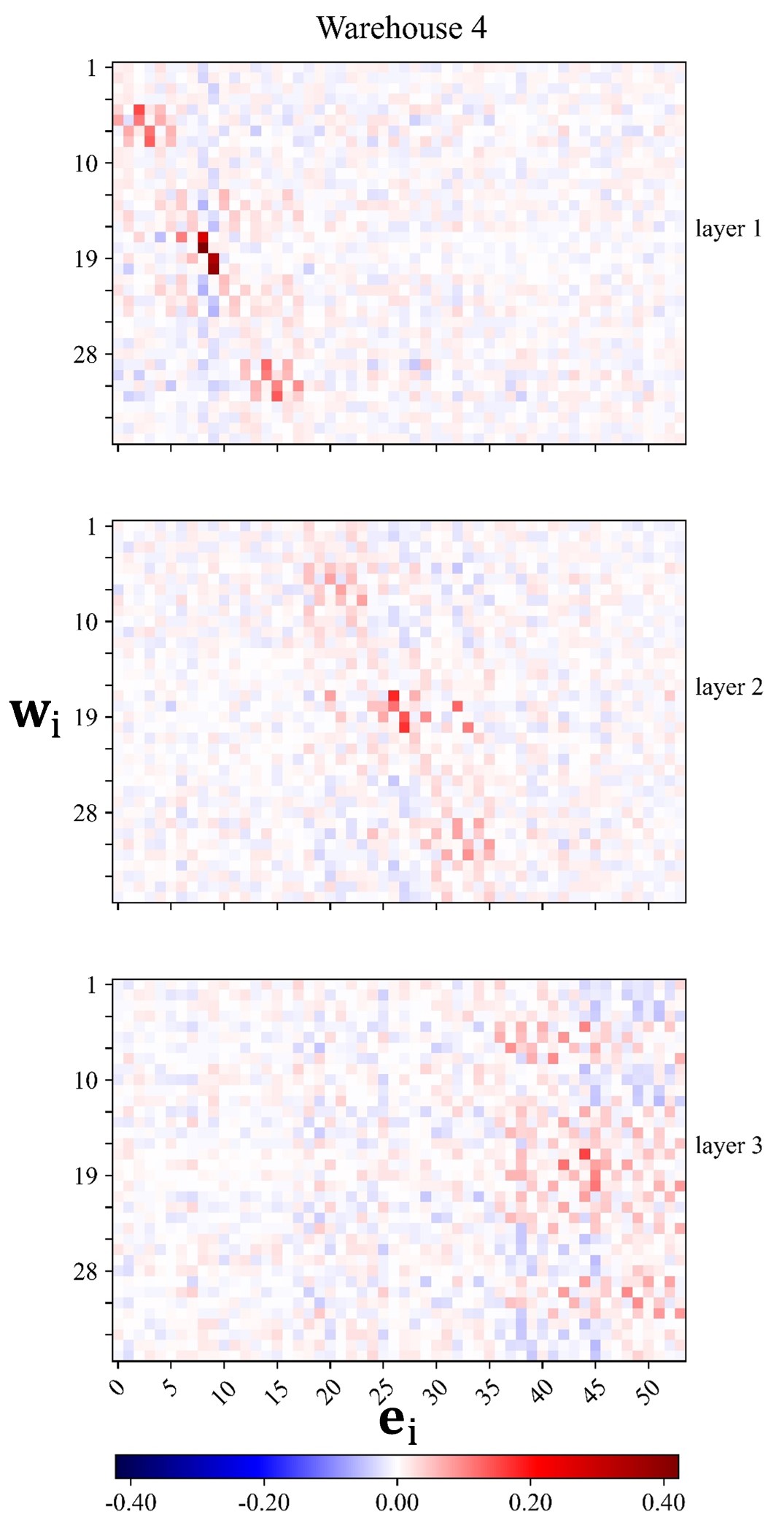}
            \end{center}
        \end{minipage}
    \subcaption{}
    \end{minipage}
    \caption{Visualization of statistical mean values of learnt attention $\alpha_{ij}$ in each warehouse for KernelWarehouse with different attentions initialization strategies. The results are obtained from the pre-trained ResNet18 backbone with KW ($1/2\times$) for all of the 50,000 images on the ImageNet validation dataset. Best viewed with zoom-in.
    The attentions initialization strategies for the groups of visualization results are as follows:
    (a) building one-to-one relationships between kernel cells and linear mixtures; (b) building one-to-two relationships between kernel cells and linear mixtures.}
    \label{fig:visualization_initialization_strategy_1d2x}

\end{figure}

\textbf{Visualization Results for KernelWarehouse with Attentions Initialization Strategies.}
The visualization results for KernelWarehouse with different attentions initialization strategies are shown in Figure~\ref{fig:visualization_initialization_strategy_1x}, Figure~\ref{fig:visualization_initialization_strategy_4x} and Figure~\ref{fig:visualization_initialization_strategy_1d2x}, which are corresponding to the comparison results of Table~\ref{table:ablation_initialization}. From which we can observe that: (1) with all-to-one strategy or without initialization strategy, the distribution of scalar attentions learnt by KernelWarehouse seems to be disordered, while our proposed strategy can help the ConvNet learn more appropriate relationships between kernel cells and linear mixtures;
(2) for KW ($4\times$) and KW ($1/2\times$), it's hard to directly determine which strategy is better only according to the visualization results. While the results demonstrate that the learnt attentions of KernelWarehouse are highly related to our setting of $\alpha_{ij}$;
(3) for KW ($1\times$), KW ($4\times$) and KW ($1/2\times$) with our proposed initialization strategy, some similar patterns of the value distributions can be found.
For example, the maximum value of $\alpha_{ij}$ in each row mostly appears in the diagonal line throughout the whole warehouse. It indicates that our proposed strategy can help the ConvNet learn stable relationships between kernel cells and linear mixtures.


\end{document}